\documentclass[11pt]{article}

\usepackage[preprint]{acl}

\usepackage{times}
\usepackage{latexsym}

\usepackage[T1]{fontenc}

\usepackage[utf8]{inputenc}
\usepackage{microtype}

\usepackage{inconsolata}

\usepackage{graphicx}
\usepackage{amsmath}
\usepackage{amssymb}
\usepackage{dsfont}
\usepackage{booktabs}
\usepackage{multirow}
\usepackage{array}
\usepackage{subcaption}
\usepackage{longtable}
\usepackage{hyperref}
\usepackage{cleveref}
\usepackage{algorithm}
\usepackage{algpseudocode}

\newcommand{\km}{\mathrm{K}_{\mathcal{M}}}

\newcommand{\squishlist}{
\begin{list}{$\bullet$}
{   \setlength{\itemsep}{0pt}
   \setlength{\parsep}{3pt}
   \setlength{\topsep}{3pt}
   \setlength{\partopsep}{0pt}
   \setlength{\leftmargin}{1.5em}
   \setlength{\labelwidth}{1em}
   \setlength{\labelsep}{0.5em} } }
\newcounter{Lcount}
\newcommand{\squishlisttwo}{
\begin{list}{\arabic{Lcount}. }
  { \usecounter{Lcount}
 \setlength{\itemsep}{0pt}
 \setlength{\parsep}{0pt}
 \setlength{\topsep}{0pt}
 \setlength{\partopsep}{0pt}
 \setlength{\leftmargin}{1.5em}
 \setlength{\labelwidth}{1em}
 \setlength{\labelsep}{0.5em} } }
\newcommand{\squishend}{\end{list} }
\usepackage{enumitem}
\usepackage{csquotes}

\usepackage{adjustbox}

\usepackage{titletoc}
\newcommand\DoToC{%
  \startcontents
  \printcontents{}{1}{}{}
}

\setlist[enumerate]{itemsep=1pt,topsep=1pt}

\title{The Trilemma of Truth in Large Language Models}

\author{
Germans Savcisens\textsuperscript{\textnormal{1,2}} and
Tina Eliassi-Rad\textsuperscript{\textnormal{1,2,3}}
\\
 \textsuperscript{1} Khoury College of Computer Sciences,
Northeastern University, Boston, MA, USA \\
 \textsuperscript{2} Network Science Institute,
Northeastern University, Boston, MA, USA \\
 \textsuperscript{3} Santa Fe Institute, Santa Fe, NM, USA
\\
 \small{
 \texttt{\{g.savcisens, t.eliassirad\}@northeastern.edu}
 }
 \\
\small{Accepted to EMNLP 2026 (Main Conference)}
}

\begin{document}
\maketitle

\begin{abstract}
The public often attributes human-like qualities to large language models (LLMs), assuming that they ``know'' certain things. In reality, LLMs encode information retained during training as internal probabilistic knowledge. This study examines existing methods for probing the veracity of that knowledge and identifies three flawed underlying assumptions. To address these flaws, we introduce \texttt{sAwMIL} (Sparse-Aware Multiple-Instance Learning), a multiclass probing framework that combines multiple-instance learning with conformal prediction. \texttt{sAwMIL} leverages LLMs' internal representations to classify statements as \textit{true}, \textit{false}, or \textit{neither}. We evaluate \texttt{sAwMIL} across 16 open-source LLMs, including default and chat-based variants, using three new curated datasets. Our results show that (1) common probing methods fail to provide a reliable and transferable veracity direction and, in some settings, perform worse than zero-shot prompting; (2) truth and falsehood are not encoded symmetrically; and (3) LLMs encode a third type of signal that is distinct from both true and false.\footnote{\textbf{Code and data}: \href{https://github.com/carlomarxdk/trilemma-of-truth}{\texttt{carlomarxdk/trilemma-of-truth}}.}
\end{abstract}

\section{Introduction}

Can we trust the content generated by large language models (LLMs)? Recent literature suggests that LLMs possess internal probabilistic knowledge~\cite{petroni-etal-2019-language, meng2022locating,chen2024journey, chanin-etal-2024-identifying, panprecise}.\footnote{In this paper, we consider factual associations to be probabilistic knowledge.} However, our understanding of \textit{how} LLMs use this internal knowledge remains fragmented.~\citet{hicks2024chatgpt} argue that LLMs lack intent regarding the veracity of their outputs and often hallucinate~\cite{huang2025survey}. Furthermore, it is often difficult for users to recognize hallucinations because LLMs produce fluent and persuasive text. For example,~\citet{church2024emerging} shows that students trust factually incorrect answers from LLMs due to their authoritative and confident tone, and~\citet{williams2025large} demonstrate that users rate disinformation generated by LLMs as equally or even more credible than human-generated content. Finally,~\citet{suzgun2025language} show that LLMs rely on inconsistent reasoning strategies for epistemic tasks and fail to separate what they believe from what is factually true. Therefore, \textit{we need novel methods to assess the truthfulness of internal probabilistic knowledge to make user interactions with LLMs more reliable.}

\begin{figure*}[!t]
    \centering
    \includegraphics[width=\linewidth]{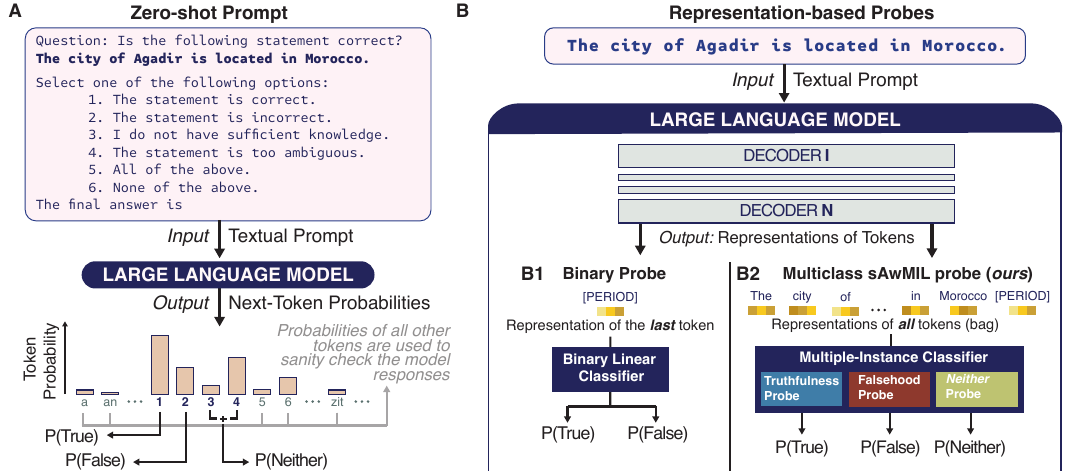}
    \caption{Overview of methods for probing veracity in LLMs.
    (\textbf{A}) In zero-shot prompting, a target statement is inserted into a structured prompt that instructs the LLM to select an answer. This method treats the model as a black box and examines its \(\langle\)input, output\(\rangle\) pairs.
    (\textbf{B})  In representation-based probing, we perform analysis on the internal representations generated by intermediate decoders.
    (\textbf{B1}) Commonly used binary probes (e.g., mean-difference) output probabilities for \textit{true} or \textit{false} statements, but cannot account for statements that lack a definitive truth value.
    (\textbf{B2}) Our probe, multiclass sparse-aware MIL (\texttt{sAwMIL}), looks at the representation of every token in a statement and provides probabilities for three classes: \textit{true}, \textit{false}, and \textit{neither}. Multiclass \texttt{sAwMIL} can account for cases when the LLM does not have any knowledge about the statement.
}
    \label{fig:flow}
\end{figure*}

Prompt-based methods (\Cref{fig:flow}.A) rely on simply asking an LLM about its knowledge. \citet{yadkori2024believe} introduce information-theoretic prompt-based evaluation, while \citet{xu-etal-2024-sayself} propose training frameworks to produce prompts with self-reflective rationales, and \citet{farquhar2024detecting} introduce uncertainty estimators to detect inconsistent text generations. However, prompt-based evaluations are sensitive to the input's phrasing~\cite{lu-etal-2022-fantastically} and content~\cite{mcilroy2024order}.

A more direct approach is to examine \textit{how} LLMs represent text internally (\Cref{fig:flow}.B). Consider an LLM $\mathcal{M}$ with a vocabulary $\mathcal{V}$.
The LLM $\mathcal{M}$ maps the input text $\boldsymbol{x}$ to a probability distribution over subsequent tokens, denoted $P_{\mathcal{M}}$:
\begin{equation}
    \mathcal{M}\left(\boldsymbol x \right) = P_{\mathcal M}(\tau \mid \boldsymbol x) \text{, where } \tau \in \mathcal{V}.
\end{equation}
For any token $\tau \in \mathcal{V}$, the distribution $P_{\mathcal{M}}\left(\tau \mid \boldsymbol x\right)$ denotes the probability that $\tau$ is the continuation of the sequence \(\boldsymbol{x}\).\footnote{\Cref{tab:notation,tab:abbreviation}, respectively, list the notations and abbreviations used in this paper.}
To compute the conditional distribution $P_{\mathcal{M}}$, an LLM transforms \(\boldsymbol{x}\) into \textit{intermediate} neural representations $h_i\left(\boldsymbol{x}\right) \in \mathbb{R}^{ L \times d}$.
Here $h_i\left(\boldsymbol{x}\right)$ denotes the representation of $\mathcal{M}$ after the $i$th decoder, $d$ stands for the dimensionality of the decoder, and $L$ stands for the length of the sequence $\boldsymbol{x}$.
We can probe these intermediate representations to identify veracity signals, i.e., to isolate patterns that correspond to truthful statements.
For example, \citet{azaria-mitchell-2023-internal} train a neural network to classify statements as \textit{true} or \textit{false} based on these internal representations.
Similarly, \citet{marks2023geometry} use a mean-difference classifier to linearly separate \textit{true} from \textit{false} statements (\Cref{fig:flow}.B1).
\citet{burger2024truth} introduce the truthfulness-polarity classifier, \citet{burns2022discovering} propose a semi-supervised method based on the contrastive pairs of statements, and \citet{liu2025training} describe a training-free ensemble approach based on correlations between the representations and truthfulness.

Collectively, these works rely on a shared idea: given statements \(\boldsymbol x\) with veracity labels \(y \in Z \) from a dataset \(\mathcal D\), train a probe \(g_i\) that maps internal representations $h_i\left(\boldsymbol{x}\right)$ to the distribution \(G_{\mathcal M}\) over the LLM \(\mathcal{M}\)'s veracity labels:
\begin{equation}
\label{eq:probe}
    g_i\left(h_i\left(\boldsymbol x\right)\right) = G_{\mathcal M} ( z \mid \boldsymbol{x}) \text{,  } z \in \{\text{true}, \text{false}\}
\end{equation}
However, we observe that existing probing methods often rely on flawed assumptions, thereby limiting the reliability of current findings. 
Instead, we argue for a three-valued logic approach (\Cref{fig:flow}.B2) as the more appropriate method to model veracity in LLMs.

Our method \texttt{sAwMIL} (short for Sparse-Aware Multiple-Instance Learning) combines Multiple-Instance Learning~\cite{bunescu2007multiple} and Conformal Prediction~\cite{romano2020classification} to allow for a flexible probe that can handle \textit{neither} statements and quantify uncertainty about statements in \(\km\).

Our contributions include the following: \textbf{(1)} We identify and discuss three flawed assumptions in the current veracity-probing literature. \textbf{(2)} We show that common linear classifiers do not capture reliable veracity directions. \textbf{(3)} We propose a novel multiclass linear-probing method, \texttt{sAwMIL}, based on Multiple-Instance Learning~(MIL) and Conformal Prediction~(CP). \textbf{(4)} We present three new datasets containing statements labeled  \textit{true}, \textit{false}, and \textit{neither}\footnote{We italicize \textit{neither}, \textit{true}, and \textit{false} to distinguish them from the regular use of these words.} to enable more rigorous evaluations of veracity probes.

\section{Background and Flawed Assumptions}

\label{sec:veracity}
A large language model \(\mathcal{M}\) has \textbf{internal probabilistic knowledge} \(\km\), which it acquires during training.
Therefore, to determine the veracity of a statement \(\phi\), the model \(\mathcal{M}\) must distinguish between the following three scenarios: 
\squishlisttwo
\item $\phi$ is \textbf{True} if there is sufficient support for $\phi$ given $\km$, i.e., $\mathrm{P}(\phi \mid \mathrm{K}_{\mathcal{M}}) \geq \zeta$, where $\zeta \in (0,1]$ is a threshold.
\item $\phi$ is \textbf{False} if there is sufficient support for $\neg\phi$ given $\km$, i.e., $\mathrm{P}(\neg\phi \mid \mathrm{K}_{\mathcal{M}}) \geq \xi $, where $\xi  \in (0,1]$ is a threshold.
\item $\phi$ is \textbf{Neither-true-nor-false} if there is insufficient support for both $\phi$ and $\neg\phi$ given $\km$, i.e., $\mathrm{P}(\phi \mid \km) < \zeta$ and $\mathrm{P}(\neg\phi \mid \km) < \xi $, where $\xi  \in (0,1]$ is a threshold.
\squishend
Consequently, if \(\mathcal{M}\) has a mechanism to determine the veracity of a statement \(\phi\), then \(\mathcal M\) should encode the signal associated with the veracity in its intermediate neural representations:

\noindent \textbf{Truthfulness}: \(\mathcal{M}\) generates an activation pattern that encodes internal support for  \(\phi\) in \(\km\).

\noindent \textbf{Falsehood}: \(\mathcal{M}\) produces an activation pattern that reflects a lack of sufficient support for \(\phi\), instead indicating that the internal knowledge \(\km\) provides stronger support for \(\neg\phi\), signaling a contradiction or misalignment with known information. 

\noindent \textbf{Neither}: \(\mathcal{M}\) encodes the lack of support for \(\phi\) and \(\neg\phi\), indicating that the veracity of \(\phi\) is currently undefined.

\subsection{Flawed Assumptions When Probing Veracity in LLMs}
\label{sec:flawed_assumptions}

A veracity probe \(g_i\) is trained on labeled datasets.
These datasets consist of \(\langle h_i(\boldsymbol{x}), y \rangle\) pairs, where \(h_i(\boldsymbol{x})\) denotes an internal representation after the \(i\)th decoder and \(y \in Z\) is the ground-truth veracity label; in most cases, \(Z=\{\text{true, false}\}\). We focus on linear probes, where learned parameters define a veracity direction \(\boldsymbol{\theta}_i\):
\begin{equation}
  \label{eq:linear_reg}
  g_i\!\left(h_i(\boldsymbol{x})\right)
  = h_i(\boldsymbol{x})\,\boldsymbol{\theta}_i^\top + \beta_i.
\end{equation}
Next, we detail three flawed assumptions prevalent in the literature.

\paragraph{Flawed Assumption I: The veracity signal lies along a single axis.}
Existing veracity probes \cite{marks2023geometry, burns2022discovering, li2023inference, levinstein2024still} implicitly assume that truth and falsehood are encoded bidirectionally along a single axis in the representation space:
\begin{equation}
\label{eq:flaw-bidirect}
 P\left(\phi \mid \km\right) = 1 - P\left(\neg\phi\mid\km\right)
\end{equation}
\Cref{eq:flaw-bidirect}~implies that truthfulness and falsehood are symmetric and consistent: any layer that encodes a signal for truth must encode an equal and opposite signal for falsehood.
However, there is little empirical evidence to support this claim. In contrast,~\citet{burger2024truth} suggest that veracity exists along more than one axis.

This assumed binary framing forces all statements into a binary partition: any statement not confirmed as \textit{true} is classified as \textit{false}, and vice versa. Yet an LLM often encounters statements for which its internal probabilistic knowledge \(\km\) provides insufficient support for \textit{either} \(\phi\) or \(\neg\phi\). These include statements about entities absent from training data, contested claims, and underspecified propositions; binary probes cannot represent this third epistemic state and instead assign such statements to \textit{true} or \textit{false}.
\Cref{fig:sup_mdcp_score_distribution} illustrates the consequence: a binary, single-axis probe (\texttt{MD+CP}) cannot carve out a region for the \textit{neither} class, placing such statements inside the high-confidence intervals of \textit{true} and \textit{false} rather than in a space of their own.

\textit{Takeaway I}: A veracity probe should therefore account for the \textit{neither}\footnote{Cases where \(\mathrm{P}(\phi \mid \km) < \zeta\) and \(\mathrm{P}(\neg\phi \mid \km) < \xi \)} statements and should not presuppose that truth and falsehood share a single geometric direction.\footnote{\(\boldsymbol \theta_i^{\text{ true}} \neq - \boldsymbol \theta_i^{\text{ false}}\)}

\paragraph{Flawed Assumption II: Ground-truth labels are faithful proxies for LLM knowledge.}
In practice, to train a probe \(g_i\), we use \(\mathcal{D}_{train}\) consisting of statement-label pairs \(\langle \boldsymbol{x},\, y\rangle\). The labels \(y\) are assigned according to \textit{our} understanding of the world: these labels follow a distribution \(G_{\mathcal D}\) grounded in human knowledge. Ideally, the probe's goal is to recover the veracity signal corresponding to the LLM's internal knowledge \(\km\), which may follow a different label distribution \(G_{\mathcal M}\). 

This mismatch introduces two associated obstacles.
First, we generally do not know what the LLM has learned: the precise composition of the LLM's training data often remains unknown~\cite{longpre2024large}, and we lack straightforward methods to verify what has been incorporated into \(\km\). 
For instance, \textit{we} know that \enquote{The city of Bissau is in Congo} has a ground-truth label \(y=\textit{false}\) since we can check maps, but we do not know how \(\mathcal{M}\) labels it. The LLM may never have encountered Bissau during training, in which case the statement falls outside \(\km\) entirely.

Existing probing methods~\cite{burns2022discovering, Harding2023operationalising, li2023inference, burger2024truth, marks2023geometry, levinstein2024still} do not account for mismatches between label distributions. Instead, these probes introduce a systematic bias, in which \(g_i\) captures a signal reflecting \textit{our} labeling choices rather than the model's true internal representations. That is, they assume \(G_{\mathcal D}\stackrel{d}{=}G_{\mathcal M}\).

The second obstacle relates to the predicted labels. Even when the probe's predictions align with the correct labels, the probe's output scores are not necessarily interpretable as degrees of belief. Probes such as SVMs and mean-difference classifiers are based on decision boundaries. They output distances from a separating hyperplane rather than calibrated probabilities. 
However, \citet{herrmann2025standards} argue that veracity probes should provide not only discrete labels but also values that can be interpreted as uncertainty estimates. 
A common method is to wrap raw scores in a sigmoid function, which forces them into \([0,1]\). However, it still does not yield calibrated probabilities. 

Since we cannot interpret the score in terms of probabilities, it is unclear what these scores represent: a probe that returns a score of \(0.5\) for a statement \(\phi\) could mean that \(\mathcal M\) is genuinely uncertain under \(\km\), or it could mean the probe \(g_i\) lacks discriminative power in the representation space. Without calibration, these cases are indistinguishable.

\textit{Takeaway II:} Existing probes do not provide accurate labels or reliable confidence estimates relative to the LLM's own knowledge \(\km\). Instead, a probe should (1) account for the possibility that \(G_{\mathcal{D}} \stackrel{d}{\not=} G_{\mathcal{M}}\) and (2) produce calibrated outputs that can be interpreted as the LLM's degree of support for a given statement.

\paragraph{Flawed Assumption III: The veracity signal is localized at the last token.}
The majority of veracity probes are trained on the representation of the last token~\cite{Harding2023operationalising, marks2023geometry, li2023inference}.
For example, it is assumed that the period alone carries the entire veracity signal in \enquote{Boston is in the US.}  Thus, the implicit assumption is that any factual signal appearing \(n\) tokens before the end of the statement will be faithfully preserved until the last token. A more reasonable strategy is to probe at the exact position where the statement is actualized, e.g., immediately after \enquote{in the}, rather than relying on the LLM to \textit{move} that signal all the way to the end of the statement. 

\Cref{fig:sup_sota_sanity} provides an example: last-token and binary probes assign confident veracity labels to \textit{non-actualized} tokens in the pre-actualized span \(\boldsymbol{x}^p\) (e.g., \enquote{the}, \enquote{city}, \enquote{of}), where no truth value has yet been declared.

\textit{Takeaway III:} The position of the relevant token is not known a priori. Veracity probes, therefore, must employ a flexible mechanism to identify the optimal token positions for extracting veracity signals, rather than relying on fixed positions. 

\paragraph{Summary.} A probe that directly addresses these flawed assumptions would more accurately reflect the internal knowledge of the LLM and provide (1)~a multiclass veracity signal with distinct directions for \textit{true}, \textit{false}, and \textit{neither} statements, rather than collapsing them onto a single axis; (2)~calibrated uncertainty measures of \(\mathcal{M}\)'s probabilistic knowledge; and~(3)~a flexible mechanism for identifying informative tokens.

In~\Cref{supsec:sota_sanity}, we present a case study demonstrating that existing probing methods often misclassify tokens that do not have any associated veracity value, as well as confidently misclassify \textit{neither} statements as \textit{true} or \textit{false}. 

\section{Method: \texttt{sAwMIL}}
\label{sec:method}
To address the flawed assumptions, we propose a multiclass probe, called \texttt{sAwMIL} (short for Sparse-Aware Multiple-Instance Learning). It classifies statements into three classes: \textit{true}, \textit{false}, and \textit{neither}, and yields statistically guaranteed prediction sets.
To do so, \texttt{sAwMIL} adopts two methodological frameworks: multiple-instance learning (MIL) and conformal prediction (CP).

\subsection{Sparse-Aware Multiple-Instance Learning}
Algorithms such as logistic regression, support vector machines, and mean-difference classifiers belong to the single-instance learning (SIL) family, where each instance in the dataset has an individual label. In contrast, multiple-instance learning is a type of weakly supervised learning that operates on a set of labeled \textit{bags}~\cite{carbonneau2018multiple}. 
A bag \(B\) contains a set of related instances (e.g., embeddings of individual words in a sentence).
Each bag has an associated binary label, but the labels for individual instances within the bag often remain \textit{unknown}.
A bag with a positive label (\(y=1\)) indicates that at least one instance in the bag \(B\) belongs to the positive class. Similarly, a negative label indicates that none of the instances in the bag \(B\) are positive.

Therefore, multiple-instance learning algorithms must consider the overall structure of the bag, suppress irrelevant instances, and simultaneously identify a few informative ones.
Our proposed algorithm, \texttt{sAwMIL}, is an extension of the multiple-instance SVM proposed by~\citet{bunescu2007multiple}.

\texttt{sAwMIL} is specifically designed for sparse bags: only a few instances carry a relevant signal. It also allows us to incorporate prior knowledge about which instances are likely to carry a relevant signal.
\texttt{sAwMIL} has three training stages (we provide a detailed description of the \texttt{sAwMIL} framework in~\Cref{supsec:sawmil} and~\Cref{alg:sawmil}):

\noindent \textbf{Stage 1 (\texttt{MIL SVM})}: We start with the MIL adaptation of an SVM that operates on bags. The objective is to identify the most important instances within the positive bags, pushing all other instances and negative bags toward the opposite side of the separating hyperplane.\\
\noindent \textbf{Stage 2 (Relabeling)}: Using the fitted SVM from Stage 1, we identify the most informative instances across all positive bags. We compute the distribution of scores assigned to each instance across all positive bags and assign a positive pseudo-label to instances scoring above the \(1-\eta\)-quantile. The remaining instances are relabeled as negative. Note that \(\eta\) is the \textit{balancing hyperparameter} that encodes our prior assumption about the number of positive instances per bag. 

We further exploit the structure of our statements. Consider the following statement: \enquote{The city of Riga is in Latvia.} We can decompose it into a \textit{pre-actualized} part (\(\boldsymbol x^p = \) \enquote{The city of Riga is in}), which cannot carry any veracity signal, and an \textit{actualized} part (\(\boldsymbol x^a=\) \enquote{Latvia}), where we expect the veracity signal to appear. 
Formally, each statement decomposes into  \(\boldsymbol{x} \leftarrow [\boldsymbol{x}^p, \boldsymbol{x}^a]\), and a token is relabeled as positive if it scores above the \((1- \eta)\)-quantile \emph{and} belongs to the actualized part.

\noindent \textbf{Stage 3 (\texttt{SIL SVM})}: Finally, we train a single-instance SVM on the relabeled instances from Stage 2. Here, we disregard the original grouping into bags and use the instances and their pseudo-labels.

\paragraph{One-vs-all \texttt{sAwMIL}.} We train three one-vs-all \texttt{sAwMIL} probes that isolate distinct veracity signals. This design choice allows us to account for Flawed Assumption I: 
\squishlist
    \item \texttt{is-true} probe: separates tokens that carry a \textit{true} signal from all others.
    \item \texttt{is-false} probe: separates tokens that carry a \textit{false} signal from all others.
    \item \texttt{is-neither} probe: separates tokens that carry a \textit{neither} signal from all others.
\squishend

\paragraph{Multiclass \texttt{sAwMIL}.}
\label{sec:multiclass-sawmil}
Ideally, we want a multiclass probe that assigns probabilities to a statement being \textit{true}, \textit{false}, or \textit{neither}.
Thus, we assemble the one-vs-all \texttt{sAwMIL} probes into a multiclass one via \textit{softmax regression}. It takes the outputs of the one-vs-all probes and transforms them into multiclass probabilities: \(p_k = \exp(z_k) / \sum_j \exp(z_j)\), where $z_k = g^k_i(\boldsymbol{x}) \,\alpha_k + \beta_k$ and \(k \in\) \{\texttt{is-true}, \texttt{is-false}, \texttt{is-neither}\}. Recall that \(g_i\) is the trained probe in~\Cref{eq:probe}.

\subsection{Conformal Prediction}
\label{sec:conformal-predictions}
Raw support-vector-machine outputs (i.e., signed distances from the separating hyperplane) are not meaningful confidence measures.  We introduce a conformal prediction framework into \texttt{sAwMIL} to obtain coverage guarantees.

Conformal prediction is a post-hoc method~\cite{romano2020classification, angelopoulos2020uncertainty} that transforms raw scores into prediction sets with  \textit{guaranteed coverage}.
That is, given a new instance \(\boldsymbol x\) and a predicted score \(z\), CP outputs a prediction set \(\mathcal{C}({\boldsymbol{x}}) \subseteq Z\) such that the ground-truth label is contained with probability at least \(1 - \alpha\): \( P\bigl(y \in \mathcal{C}(\boldsymbol{x})\bigr) \geq 1 - \alpha\).
Prediction sets that contain multiple labels indicate uncertainty, singleton sets indicate confident predictions, and empty sets indicate that the instance \(\boldsymbol x\) is too unusual to be classified as any of the classes.

For a detailed description of the nonconformity scores, refer to~\citet{johansson2017model} and~\Cref{supsec:nonconformity}.

\section{Experiments}
\begin{table*}[th]
\centering
\small{
\begin{tabular}{m{0.10\linewidth}m{0.072\linewidth}m{0.072\linewidth}m{0.072\linewidth}m{0.55\linewidth}}
\toprule
\textbf{Dataset}     & \textbf{True} & \textbf{False}  & \textbf{Neither} & \textbf{Examples} \\ \toprule
City\newline Locations      & A: 1392\newline N: 1376    & A: 1358\newline N: 1374    & A: 876\newline N: 876 & (True) The city of Mâcon is located in France.\newline (False) The city of Dhar\={a}n is located in Ecuador.\newline (Neither) The city of Staakess is located in Marbate.          \\ \midrule
Medical\newline Indications       & A: 1522\newline N: 1439   & A: 1419 \newline N: 1523   & A: 478\newline N: 522  & (True) PR-104 is indicated for the treatment of tumors.\newline (False) Zolpidem is indicated for the treatment of angina.\newline (Neither) Alostat is indicated for the treatment of candigemia. \\ \midrule
Word\newline Definitions & A: 1269\newline N: 1251  & A: 1223\newline N: 1257  & A: 1747 \newline N: 1753   & (True) Corsage is a synonym of a nosegay.\newline  (False) Towner is not a type of a resident.\newline (Neither) Kharter is not a synonym of a greging.\\   \bottomrule
\end{tabular}
}
\caption{Composition of datasets used in this work showing the number of \textit{true}, \textit{false}, and \textit{neither} statements per dataset. \textbf{A} stands for the number of affirmative statements, and \textbf{N} stands for the number of negated statements. The last column shows example statements with ground-truth labels.
}
\label{tab:data_set}
\end{table*}
This section outlines our experimental setup, including datasets, our evaluation procedure, and the set of large language models.

\subsection{Data}
\label{sec:exp_data}

We introduce three new datasets comprising factually \textit{true}, factually \textit{false}, and \textit{neither} statements. 
While several benchmarks for veracity and factuality evaluation exist, prior work has shown that some of these may be incorporated into the pretraining or fine-tuning stages of LLMs~\cite{fang-etal-2025-lastingbench}.
In contrast, our goal is to minimize the risk of data contamination while also maintaining greater control over data provenance and quality. 
Hence, we assemble new datasets that involve statements related to specific themes (see~\Cref{tab:data_set} for examples):
\squishlist
\item \textbf{City Locations} dataset contains statements about cities and their corresponding countries extracted from the \texttt{GeoNames} geographical knowledge database~\cite{geonames}. 
\item \textbf{Medical Indications} dataset consists of statements about medications and their corresponding indications from the \texttt{DrugBank 5.1.12} knowledge base~\cite{wishart2018drugbank}.
\item \textbf{Word Definitions} dataset is based on the \citet{wordsapi} dictionary; these statements involve words and their synonyms or relations. 
\squishend

We deliberately restrict our study to \emph{simple relational statements}.
This is a construct-validity choice: to test whether a probe recovers a \textit{veracity} signal, the stimuli must have an unambiguous, verifiable truth value that does not depend on context or compositional structure.
    
Every dataset consists of negated statements such as \enquote{The city of Riga \textbf{is not} located in Estonia.}\footnote{This statement is a factually true and negated statement.} and affirmative ones like \enquote{Menadione \textbf{is} indicated for the treatment of coughs.}\footnote{This statement is a factually false and affirmative statement.}

\paragraph{\textit{Neither} statements.} If a statement \(\phi\) is absent from the LLM's internal probabilistic knowledge \(\km\), then \(\phi\) is \textit{neither}, since the truth value cannot be determined at the present moment (due to the lack of information).

It is non-trivial to determine which statements are absent from \(\km\), because we generally do not have access to the training datasets used to train LLMs.  
Instead, we create \textit{neither} statements with \textit{synthetic entities}, which occur neither in the real world nor in works of fiction.  
Since these entities are specifically generated for our experiments, it is highly unlikely that an LLM has learned anything about them during training. Thus, we can use them as \textit{substitutes} for content that LLMs could not have learned: from the point of view of an LLM, these should be considered neither \textit{true} nor \textit{false}.

Refer to~\Cref{sec:sup_data_unknowns} for additional details on data provenance and generation of \textit{neither} statements.

\subsection{Evaluation}
\label{sec:lang_models}

\begin{figure*}[t]
\centering
  \includegraphics[width=1\linewidth]{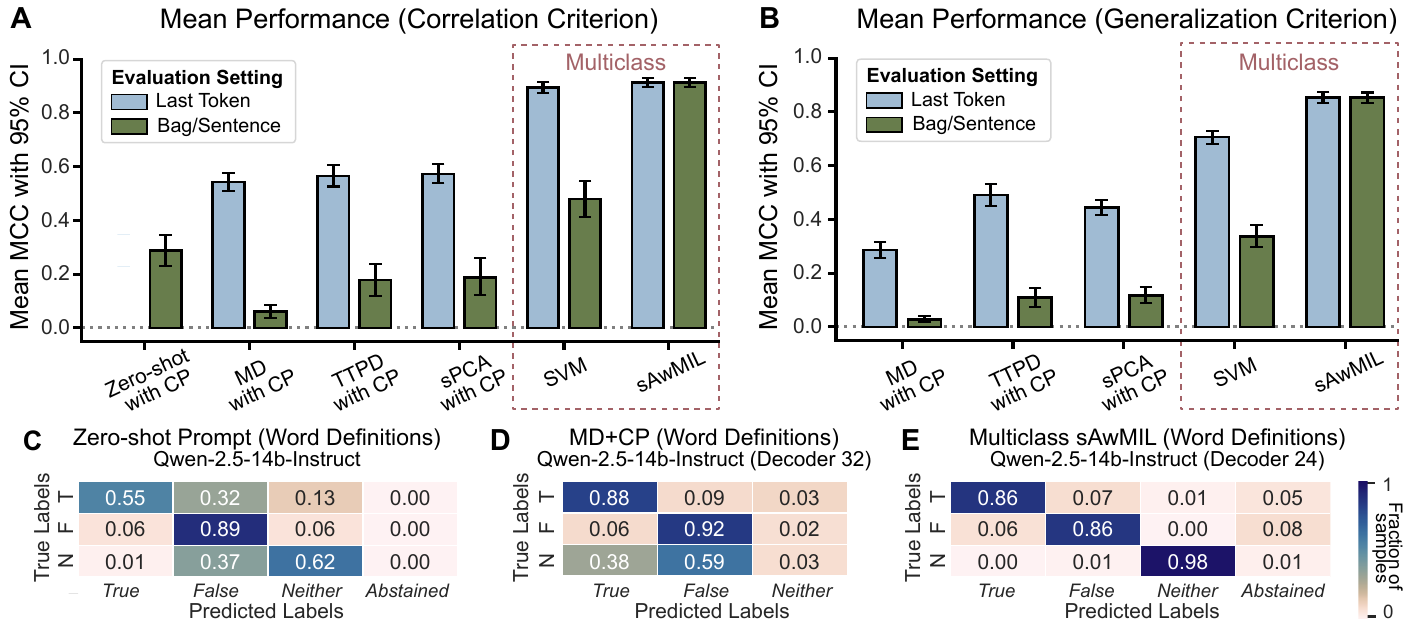}
\caption{Mean performance of probing methods aggregated across 16 models and three datasets (with \(95\%\) confidence intervals), along with examples of confusion matrices. Each probe is evaluated under two settings: using only the representation of the last token and using predictions aggregated across the entire bag/sentence. For each probe, we aggregate the statistics using the best-performing layers.
(\textbf{A}) \textit{Correlation criterion}. Mean probe performance on the test split of the dataset on which each probe was trained.
(\textbf{B}) \textit{Generalization criterion}. Mean probe performance on datasets that were not used for training.
(\textbf{C}) Confusion matrix for zero-shot prompting, where the LLM overpredicts the \textit{false} class. (\textbf{D}) Confusion matrix for \texttt{MD+CP} evaluated on the last token representation, where the probe incorrectly predicts \textit{neither} statements as \textit{true} or \textit{false}. (\textbf{E}) Confusion matrix for multiclass \texttt{sAwMIL} evaluated on the full bag.
Binary probes trained on the last token representation (\texttt{MD+CP}, \texttt{TTPD+CP}, and \texttt{sPCA+CP}) exhibit both lower performance and poorer generalization. The multiclass \texttt{SVM} has comparable performance on the last token. However, performance degrades when evaluated on the full-bag setting; it also achieves lower generalization than \texttt{sAwMIL}.}
  \label{fig:bar_performance}
\end{figure*}

\paragraph{Language Models.} We evaluate 16 open-source LLMs (ranging from approximately 3 to 14 billion parameters) across four families: \texttt{Gemma}/\texttt{Gemma-2}, \texttt{Llama-3} (\texttt{v3.1} and \texttt{v3.2}), \texttt{Mistral-v0.3}, and \texttt{Qwen-2.5}. These models run on consumer-grade hardware and are publicly available through \href{https://huggingface.co/}{Hugging~Face}~\cite{wolf-etal-2020-transformers}. We provide a detailed overview of these models in~\Cref{supsec:llms}. 

\paragraph{Evaluation Metric.} We evaluate probes based on their ability to correctly classify statements into three classes: \textit{true}, \textit{false}, and \textit{neither}.
We use Matthews correlation coefficient (MCC) to summarize the performance of each probe (on the test sets); refer to~\Cref{supsec:evaluation} for the definition of MCC.

\paragraph{Probing Methods.} We compare performance across \textit{six} probing methods, grouped into three categories. In \textit{zero-shot prompting}, we insert each statement into a prompt formulated as a multiple-choice question. 
\Cref{sec:sup-zero-shot} provides additional results for the zero-shot evaluation.

We also evaluate three \textit{binary representation-based probes} trained on the representations of the last token: the mean-difference (\texttt{MD}) probe introduced by \citet{marks2023geometry}, the \texttt{TTPD} probe (short for Training of Truth and Polarity Direction) introduced by \citet{burger2024truth}, and a supervised PCA classifier (\texttt{sPCA}). We refer the reader to~\Cref{alg:mean_diff,alg:ttpd,alg:spca} for details. 
Since \texttt{MD},~\texttt{TTPD}, and~\texttt{sPCA} probes are trained to separate only \textit{true} and \textit{false} statements, we augment each with conformal prediction. 
Samples whose conformal prediction set does not contain exactly one of \textit{true}/\textit{false} are labeled \textit{neither}.
We refer to these methods as \texttt{MD+CP}, \texttt{TTPD+CP}, and \texttt{sPCA+CP}, respectively.

Finally, we include two \textit{multiclass probes}:  a multiclass \texttt{SVM} trained on the representation of the last token,\footnote{Multiclass \texttt{SVM} is not equipped with conformal prediction.} and the multiclass \texttt{sAwMIL} probe, which operates on the whole sentence---all token representations within a statement---and uses a conformal prediction framework.

\paragraph{Validity Criteria.} We compare all probes under five criteria: (1) \textit{Correlation}, (2) \textit{Generalization}, (3) \textit{Manipulation}, (4) \textit{Locality}, and (5) \textit{Selectivity}. See~\Cref{supsec:evaluation,supsec:interventions} for details on the validity criteria. Here we focus on the first three criteria.

The \textbf{correlation criterion}~\cite{Harding2023operationalising, herrmann2025standards} assesses how well a probe  \(g_i\), trained on \(\mathcal{D}_{train}\), performs on the corresponding \(\mathcal{D}_{test}\), assuming that both are drawn from the same distribution. 
For example, a probe trained on statements about city locations should accurately classify other statements from the same domain.
The \textbf{generalization criterion}~\cite{burger2024truth} evaluates whether a probe \(g_i\), trained on \(\mathcal{D}_{train}\), successfully generalizes to datasets \(\mathcal{D}'_{test}\) containing statements from a different domain. 

The \textbf{manipulation} criterion tests whether the identified direction \(\boldsymbol{\theta}_i\) \textit{causally} affects \(\mathcal{M}\)'s outputs.
In addition, the \textbf{locality} criterion verifies whether these manipulations predominantly affect the veracity-relevant tokens rather than shifting the distribution of unrelated tokens~\cite{Harding2023operationalising}.

\paragraph{Contrastive Token Intervention (CTI).} We assess the manipulation and locality criteria by combining activation steering~\cite{li2023inference, todd2024function} with a difference-in-differences (DiD) design.
Recall that we have a tokenized statement \(\boldsymbol{x} = [\boldsymbol{x}^p, \boldsymbol{x}^a]\) with a pre-actualized part \(\boldsymbol{x}^p\) and an actualized part \(\boldsymbol{x}^a\).
Further, let \(\boldsymbol{\tau}_c\) denote the correct actualized sequence (\(\boldsymbol{x}^a = \boldsymbol{\tau}_c\)) and \(\boldsymbol{\tau}_r\) a length-matched random control (\(|\boldsymbol{\tau}_c| = |\boldsymbol{\tau}_r|\)).
At layer \(i\), we translate the last-token representation of \(\boldsymbol{x}^p\) along the veracity direction \(\boldsymbol{\theta}_i\):
\begin{equation}
\label{eq:intervention-paper}
    \tilde{h}^{\kappa}_i(\boldsymbol{x}^p) \leftarrow 
    h_i(\boldsymbol{x}^p) + \kappa \cdot \alpha \cdot 
    \boldsymbol{\theta}_i,
\end{equation}
where \(\kappa \in \{-1, +1\}\) is the sign of the intervention and \(\alpha\) its dose (\Cref{supsec:interventions}).
For each target sequence \(\boldsymbol{\tau} \in \{\boldsymbol{\tau}_c, \boldsymbol{\tau}_r\}\) and \(\kappa\), we record the resulting change in the output log-probability \(\Delta_{\boldsymbol{\tau}}^{\kappa}\) relative to the unperturbed model.
The positive and negative translations, each for the correct and control tokens, result in four measurements (and form the outcome of a \(2\times2\) model):
\begin{equation}
\begin{aligned}
    \Delta_{\boldsymbol{\tau}}^{\kappa} = 
    &\beta_0 
    + \beta_1 \cdot \mathds{1}[\boldsymbol{\tau}=\boldsymbol{\tau}_c] 
    + \beta_2 \cdot \mathds{1}[\kappa=+] \\
    &+ \beta_3 \cdot \mathds{1}[\boldsymbol{\tau}=\boldsymbol{\tau}_c] \cdot \mathds{1}[\kappa=+] 
    + \varepsilon
\end{aligned}
\label{eq:did}
\end{equation}
The interaction term \(\beta_3\) is the veracity-specific effect: the additional change in log-probability on the correct tokens \(\boldsymbol{\tau}_c\) after we account for the shift \(\beta_2\) that the intervention induces on all tokens.
A non-significant \(\beta_3\) means the direction \(\boldsymbol{\theta}_i\) has the same effect on correct \(\boldsymbol{\tau}_c\) and control tokens \(\boldsymbol{\tau}_r\); a significant \(\beta_3\) meets both criteria: \(\boldsymbol{\theta}_i\) causally affects \(\mathcal{M}\)'s outputs (manipulation) and does so specifically for veracity-relevant tokens (locality).
Since the raw \(\beta_3\) is scale-dependent across \(\langle\)LLM, dataset\(\rangle\) configurations, we summarize results for each probe via the \textit{normalized interaction effect} (NIE):  \(|\beta_3/\hat{\sigma}|\), where \(\hat{\sigma}\) is the residual standard deviation of the fit.
Thus, NIE is a scale-independent measure of the veracity-specific shift.

We estimate the pooled NIE across the 48 \(\langle\)LLM, dataset\(\rangle\) configurations using a DerSimonian--Laird random-effects estimator~\cite{kelley2023evolution}, which accounts for between-configuration heterogeneity.
A complementary significance-based \textit{success rate}  is reported in~\Cref{supsec:interventions}.
We apply CTI to the four probes (\texttt{MD}, \texttt{TTPD}, and the \texttt{is-true} directions of \texttt{SVM} and \texttt{sAwMIL}); \Cref{supsec:interventions} provides full details.

\section{Results}
\label{sec:results}

We first look at the performance of probes under the \textit{correlation} and \textit{generalization} criteria. 
~\Cref{fig:bar_performance} shows the mean performance aggregated across all 16 LLMs and three datasets. For representation-based probes, we report results for two settings: (1) using only the last token's representation, and (2) using all token representations (the full bag).\footnote{For single-instance probes, when the full bag is provided, the predicted label corresponds to the class with the highest score among all tokens in the statement.}

\textbf{Binary representation-based probes}.
The \texttt{MD+CP}, \texttt{TTPD+CP}, and  \texttt{sPCA+CP} probes achieve moderate MCC when evaluated on the last-token representation (\Cref{fig:bar_performance}.A). 
However, their performance drops substantially when evaluated on the full-bag setting, suggesting sensitivity to non-actualized tokens or noise.
Under the generalization criterion (\Cref{fig:bar_performance}.B), these probes exhibit limited transferability and degraded performance on the full-bag setting. 
A common failure, illustrated in~\Cref{fig:bar_performance}.D and detailed in~Tables~\ref{sup_tab:cm_mdcp_instance}--\ref{sup_tab:cm_spca_bag}, is that these probes frequently and confidently misclassify \textit{neither} statements as \textit{true} or \textit{false}.
Thus, while effective under controlled conditions, these binary probes appear to identify proxy directions that partially reflect veracity but are confounded by spurious correlations.

\textbf{Zero-shot prompting}. Unlike representation-based probes, prompting does not require information about where a factual statement begins or ends. This makes it more robust when working with unstructured text.
However, zero-shot prompting shows a strong preference for a single class: models tend to overpredict one label. 
For example, in~\Cref{fig:bar_performance}.C, Qwen-2.5-14B-Instruct overpredicts the false label under prompting. We see similar skews in the confusion matrices of other models with zero-shot prompting (\Cref{sup_tab:cm_zero_shot}). 

\textbf{Multiclass representation-based probes}. 
Both multiclass probes, \texttt{SVM} and \texttt{sAwMIL}, achieve substantially higher performance (\Cref{fig:bar_performance}.A). 
However, the \texttt{SVM} probe, like its binary counterparts, exhibits a performance drop under the bag setting, indicating reliance on non-generalizable cues.
In contrast, the \texttt{sAwMIL} probe maintains consistent performance across both settings and yields superior generalization performance (\Cref{fig:bar_performance}.B). 

\begin{figure}[ht]
\centering
  \includegraphics[width=0.9\linewidth]{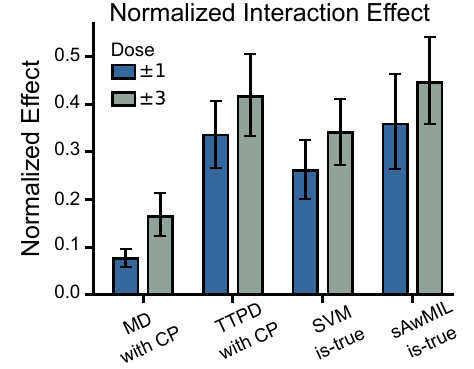}
\caption{Manipulation \& locality evaluation pooled over all 48 \(\langle\)LLM, dataset\(\rangle\) configurations for each probe: pooled normalized interaction effect (NIE) adjusted via the DerSimonian–Laird random-effects estimator~\cite{kelley2023evolution}. The error bars are 95\% confidence intervals.
The normalized effect size quantifies the magnitude of the effect.
\texttt{MD} shows the weakest causal effect, followed by the single-instance \texttt{SVM};
\texttt{TTPD} and \texttt{sAwMIL}'s \texttt{is-true} achieve similar effect sizes.
These results are also presented in~\Cref{tab:did-pooled}.}
\label{fig:nie-main}
\end{figure}
\paragraph{Manipulation \& locality.} Finally, we test whether the veracity directions causally affect \(\mathcal{M}\)'s outputs.
At the strongest dose, steering along the \texttt{is-true} direction from \texttt{sAwMIL} produces the largest normalized interaction effect (NIE \(0.439\); \Cref{fig:nie-main}), while \texttt{MD} achieves the weakest (\(0.155\)); NIE increases with dose (\Cref{tab:did-pooled}), consistent with a valid directional effect.
Notably, \texttt{TTPD} nearly matches \texttt{sAwMIL} (NIE \(0.411\)).
That is, it too locates a causally relevant direction, but \texttt{TTPD}'s direction does not satisfy the correlation and generalization criteria, suggesting it captures a causally relevant but non-generalizable (noisy) signal.
Together, these findings indicate that \texttt{sAwMIL} is the only probe whose veracity directions are at once robust, transferable, and causally relevant.

\subsection*{Veracity Directions}
To show that the truthfulness and falsehood directions are not perfect opposites, we examine the \texttt{is-true} and \texttt{is-false} directions identified by both the multiclass \texttt{SVM} probe and \texttt{sAwMIL} probe.

\begin{figure}[t]
\centering
  \includegraphics[width=\linewidth]{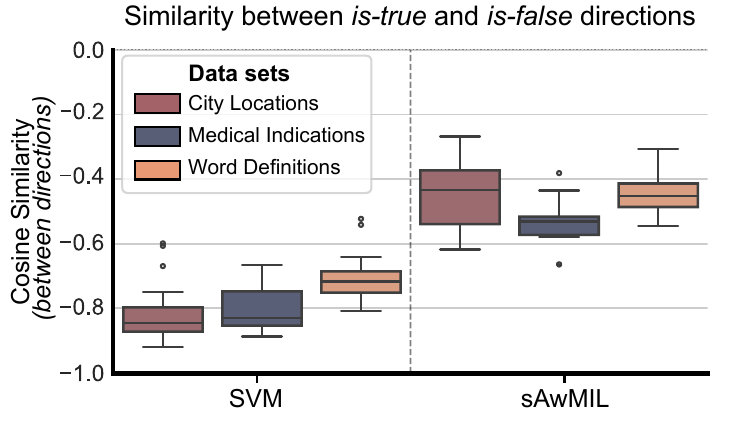}
\caption{Cosine similarity between the \texttt{is-true} and \texttt{is-false} directions across datasets and probes, using values extracted from the best-performing layer and averaged across 16 LLMs. If the directions were perfectly opposite, their cosine similarity would be \(-1\), indicating a single bidirectional axis of truth and falsehood. Instead, all probes exhibit less opposition, with the better-performing and more generalizable \texttt{sAwMIL} probe yielding higher (less negative) cosine similarity.}
\label{fig:dir_similarity_simple}
\end{figure}

In~\Cref{fig:dir_similarity_simple}, we see how different these directions are by computing their cosine similarity.
If the two were perfectly opposite, we would expect a cosine similarity of approximately \(-1\).
However, this is not the case. The better-performing and more generalizable \texttt{sAwMIL} probe yields directions that are \textit{less opposed}, suggesting that LLMs encode \textit{true} and \textit{false} as related rather than polar concepts.

Similarly, when forming a matrix \(\Theta = [\boldsymbol \theta^{\text{true}}, \boldsymbol \theta^{\text{false}}]\), we find that its rank is \(2\) for both \texttt{SVM} and \texttt{sAwMIL}, while the effective rank is \(1.73\ \pm 0.012\) for \texttt{SVM} and \(1.93 \pm 0.004\) for \texttt{sAwMIL}.\footnote{The number after \(\pm\) is standard error.}
If the two directions were linear combinations of one another (i.e., lying on a single axis), the rank would be \(1\) and the effective rank would be closer to \(1\).

These findings indicate that the \texttt{is-true} and \texttt{is-false} directions are not strict opposites but span a low-dimensional subspace that captures shared yet distinct representational components.

\section{Conclusion}

In this work, we critically examine popular methods for probing the veracity of large language models (LLMs) and show that these methods fail to learn reliable and transferable veracity patterns. We argue that probing methods rest on flawed assumptions about how veracity is encoded in LLMs' internal representations.

To address the flaws, we introduce \texttt{sAwMIL}, a multiclass linear probe that combines multiple-instance learning with conformal prediction.
Unlike prior methods, \texttt{sAwMIL} models veracity along three axes: \textit{true}, \textit{false}, and \textit{neither}. Across sixteen models and three datasets, \texttt{sAwMIL} outperforms existing probes and shows that truthfulness and falsehood are not represented as simple opposites within LLMs, but as directions spanning a subspace.
These findings suggest that veracity-related signals in language models span multiple dimensions rather than a single axis of truthfulness.

Beyond diagnosis, \texttt{sAwMIL} is a practical tool for knowledge auditing and editing. Its calibrated three-way signal indicates whether a given statement is encoded in \(\mathcal{M}\) as \textit{true}, \textit{false}, or \textit{neither}, and our intervention experiments indicate that the recovered directions are causally related to \(\mathcal{M}\)'s outputs.

\section*{Limitations}

\paragraph{Data and Statements.} This study focuses on a specific subset of factual statements, namely those that involve the relation between two entities: city-country, medicine-disease, and noun-noun pairs. 
As such, it remains unclear how the \texttt{sAwMIL} probe behaves in cases involving multiple relational facts, or in statements describing relations among more than two entities.
All statements in our datasets are written in English, and negated statements are limited to the simple \enquote{not} construction.

Furthermore, the \textit{neither} statements used in this study are synthetically generated to simulate cases where LLMs lack knowledge.
While these proxies enable controlled evaluation, they should be further validated to ensure they accurately capture how models represent unknown, ambiguous, or context-dependent facts.
Future work could explore more natural examples from temporal holdout datasets or corpora that reflect human uncertainty. 

\paragraph{Evaluation.}
While our experiments cover 16 LLMs spanning multiple families, our results are limited to models ranging from approximately \(3\) billion to \(14\) billion parameters. As such, results might not generalize to much smaller or larger LLMs. Moreover, our study considers only publicly available LLMs and excludes proprietary (closed-source) models. That is, we cannot guarantee that our findings generalize beyond the models that we use. 

Our evaluation covers only single-turn scenarios, and we do not explore how probes behave in multi-turn interactions or under varying contexts and instructions. Finally, we consider probes that can only identify linear interactions. It remains an open question whether non-linear probes can achieve better generalization or identify more complex veracity-related patterns.

\section*{Ethical Considerations}
In this study, we introduce a probing method, \texttt{sAwMIL}, designed to identify veracity-related signals encoded within large language models (LLMs).
The primary applications of our approach are auditing and enhanced interpretability of LLMs.
Our framework offers a clearer overview of the information captured by LLMs. Additionally, our method can be used to mitigate harmful or incorrect outputs.
However, the method may also be exploited by malicious actors to manipulate LLM behavior.

For instance, activation steering along learned veracity directions could bias model outputs toward incorrect or misleading information.
Knowledge of where truthfulness signals arise could be exploited to suppress or modify information (e.g., by removing specific knowledge, censoring information, or introducing targeted misinformation).

Importantly, we do not assess the robustness of \texttt{sAwMIL} against adversarial actors and cannot guarantee that the identified veracity directions are resistant to manipulation or obfuscation. Nevertheless, we release our code and data to promote open, reproducible research and encourage future investigations into the adversarial robustness of \texttt{sAwMIL}.

\section*{Use of AI}
We used Claude and GitHub Copilot for code annotations and formatting assistance. Grammarly was used for assistance with grammar, spelling, and style. Claude and GPT were used for a final proofread of the manuscript.

\clearpage
\appendix
\setcounter{figure}{0}
\setcounter{table}{0}
\setcounter{equation}{0}
\renewcommand{\thefigure}{A\arabic{figure}}
\renewcommand{\thetable}{A\arabic{table}}
\renewcommand{\theequation}{A\arabic{equation}}

\onecolumn
\section*{Appendix}
\DoToC

\clearpage
\onecolumn
\section{Notations and Abbreviations}
\label{supsec:notations}

\begin{table*}[ht]
  \centering
\begin{adjustbox}{width=0.9\textwidth}
  \begin{tabular}{@{}lll@{}}
    \toprule
    \textbf{Symbol} & \textbf{Description} & \textbf{Shape / Notes} \\
    \midrule
    \(\mathcal{M}\) & Large language model (LLM) & \\
    \(\km\) & Internal probabilistic knowledge of the model \(\mathcal{M}\) & \\
    \(\mathcal{V}\) & Vocabulary of the model \(\mathcal{M}\), consists of tokens & \([\tau_1, \dots, \tau_{|\mathcal V|}] \in \mathcal{V} \) \\
    \(P_{\mathcal{M}}\left(\tau \mid \boldsymbol x\right )\) & Output of \(\mathcal M\): a conditional probability distribution on tokens  & \\
    \midrule
    \multicolumn{3}{c}{\textbf{Inputs and Datasets}} \\
    \midrule
    \(\mathcal{D}\) & Dataset of statements & \(\mathcal{D} = \{\mathcal{D}_{\text{train}},\, \mathcal{D}_{\text{test}},\, \mathcal{D}_{\text{cal}}\}\)  \\
     \(\boldsymbol{x}\)
      & Input token sequence, e.g., \enquote{The city of Riga is in Latvia.}
      & $L = |\boldsymbol{x}|$ \\
    \(\boldsymbol{x}^p\) & Pre-actualized part of a statement, e.g., \enquote{The city of Riga is in} & \\
    \(\boldsymbol{x}^a\) & Actualized part of a statement, e.g., \enquote{Latvia.} & \\
    \(B\) & Bag of token representations (contains related instances) &  \\
    \(y\) & Veracity label assigned to \(\boldsymbol x\) & \( y \in Z\) \\
    \(\phi\) & A statement evaluated for truthfulness/veracity &  \\
    \(\mathcal{T}_S\) & Transition matrix for n-gram generation & See~\Cref{eq:ngram_markov} \\
\midrule
\multicolumn{3}{c}{\textbf{Internal Representations}} \\
\midrule
    \(d\) & Size of the hidden representation (of a decoder) & \\
    $h_i(\boldsymbol{x})$ & Activations after the $i$th decoder& $\mathbb{R}^{L\times d}$ \\
    $h_i(\boldsymbol{x})_{[j]}$ & Activation of the token at index \(j\) after $i$th decoder  & 
    $\mathbb{R}^{1\times d}$ and \(j \in \{1 \dots L\}\)\\
    $h_i(\boldsymbol{x})_{[a{:}b]}$ & Activations from \(a\)th to \(b\)th tokens after \(i\)th decoder & $\mathbb{R}^{(b-a+1)\times d}$\\
\midrule
\multicolumn{3}{c}{\textbf{Veracity, Probes and Distributions}} \\
\midrule
        \(Z\) & Set of veracity labels, e.g., \{\textit{true}, \textit{false}, \textit{neither}\} & \\
        \(G_{\mathcal D}\left(z \mid \boldsymbol x\right)\) & Distribution of veracity labels \(z \in Z\) in a dataset \(\mathcal D\) & \\
    \(G_{\mathcal M}\left(z\mid \boldsymbol x\right)\) & Distribution of veracity labels \(z \in Z\) in the model \(\mathcal M\) & \\
    $g_i$ & Veracity probe trained on representations of the $i$th decoder & \(g_i: h_i(\boldsymbol x) \mapsto G_{\mathcal M}\) \\
    \(\boldsymbol \theta_i\) & Linear veracity direction extracted from the probe \(g_i\)& \( \mathbb{R}^{1 \times d}\)\\
    \(\eta\) & Balancing hyperparameter for  \texttt{sAwMIL} & \(\eta \in (0,1]\) \\
        \(\boldsymbol{m}\) & \texttt{sAwMIL}'s intra-bag labels (i.e., per-token labels in each \(\boldsymbol{x}\))  & \(\boldsymbol{m} \in \{0,1\}^{L}\), \(L = |\boldsymbol{x}|\) \\
    \(\mathcal{C}(\boldsymbol{x})\) & Prediction set (related to conformal prediction) \(\mathcal{C}(\boldsymbol{x}) \subseteq Z\) \\
    \(P\left(y\in\mathcal{C}\left(\boldsymbol{x}\right)\right)\) & Probability that label \(y\) is part of the prediction set & \\
    \midrule
    \multicolumn{3}{c}{\textbf{Interventions via Activation Steering} (\Cref{supsec:interventions})}\\
    \midrule
    \(\kappa\) & Sign for the intervention direction  & \(\kappa \in \{-1, +1\}\) \\
    \(\alpha\) & Intervention dose & \(\alpha = d' \cdot \sigma,\; d' \in \{1,3\}\) \\
    \(d'\) & Dose scaler & \(d' \in \{1,3\}\) \\
    \(\sigma\) & Standard deviation of the scores provided by \(g_i\) (based on \(\mathcal{D}_{cal}\))&  See~\Cref{eq:dosage} \\
    \(\tilde{h}^{\kappa}_i(\boldsymbol{x}^p)\) & Steered hidden representation at layer \(i\) (along \(\alpha\cdot\kappa\cdot\boldsymbol \theta_i)\) & \\
    \(\boldsymbol{\tau}_c\) & Actualized sequence (containing correct tokens) & \(\boldsymbol{x}^a = \boldsymbol{\tau}_c\) \\
    \(\boldsymbol{\tau}_r\) & Actualized sequence (contains random control tokens) & \(|\boldsymbol{\tau}_r| = |\boldsymbol{\tau}_c|\) \\
    \(\Delta^{\kappa}_{\boldsymbol{\tau}}\) & Log-probability shift after intervention (along \(\alpha\cdot\kappa\cdot\boldsymbol \theta_i)\) & See \Cref{eq:delta} \\
    \(\beta_0, \beta_1, \beta_2, \beta_3\) & DiD regression coefficients & See \Cref{eq:sup_did} \\
    \(\hat{\sigma}\) & Residual standard deviation of the DiD fit & See \Cref{eq:sup_normalized_interaction} \\
    \(\tau^2\) & Between-configuration heterogeneity (DerSimonian–Laird estimator) & \\
    \bottomrule
  \end{tabular}
  \end{adjustbox}
  \caption{Notations used throughout the paper. 
Symbols are grouped by category: model definitions, inputs and datasets, internal representations, veracity distributions, and intervention-related symbols.}
    \label{tab:notation}
\end{table*}

\begin{table*}[!t]
\begin{adjustbox}{width=1\textwidth}
  \begin{tabular}{@{}lll@{}}
    \toprule
    \textbf{Abbreviation} & \textbf{Full Form}                          & \textbf{Description} \\
    \midrule
    LLM & Large Language Model & \\
    SIL       & Single-Instance Learning      & Probes trained on single instances \\
    MIL & Multiple-Instance Learning & Probes trained on bags (collections of instances) \\ 
    SVM & Support Vector Machine & Type of a classifier \\ 
    CP        & Conformal Prediction           & Uncertainty calibration method \\
    \texttt{MD+CP}    & Mean-Difference with conformal prediction  & It classifies \textit{neither} via abstention \\
    \texttt{TTPD+CP}    & Training of Truth and Polarity Direction (probe) with CP  & It classifies \textit{neither} via abstention \\
    \texttt{sPCA+CP}    & Supervised PCA (probe) with conformal prediction  & It classifies \textit{neither} via abstention \\
    \texttt{sAwMIL}    & Sparse-Aware MIL probe & Multiclass probe handling unknowns \\
    MCC       & Matthews Correlation Coefficient & Multiclass performance measure in~\cref{eq:sup_mcc} \\
    NER & Named Entity Recognition & See~\Cref{supsec:datasets} \\
    DiD & Difference-in-differences design & See~\Cref{supsec:interventions} \\
    CTI & Contrastive Token Intervention & See~\Cref{supsec:interventions}  \\
    NIE & Normalized Interaction Effect (related to CTI) & See~\Cref{supsec:interventions} \\
    SR & Success rate (related to CTI) & See~\Cref{supsec:interventions} \\
    \bottomrule
  \end{tabular}
  \end{adjustbox}
   \caption{Abbreviations and naming conventions used throughout this paper.}
\label{tab:abbreviation}
\end{table*}

\clearpage
\twocolumn
\section{Datasets}
\label{supsec:datasets}

We introduce three new datasets: \textit{City Locations}, \textit{Medical Indications}, and \textit{Word Definitions}. 
Each dataset consists of statements that are factually \textit{true}, factually \textit{false}, or \textit{neither}. 
These datasets contain both affirmative and negated statements. 
The datasets are publicly available on Hugging Face at \href{https://huggingface.co/datasets/carlomarxx/trilemma-of-truth}{\texttt{carlomarxx/trilemma-of-truth}}   (DOI:~\href{https://doi.org/10.57967/hf/5900}{\texttt{10.57967/hf/5900}}).

An example of a false negated statement is \enquote{Guaifenesin is \textbf{not} indicated for the treatment of coughs}, and an example of the true affirmative statement is \enquote{Shouter is a type of a communicator.} 

\paragraph{Data Splits.} We split each dataset into train, calibration, and test sets using \textit{approximately} 55/20/25 percentage splits (see~\Cref{sup_tab:data_split}).
We ensure that each \(\langle\)subject, object\(\rangle\) pair is exclusive to one split, so a statement and its negation always end up in the same split.

\begin{table*}[ht]
\centering
\begin{tabular}{@{}rllll@{}}
\toprule
\multicolumn{1}{c}{\textbf{Dataset}} & \multicolumn{1}{c}{\textbf{Train}} & \multicolumn{1}{c}{\textbf{Calibration}} & \multicolumn{1}{c}{\textbf{Test}} & \multicolumn{1}{c}{\textbf{Total}} \\ \midrule
City Locations & 3999 (.55) & 1398 (.19) & 1855 (.26) & 7252 (1.00) \\
Medical Indications & 3849 (.56) & 1327 (.19) & 1727 (.25) & 6903 (1.00) \\
Word Definitions & 4717 (.55) & 1628 (.19) & 2155 (.25) & 8500 (1.00) \\ \bottomrule
\end{tabular}

\caption{\textbf{Dataset splits.} 
The number of statements per split. In the parentheses, we specify the fraction of the total number of statements. }
\label{sup_tab:data_split}
\end{table*}

\subsection{\textit{Neither} Statements}
Since we do not have access to LLM training datasets, we cannot validate whether LLMs retain information about specific facts or entities.
To overcome this issue, we create \textit{neither} statements that contain \textit{synthetic entities}: entities that do not exist in the real world or fictional works. The \textit{neither} statements are the ones whose value cannot be determined at present (e.g., due to lack of information).

\subsubsection*{Generation of \textit{Neither} Statements}
\label{sec:sup_data_unknowns}
We use synthetic names to generate \textit{neither} statements.
For example, \enquote{The city of \textit{Staakess} is located in \textit{Soldovadago}} mentions a town and a country that do not exist.
From the point of view of an LLM, these statements should be considered \textit{neither-true-nor-false}, as LLMs could not have captured anything about these in \(\km\).

To generate the \textit{neither} statements, we use the Markov Chain technique~\cite{jm3}.
Given a set of existing words \([w_1, w_2 \dots, w_n] \in S\), we break each word \(w_i \) into \(n\)-grams.  
For instance, we break \enquote{ability} into the following 2-grams: \texttt{[start]a ab bi il li it ty y[end]}.
We then compute a transition matrix \(\mathcal{T}_S\), which provides the probability of transitioning from the \(n\)-gram \(i\) to \(n\)-gram \(j\):
\begin{equation}
\label{eq:ngram_markov}
    \mathcal{T}_S\left(j \mid i\right)=\frac{\operatorname{count}\left(i \rightarrow j\right)}{\sum_x \operatorname{count}\left(i \rightarrow x\right)}
\end{equation}   
We use \(\mathcal{T}_S\) to sample new synthetic words that follow the \(n\)-gram distribution of words in \(S\).
We generate the following types of synthetic entities: city names, country names, drug/medicine names, disease/indication names, and English nouns.

In our experiments, we use 3-grams for most entities, except for country names, which we generate with 2-grams.
We use the \texttt{namemaker}~\cite{namemaker2024github} package that implements a Markov chain word generator.

\subsection{Data Selection and Processing}
Next, we provide details on the source and processing steps for each dataset. 

\subsubsection*{City Locations}
The \textit{City Locations} dataset is based on the \texttt{GeoNames} database~\cite{geonames}; \texttt{GeoNamesCache}~\cite{geonamescache2024pypi} is a Python package that interacts with the \texttt{GeoNames} API (accessed under the CC BY 4.0 license).
We use the following criteria to select a \(\langle\)city, country\(\rangle\) pair:
\begin{enumerate}[itemsep=1pt,topsep=1pt]
    \item The population of the city is at least 30,000.
    \item The city is associated with a country. 
    \item If a city name is associated with multiple countries, we include the \(\langle\)city, country\(\rangle\) pair for each country. 
    We exclude all cities that have \enquote{Antarctica} as a location or a country. 
\end{enumerate}

Since the resulting set of \(\langle\)city, correct country\(\rangle\) pairs is relatively large, we reduce the number of pairs by downsampling. In total, we select 1,400 unique city names: \(700\) cities with the highest populations, and \(700\) cities randomly sampled from the rest of the names.

\paragraph{Statement Structure.}
For each \(\langle\)city, country\(\rangle\) pair, we create statements of the form:  
\begin{quote}
    The city of \texttt{[city]} is (not) located in \texttt{[country]}.
\end{quote}
If a city name already contains the word \enquote{City} (e.g., \enquote{Guatemala City}), we do not start a sentence with \enquote{The city of.}
Finally, we sample \(\langle\)city, incorrect country\(\rangle\) pairs, and generate statements according to the template above.

\paragraph{Synthetic Entities.}
We use the technique described in~\Cref{sec:sup_data_unknowns} to generate synthetic city and country names.

To generate synthetic \textit{city} names, we collect all the city names in our dataset (including those that we did not include) and input them to \texttt{namemaker} (with \(n\)-gram length of 3).
We generate \(500\) synthetic city names. 
We validate these synthetic names in two stages:

\begin{enumerate}
    \item We check whether a synthetic name exists in the \texttt{GeoNames} database by looking for matches across \textit{name} and \textit{alternative name} fields. We keep \(310\) cities after this first stage.
    \item We use Google Search to validate that a synthetic city name does not exist via the following prompt: \enquote{city [city name]}. 
    If the search result returns a city with \(1\)-\(2\) character difference, we remove the synthetic name from the list. We keep \(219\) cities after this second stage.
\end{enumerate}

For the synthetic country names, we collect all the country names and input them to \texttt{namemaker} (with \(n\)-gram length of \(2\)).
We generate \(250\) synthetic names and validate them using the workflow described in the previous paragraph.
We keep \(238\) country names after the first stage, and \(138\) after the second stage. The Google Search prompt is: \enquote{country [country name]}.

Finally, with probability \(25\%\), we append a prefix or suffix to the synthetic country name.
The list of prefixes and suffixes includes \enquote{Island,} \enquote{Republic of,} \enquote{Kingdom,} \enquote{West,} \enquote{East,} \enquote{North,} \enquote{South,} and \enquote{Land.}
Finally, we randomly assign each synthetic city name to a synthetic country name. 

\subsubsection*{Medical Indications}
The \textit{Medical Indications} dataset is based on the \texttt{DrugBank} (version 5.1.12)~\cite{wishart2018drugbank}.
We obtained access to the \texttt{DrugBank} on October 4th, 2024, via the academic license (for research purposes only). 
Our code repository does not contain the raw data from the \texttt{DrugBank}, but the reader can apply for the academic license.\footnote{Here is the link to the \texttt{DrugBank}'s academic license: \href{https://go.drugbank.com/releases/5-1-12}{https://go.drugbank.com/releases/5-1-12}}
We extract 2 fields from this knowledge base:
\begin{enumerate}[itemsep=1pt,topsep=2pt]
    \item \textbf{Name}, which specifies the official name of the drug or the chemical (e.g., Lepirudin).
    \item \textbf{Indication}, which is a text field that describes the indication of the drug. 
    If this field consists of multiple sentences, we keep only the first sentence (e.g., \enquote{Lepirudin is indicated for anticoagulation in adult patients with acute coronary syndromes (ACS) such as unstable angina and acute myocardial infarction without ST elevation.})
\end{enumerate}

To extract diseases and conditions from the \textit{Indications} field, we use two named entity recognition (NER) models: 
\begin{enumerate}[itemsep=1pt,topsep=2pt]
    \item \texttt{SciSpaCy}'s \texttt{en\_ner\_bc5cdr\_md} model for the biomedical term annotations~\cite{neumann-etal-2019-scispacy},
    \item \texttt{BioBERT}-based NER~\cite{biobert2023tf,lee2020biobert} for disease annotations.
\end{enumerate}
We input the \enquote{Indication} text to both models. 
The disease/condition terms are extracted only if both models mark it as a disease or condition.
For example, for \enquote{Lepirudin is indicated for anticoagulation in adult patients with acute coronary syndromes (ACS),} the \texttt{SciSpaCy} model marks \textit{coronary syndromes} as a disease, but \texttt{BioBERT} does not. Thus, we do not add it to Lepirudin's disease/condition list. 
Similarly, we remove the abbreviation if the disease list contains the full name \textit{and} its abbreviation, such as [acute coronary syndromes, ACS]. 

\paragraph{Name validation.} We further validate the drug names using the \texttt{SciSpaCy} model and retain only those marked as \texttt{CHEMICAL}. Otherwise, we remove the drug from our dataset.
Finally, if the disease list (for a given drug) is empty after the preprocessing, we remove the drug from our dataset.

Additionally, we use the \texttt{wordfreq} package~\cite{speer2022wordfreq} to check whether the name of the drug or the name of the indication appears in widely used corpora (e.g., Wikipedia or books dataset).
In other words, we remove the pair if either the drug name or the indication has a Zipf frequency of \(0\) -- i.e., the word does not appear in any of the \texttt{wordfreq} corpora.

\paragraph{Statement Structure.}
For each \(\langle\)drug,  correct disease\(\rangle\) pair, we create statements of the form:
\begin{quote}
    \texttt{[drug]} is (not) indicated for the treatment of \texttt{[disease/condition]}.
\end{quote}
We also sample the \(\langle\)drug, incorrect disease\(\rangle\) pairs.
We ensure that the \enquote{incorrect disease} did not share any words with the diseases in the correct list. 

\textbf{Synthetic Entities}. 
To generate synthetic drug names and disease names, we use the approach described in~\Cref{sec:sup_data_unknowns} (with \(n\)-gram length of 3).
We generate \(500\) synthetic drug names.
We validate these synthetic names in two stages:
\begin{enumerate}[itemsep=1pt,topsep=2pt]
    \item We pass each generated name through the \texttt{SciSpaCy} model and remove the ones marked as \texttt{CHEMICAL}. We keep \(315\) names after this first stage.
    \item We use Google Search to validate that each drug name does not exist via the prompt \enquote{medicine [drug name].}
    If the search result returned a drug with \(1\)-\(2\) character difference, we remove it from the dataset. We keep \(243\) names after this second stage.
\end{enumerate}

We generate \(200\) disease names and check whether they exist in our list of diseases. We keep \(181\) names after this first stage.
Next, we use Google Search with the prompt \enquote{disease [disease/condition name].}  We keep \(131\) disease names after this second stage. 
Finally, we randomly match synthetic drug names to synthetic disease names to generate \textit{neither} statements.

\subsubsection*{Word Definitions}
The \textit{Word Definitions} dataset is based on a publicly available sample from~\citet{wordsapi} database.
It contains \(10\%\) of randomly sampled words from the database. 

For each word in the sample, we keep the ones that satisfy the following criteria:
\begin{enumerate}[itemsep=1pt,topsep=2pt]
    \item The word is a noun.
    \item The word has at least one definition in the \textit{definition} field.
    \item The word has at least one of the following fields:  \textit{synonym},  \textit{typeOf}, or \textit{instanceOf}.
\end{enumerate}

\textbf{Statement Structure.}
Depending on the specified field (i.e.,~\textit{synonym}, \textit{typeOf}, \textit{instanceOf}), we generate three types of statements:
\begin{enumerate}[itemsep=1pt,topsep=2pt]
    \item  \enquote{\texttt{[word]} is (not) \texttt{[instanceOf]}.}
    \item  \enquote{\texttt{[word]} is (not) a type of \texttt{[typeOf]}.}
    \item  \enquote{\texttt{[word]} is (not) a synonym of \texttt{[synonym]}.}
\end{enumerate}
Before inserting a word from \textit{synonym}, \textit{typeOf}, \textit{instanceOf} fields into a corresponding spot, we determine whether the appropriate indefinite article is `a' or `an.' When possible, we change words into singular forms.
To do so, we use the \texttt{inflect} package~\cite{inflect}.

\textbf{Synthetic Entities.}
To generate synthetic entities, we use the approach described in Supplementary Sec.~\ref{sec:sup_data_unknowns} (with \(n\)-gram length of 3).
We generate four categories of synthetic entities:
\begin{enumerate}[itemsep=1pt,topsep=2pt]
    \item Words that go at the beginning of each statement: We use all the words we have in the dataset.
    \item Types: We use all the words from the \textit{typeOf} field for the Markov Chain generation.
    \item Synonyms: We use all the words from the \texttt{synonym} field.
    \item Instances: We use words from the \textit{instanceOf} field.
\end{enumerate}
We generate 1,000 synthetic words for each of the four categories.
We validate the non-existence of words. 
We use the \texttt{english\_words} package\footnote{\href{https://pypi.org/project/english-words/}{pypi.org/project/english-words/}} to check whether a word exists in \enquote{GNU Collaborative International Dictionary of English \(0.53\),} or \texttt{web2} word list. Furthermore, we check whether there is a word in the \textit{words} list of the \texttt{nltk} package~\cite{bird-2006-nltk}. 
After this stage, we end up with 3,305 words.
Finally, we randomly sample pairs of \(\langle\)word, property\(\rangle\), where the property is a type, instance, or synonym.
\clearpage
\onecolumn
\section{Selection of Large Language Models}
\label{supsec:llms}
In this section, we provide an overview of the large language models used in our experiments. \Cref{tab:llm_overview} provides a list of all 16 models. 

We use default models (pre-trained on general tasks) and chat models fine-tuned on instruction- and chat-like interactions.
Every default model in our selection has a corresponding chat-based model. 
We add two additional chat-tuned \texttt{Llama} models, each fine-tuned on biomedical data.
Furthermore, we do not use the full official model names but use short names with a version, such as \enquote{chat} or \enquote{default}.
For example, \texttt{Llama-3.2} (chat) refers to the \texttt{Llama-3.2-3b-Instruct} model. Finally, we list the Hugging Face repositories in the notes for each reference.

\begin{table*}[h]
\centering
\begin{adjustbox}{width=1\textwidth}
\begin{tabular}{@{}llrrrl@{}}
\toprule
Official Model Name & Type & \# Decoders & Size &  Release Date & Source \\ \midrule
Gemma-7b & Default & 28 & 8.54 B & Feb 21, 2024 & \citet{team2024gemma} \\ 
Gemma-2-9b & Default & 26 & 9.24 B & Jun 27, 2024 & \citet{team2024gemma2} \\ 
Llama-3-8b & Default & 32 & 8.03 B & Apr 18, 2024 & Meta~\cite{grattafiori2024llama} \\ 
Llama-3.2-3b & Default & 28 & 3.21 B & Sep 25, 2024 & Meta~\cite{grattafiori2024llama} \\ 
Mistral-7B-v0.3 & Default & 32 & 7.25 B & May 22, 2024 & Mistral AI~\cite{mistral2023} \\ 
Qwen2.5-7B & Default & 28 & 7.62 B & Sep 19, 2024 & Alibaba Cloud~\cite{qwen2024report} \\ 
Qwen2.5-14B & Default & 38 & 14.80 B & Sep 19, 2024 & Alibaba Cloud~\cite{qwen2024report}\\ 
\midrule
Gemma-7b-it & Chat & 28 & 8.54 B & Feb 21, 2024 & \citet{team2024gemma} \\ 
Gemma-2-9b-it & Chat & 26 & 9.24 B & Jun 27, 2024 & \citet{team2024gemma2} \\ 
Llama-3.2-3b-Instruct & Chat & 28 & 3.21 B & Sep 25, 2024 & Meta~\cite{grattafiori2024llama} \\ 
Llama-3.1-8b-Instruct & Chat & 32 & 8.03 B & Jul 23, 2024 & Meta~\cite{grattafiori2024llama} \\ 
Llama3-Med42-8B & Chat & 32 & 8.03 B & Jul 2, 2024 & \citet{med42v2} \\ 
Bio-Medical-Llama-3-8B & Chat & 32 & 8.03 B & Aug 8, 2024 & \citet{contactdr2024} \\ 
Mistral-7B-Instruct-v0.3 & Chat & 32 & 7.25 B & May 22, 2024 & Mistral AI~\cite{mistral2023} \\ 
Qwen2.5-7B-Instruct & Chat & 28 & 7.62 B & Sep 19, 2024 & Alibaba Cloud~\cite{qwen2024report} \\ 
Qwen2.5-14B-Instruct & Chat & 38 & 14.80 B & Sep 19, 2024 & Alibaba Cloud~\cite{qwen2024report} \\ 
\bottomrule
\end{tabular}
\end{adjustbox}
\caption{\textbf{List of LLMs used in our experiments.} We provide the official names of the models in the Hugging~Face repository.
Furthermore, we specify the model's \textit{type}: `default' denotes the pre-trained models, and `chat' denotes the chat- or instruction-tuned versions. 
Finally, we provide the number of decoders, the number of parameters, the release date, and the source of the model. These models are publicly available through \href{https://huggingface.co/}{Hugging~Face}~\cite{wolf-etal-2020-transformers}.}
\label{tab:llm_overview}
\end{table*}

\clearpage
\twocolumn
\section{Criteria for Validating a Veracity Probe}
\label{sec:sup_validity}

\begin{table*}[!ht]
\begin{adjustbox}{width=1\textwidth}
\begin{tabular}{p{0.15\linewidth} p{0.32\linewidth}  p{0.23\linewidth} p{0.25\linewidth}}
\toprule
\textbf{Criteria} & \textbf{Definition} & \textbf{If Satisfied} & \textbf{Similar Concepts} \\ \midrule
\textbf{Correlation} & A probe \(g_i\) trained on  \(\langle h_i(\boldsymbol{x}), y\rangle\in \mathcal{D}_{\text{train}}\) should perform well (i.e., have high predictive accuracy) on \(\mathcal{D}_{\text{test}}\), assuming the same input and label distributions. & $\mathcal{M}$ encodes information correlated with veracity; for the result refer to~\Cref{fig:bar_performance}.A. & Information \cite{Harding2023operationalising}, Accuracy \cite{herrmann2025standards} \\ \midrule
\textbf{Generalization} & A probe \(g_i\) trained on \(\langle h_i(\boldsymbol{x}), y\rangle\in \mathcal{D}_{\text{train}}\) should have 
high predictive accuracy on data from different domains. 
& \(\mathcal{M}\) has a universal activation pattern correlated with veracity; for the results refer to~\Cref{fig:bar_performance}.B and~\Cref{fig:generalization_performance}.  & Generalization as defined by \citet{burger2024truth}, Uniformity \cite{herrmann2025standards} \\ \midrule
\textbf{Selectivity} & A probe \(g_i\) trained on \(\langle h_i(\boldsymbol{x}), y \rangle \in \mathcal{D}_{\text{train}}\) should not assign \textit{true} or \textit{false} labels to samples where truthfulness is absent or undefined. & \(\mathcal{M}\) has a distinct mechanism that correlates exclusively with veracity. & Misrepresentation as defined by \citet{Harding2023operationalising}, Control Task \cite{hewitt-liang-2019-designing} \\ \midrule
\textbf{Manipulation} & Modifying \(h_i(\boldsymbol{x})\) along \(\boldsymbol \theta_i\) should systematically alter $P_{\mathcal{M}}(\tau\mid\boldsymbol{x})$ for tokens $\tau$ related to the veracity property \(Z\). & \(\mathcal{M}\) has a \textit{linear} mechanism to track veracity and uses it to compute the output $P_{\mathcal{M}}(\tau\mid\boldsymbol{x})$. & Use \cite{Harding2023operationalising}, Addition~\cite{arditi2025refusal}, Intervention~\cite{meng2022locating} \\ \midrule
\textbf{Locality} & Modifying \(h_i(\boldsymbol{x})\) along \(\boldsymbol \theta_i\) should not significantly alter $P_{\mathcal{M}}(r\mid\boldsymbol{x})$ for random tokens \(r\) that are unrelated to \(Z\). & $\mathcal{M}$ maintains a separate mechanism that tracks veracity, without being confused with other concepts. & Misrepresentation~\cite{Harding2023operationalising}, Leakage \cite{elazar-goldberg-2018-adversarial} \\ \bottomrule
\end{tabular}
\end{adjustbox}
\caption{\textbf{Validity criteria for representation-based probes.}
If satisfied, these criteria serve as validation that \(g_i\) indeed captures signals associated with veracity \(Z\).
Here, we provide a formal definition of each criterion, along with the implications of satisfying the criterion. 
Finally, we provide the list of similar criteria and concepts used in the literature.
}
\label{tab:criteria}
\end{table*}

\citet{herrmann2025standards}, \citet{zou2023representation} and \citet{Harding2023operationalising} have proposed criteria for measuring the validity of veracity probes.
We aggregate these into five major categories and provide an overview in~\Cref{tab:criteria}.
We propose to evaluate a probe \(g_i\) along the following criteria:
\begin{enumerate}[label=(\roman*), topsep=1pt, itemsep=1pt]
    \item \textbf{Correlation}.
    The probe, trained to predict a veracity property \(Z\), should achieve high predictive accuracy on unseen samples that possess this property, i.e., on samples from \(\mathcal{D}_{test}\).
    When the criterion is satisfied, the 
    \(i\)th decoder embeds the information about \(Z\) to some degree. We cannot rule out the fact that it captures proxies associated with \(Z\).
    \item \textbf{Generalization} extends \textit{Correlation} by requiring that the probe generalizes beyond the dataset it was trained on.
    The probe should achieve high predictive accuracy on samples with veracity \(Z\), but have different phrasing or come from different domains.
    For example, if a probe is trained on representations associated with veracity for ecological statements, it should exhibit similar predictive accuracy on geographical statements. 
    \item 
    \textbf{Selectivity}.
    The probe \(g_i\) should avoid classifying statements that are not (or cannot be) \textit{true} or \textit{false}.
    Instead, the probe \(g_i\) should abstain from making predictions on the \textit{neither} statements---i.e., statements that the LLM could not have learned from its training data or that inherently lack any truthfulness or falsehood.
    Poor selectivity indicates that the probe might capture spurious correlations.
    \item 
    \textbf{Manipulation}.
    We should be able to use the identified direction \(\boldsymbol \theta_i\) to update \(h_i(\boldsymbol{x})\) and have a predictable change in the distribution of the output tokens \(P_{\mathcal{M}}(\tau \mid\boldsymbol{x})\).
    \item
    \textbf{Locality}.
    When asking a question such as \enquote{Is X true? Answer yes or no,} the manipulation should primarily influence the probabilities related to the tokens \enquote{yes} or \enquote{no}. It should minimally affect unrelated tokens.
    For example, if a manipulation does not increase the likelihood of the LLM generating \enquote{no,} but increases the likelihood of generating tokens such as \enquote{elephant,} then the identified veracity direction affects not only veracity-related signals but also unrelated ones. 
\end{enumerate}

Evaluating a probe according to these criteria allows us to determine how well \(g_i\) captures the signals associated with veracity \(Z\) and how manipulations (a.k.a.~interventions) affect the LLM's output \(P_{\mathcal M}\).
Part of our future work includes adding a new criterion to assess whether a probe can determine whether an LLM can \enquote{reason} logically. For example, if \(\mathcal M\) classifies a statement $\phi_1$ as true and another statement $\phi_2$ as true, then will \(\mathcal M\) also classify $\phi_1 \wedge \phi_2$ as true?

\paragraph{Results.} 
We evaluate all five criteria across our experiments. Results for \textbf{Correlation}, \textbf{Selectivity}\footnote{The multiclass MCC allows us to evaluate both the correlation and selectivity criteria at the same time.} and \textbf{Generalization} are reported in~\Cref{fig:bar_performance} and additional details are provided in~\Cref{supsec:evaluation}. The results for \textbf{Manipulation} and \textbf{Locality} are reported in~\Cref{fig:nie-main} and additional details are provided in~\Cref{supsec:interventions}.

\clearpage
\twocolumn
\section{Conformal Prediction Sets}
\label{supsec:nonconformity}

In our work, we focus on \textit{split conformal learning}~\cite{angelopoulos2020uncertainty}, which requires a hold-out (or calibration) dataset to compute conformal prediction sets. 
Given a probe \(g_i\), we use a calibration dataset $\mathcal{D}_{cal}$ of activation-label pairs $\langle h_i(\boldsymbol x), y \rangle$ to find prediction sets \(C(\boldsymbol{x}) \subset \mathcal{Z}\) with a marginal coverage guarantee; assuming the calibration and test data are exchangeable, \(C(\boldsymbol{x})\) contains the correct label with probability at least \(1-\alpha\).
Given \(\alpha=0.1\), the correct label is covered at least $90\%$ of the time \textit{on average over the data}.
We provide pseudocode for the nonconformity functions in~\Cref{alg:sup_cp_binary,alg:sup_cp_multiclass}.

\paragraph{Nonconformity scores.}
To identify the conformal sets, we compute a nonconformity score for each sample in $\mathcal{D}_{cal}$.
For the binary cases (such as mean-difference probe and one-vs-all \texttt{sAwMIL}), we use the binary nonconformity scoring; see~\Cref{alg:sup_cp_binary}.
It is based on the distance between the prediction and the classifier's separating hyperplane.
In~\Cref{eq:nc_symmetric}, $s$ is the signed distance of the sample to the separation hyperplane, and $y$ is a ground-truth (or candidate) label:\footnote{In this case, labels should be either \(-1\) or \(1\). Thus, all samples with label \(0\) are assigned label \(-1\).}
\begin{equation}
\label{eq:nc_symmetric}
    \operatorname{binaryNC}\,(s,y) = \exp\bigl(-y \,\cdot\,s\bigr), 
\end{equation}
where \(\:y\in\{-1,1\}\text{ and } s\in\mathbb{R}.\)
If the sample ends up on the wrong side of the separation hyperplane (e.g., $s>0$ and $y=-1$), then the nonconformity score in~\Cref{eq:nc_symmetric} is high and the candidate label is weakly supported by the model.

For the multiclass \texttt{sAwMIL}, we use the multiclass nonconformity score~\cite{johansson2017model};  see~\Cref{alg:sup_cp_multiclass}.
For a given candidate label \(y\), the score is defined in terms of the difference between the predicted probability of the true class and the highest probability among the other classes (with \(K\) denoting the total number of classes). Formally, for a candidate label \(y\) with predicted probability \(p_y\), we calculate the multiclass nonconformity score with the following function:
\begin{equation}
\label{eq:nc_margin}
        \operatorname{multiNC}\left(\mathbf{p}, y\right) = \frac{1 - (p_y - \max_{i \neq y} p_i)}{2},
\end{equation}
where $\mathbf{p} \in \Delta^{K-1}$ is the predicted probability vector, 
$\Delta^{K-1} := \{\mathbf{p} \in \mathbb{R}^K \mid p_i \geq 0,\, \sum_i p_i = 1\}$, 
and $p_y$ denotes the predicted probability assigned to the candidate label $y$.

In both cases, lower scores in~\Cref{eq:nc_symmetric,eq:nc_margin} indicate that the candidate label \(y\) is strongly supported by the model. 
For each candidate label \(y\), we compute its conformal \(p\)-value \(\psi_y\) from the nonconformity scores and include \(y\) in the conformal prediction set \(C\) if \(\psi_y > \alpha\), where \(\alpha=0.1\). The probe abstains when the resulting prediction set is empty or contains more than one label (see~\Cref{alg:sup_cp_binary,alg:sup_cp_multiclass}). 
The addition of conformal predictions allows us to distinguish between cases where the statements originate from different distributions, as compared to those in the calibration dataset.
\clearpage
\section{More on Representation-based Probing Methods}
\label{supsec:sota_sanity}

\subsection{Mean-difference Probe with Conformal Prediction}
\label{sec:sup_mdcp_details}

The mean-difference probe (\texttt{MD+CP}) consists of two components: a binary mean-difference classifier (\texttt{MD}) and the conformal prediction (\texttt{CP}). 

First, we fit the binary classifier with a linear decision boundary~\cite{marks2023geometry}.
We use it to separate \textit{true} and \textit{false} statements based on the internal representations \(h_i\).
For each pair \(\langle \boldsymbol x_j, y_j\rangle\), we extract the activation of the last token \(h_{i}(\boldsymbol x_j)_{[L]}\) and assemble a set of factually true \(\mathcal{X}^+ = \{h_{i}(\boldsymbol{x_j})_{[L]}: y_j=\text{true}\}\) and a set of false \(\mathcal{X}^- = \{h_{i}(\boldsymbol{x_j})_{[L]}: y_j = \text{false}\}\) activations. Here, \(L\) is the index of the last token in \(\boldsymbol x\).
We then compute the means of each set, denoted \(\boldsymbol{\mu}^+\) and \(\boldsymbol{\mu}^-\), and get the mean-difference vector
\begin{equation}
\label{eq:mean_diff}
    \boldsymbol{\theta} = (\boldsymbol{\mu}^+ - \boldsymbol{\mu}^-)\text{, where } \boldsymbol{\theta}\in \mathbb{R}^{1\times d}.
\end{equation}
Statements are then scored along the whitened direction \(\boldsymbol{\Sigma}^{-1}\boldsymbol{\theta^{\top}}\), where \(\boldsymbol{\Sigma}\) is the pooled within-class covariance matrix; refer to the pseudocode in~\Cref{alg:mean_diff}.

Second, we augment \texttt{MD} with conformal prediction~\cite{angelopoulos2020uncertainty}.
Conformal predictions help detect statements that fall outside \texttt{MD}'s high-confidence regions for \textit{true} or \textit{false} classes.
We use $\alpha = 0.1$ in our experiments; that is, the conformal prediction sets (or regions) have a marginal coverage of at least $90\%$ (assuming the calibration and test data are exchangeable), the correct label lies in the region at least $90\%$ of the time \textit{on average over the data}.
Note that we use the \textit{true} and \textit{false} statements from the calibration dataset to find the conformal prediction regions. 

Finally, we test the \texttt{MD+CP} probe using \textit{true}, \textit{false}, and \textit{neither} statements from the test set. \emph{How can the binary \texttt{MD+CP} classifier identify \textit{neither} statements in addition to \textit{true} and \textit{false} statements?} 
If the \texttt{MD+CP} probe accurately captures the veracity signal, the \textit{neither} statements (from the test dataset) should fall outside of the conformal regions.
We observe that this is not the case: \texttt{MD+CP} assigns high-confidence scores to the \textit{neither}-valued statements in the \textit{true} or \textit{false} regions.

\begin{figure}[ht]
\centering
  \includegraphics[width=\linewidth]{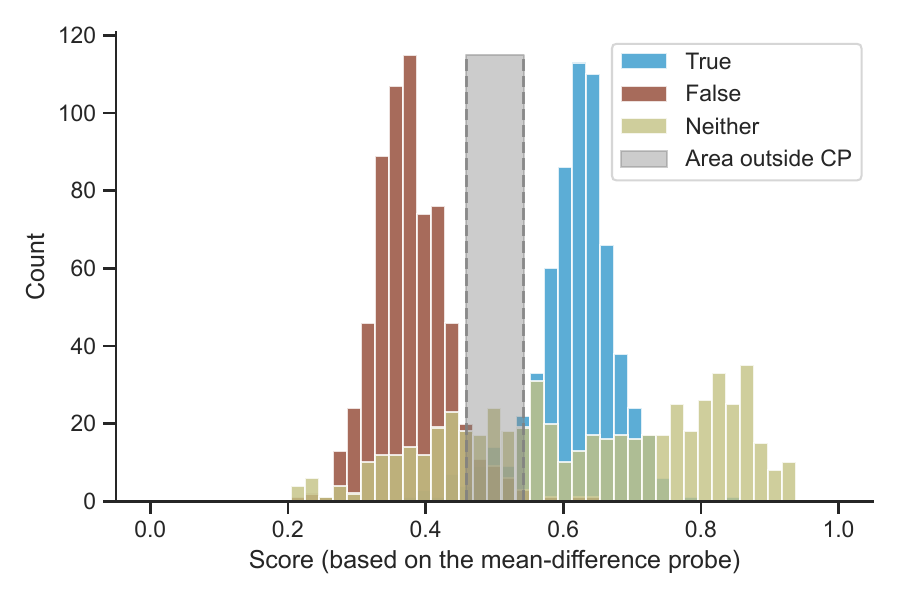}
  \caption{\textbf{Score distributions of mean-difference probe with conformal prediction (\texttt{MD+CP}) on the 13\textsuperscript{th} decoder activations of the default \texttt{Llama-3-8B} model for the City Locations dataset.}
The probe provides a good separation between \textit{true} and \textit{false} statements. 
If the \texttt{MD+CP} probe \textit{truly} captures the veracity signal, we expect the scores for the \textit{neither} statements to fall outside the conformal region (i.e., in the gray area), providing an \textbf{empty prediction set} (so the statement is labeled \textit{neither}).
However, \texttt{MD+CP} assigns high-confidence scores for the \textit{neither} statements, labeling them \textit{true} or \textit{false}. This finding suggests that \texttt{MD+CP} relies on spurious proxies rather than genuine veracity signals.}
  \label{fig:sup_mdcp_score_distribution}
\end{figure}

\Cref{fig:sup_mdcp_score_distribution} shows the score distributions\footnote{We use the embedding of the last token to compute the scores in~\Cref{fig:sup_mdcp_score_distribution}:
$g_i\left(h_i\left(\boldsymbol x\right)\right) = \boldsymbol \theta^{\top} h_i(\boldsymbol{x})_{[L]} + \beta$. Here, \(L\) is the total length of the statement, which is the same as the index of the last token.}  of \texttt{MD+CP} on the best-performing decoder (i.e., 13\textsuperscript{th}) of the default \texttt{Llama-3-8B} model on the \textit{City Locations} dataset.  There are three distributions: one for \textit{true}, one for \textit{false}, and one for \textit{neither}. 
If \texttt{MD+CP} correctly captures the veracity signal, we expect the distribution of \textit{neither} statements (green bars) to be outside of the conformal prediction region (i.e., in the gray area) and it is not included in the conformal prediction set.
However, this is not the case. 
Most of the \textit{neither} statements fall within the conformal prediction region (i.e., not in the gray area) and are labeled as \textit{true} or \textit{false}.

\begin{figure*}[t]
  \includegraphics[width=\linewidth]{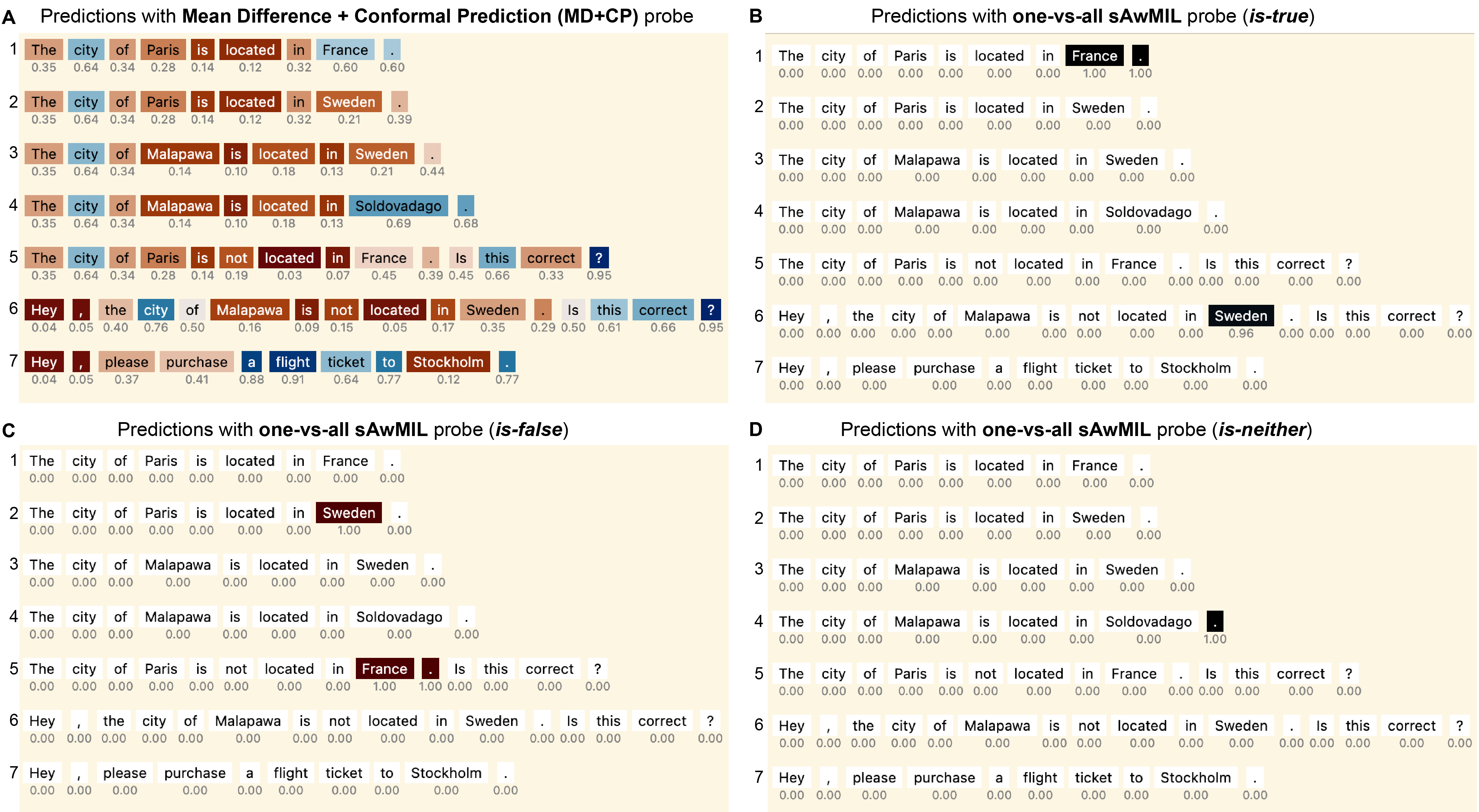}
  \caption{\textbf{Per-token predictions of mean-difference with conformal prediction (\texttt{MD+CP}) and one-vs-all \texttt{sAwMIL} on the 13\textsuperscript{th} decoder activations of the default \texttt{Llama-3-8B} model.}
Statements are from the \textit{City Locations} dataset. We show per-token probabilities (printed beneath each word), assigned based on the token's representation. Words are shaded based on the predicted probability. When MD+CP outputs 0, the statement is labeled false; when it outputs 1, the statement is labeled true. If the per-token score falls outside the conformal prediction region, \texttt{MD+CP} assigns a score of \(0.5\) to that token (which corresponds to the highest uncertainty). The one-vs-all \texttt{sAwMIL} probe for \texttt{is-true} outputs 1 when the probe is 100\% confident that the statement is true; and it outputs 0 when the per-token score is outside the conformal prediction region (i.e., there is an absence of truthfulness signal). Similarly, the one-vs-all \texttt{sAwMIL} probe for  \texttt{is-false} outputs 1 when the probe is 100\% confident that the statement is false; and it outputs 0 when the per-token score is outside the conformal prediction region (i.e., there is an absence of falsehood signal). The same logic applies for the one-vs-all \texttt{sAwMIL} probe for \texttt{is-neither}.
\textbf{Panel A} shows the \texttt{MD+CP} predictions. It often assigns high confidence scores to pre-actualization tokens and makes mistakes on the \textit{wrapped} prompts in cases \(\#5 \textrm{ and } 6\) (e.g.,\ statement \#6: ``Hey,\_\_\_ Is this correct?''). 
Also, \texttt{MD+CP} probe assigns labels to the statement without any truth value (e.g., case \(\#7\)).
\textbf{Panels B--D} display one-vs-all \texttt{sAwMIL} probes (\texttt{is-true}, \texttt{is-false}, and \texttt{is-neither}). Unlike \texttt{MD+CP}, one-vs-all \texttt{sAwMIL} localizes the veracity signal to the actualized token and abstains elsewhere, demonstrating superior selectivity.}
  \label{fig:sup_sota_sanity}
\end{figure*}

\Cref{fig:sup_sota_sanity}.A illustrates per-token \texttt{MD+CP} predictions across entire statements. If \texttt{MD+CP} correctly identifies the veracity signal, then (1) it should not assign any labels to the tokens in the pre-actualized parts of statements \(\boldsymbol x^p\), and (2) the label should be consistent across the actualized part \( \boldsymbol x^a\). 
Given a statement \enquote{The city of Tokyo is in Japan.}, the pre-actualized part is \enquote{The city of Tokyo is in}, and the actualized part is \enquote{Japan.}

We observe that \texttt{MD+CP} assigns scores to the pre-actualized tokens that fall within the conformal prediction regions in cases \#\(1\)--\(5\) (see~\Cref{fig:sup_sota_sanity}).
In case \#5, \texttt{MD+CP} assigns a correct prediction at the \textit{period} (\(p = 0.39\) corresponds to a false label), but the prediction flips at the end of the text, where the \textit{question mark} gets \(p = 0.95\) corresponding to the \textit{true} label. 
Similarly, in \#\(7\) case of~\Cref{fig:sup_sota_sanity}.A, the sentence does not have \textit{any} truth value (i.e., it is not a factual claim). 
The \texttt{MD+CP} probe assigns high confidence scores to some of its tokens.
These findings suggest that \texttt{MD+CP} probe captures proxies or spurious correlations. 
One cannot use it in real scenarios, where we do not know a priori where the factual claim ends.
In contrast,~\Cref{fig:sup_sota_sanity}.B--D show the per-token predictions for the one-vs-all \texttt{sAwMIL}.
These probes correctly identify positions where the veracity is actualized. For example, they do not assign predictions to the non-actualized parts of the statements and only label tokens in the actualized part. 
Moreover, these probes do not label tokens in cases where veracity is absent (e.g., see case \(\# 7\) in~\Cref{fig:sup_sota_sanity}.D).

\subsection{Multiclass Single-Instance Support Vector Machine}
\label{sec:sup_svm_details}

\begin{figure*}[ht]
\centering
  \includegraphics[width=\linewidth]{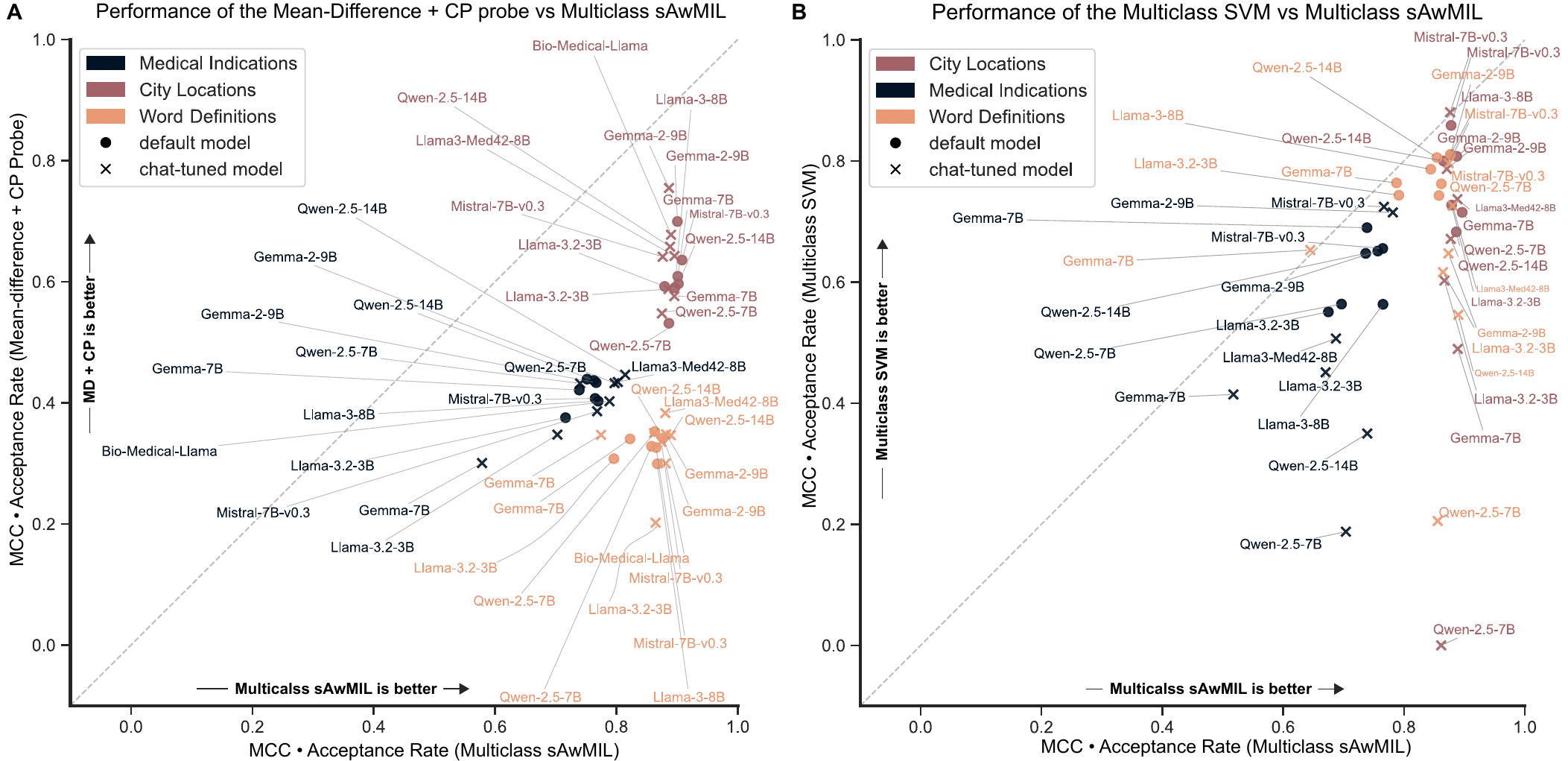}
\caption{\textbf{Comparison of performance for \texttt{MD+CP}, a multiclass SVM and multiclass \texttt{sAwMIL} probes.}
\textbf{Panels A \& B:} Each marker shows a probe's performance for a $\langle$model, dataset$\rangle$ pair.
Default models are shown with circles, while chat models are shown with crosses. Different colors indicate different datasets.  
\textbf{Panel A} shows the comparison between the multiclass \texttt{sAwMIL} probe on the x-axis and the mean-difference probe with conformal prediction (\texttt{MD+CP}) on the y-axis. 
\textbf{Panel B}: Performance of multiclass \texttt{sAwMIL} vs.~multiclass single-instance SVM probes.
The performance of the multiclass \texttt{sAwMIL} probe is specified on the x-axis, while the performance of the multiclass SVM is specified on the y-axis. We observe that multiclass \texttt{sAwMIL} outperforms multiclass SVM. The only exceptions are the \texttt{Gemma-7B} chat model and the \texttt{Mistral-7B-v0.3} chat model on \textit{Word Definitions}, where the performances of the single-instance and multiple-instance probes are competitive. This experiment shows that multiple-instance learning (i.e., training on all the tokens in the statement) is beneficial when tracking the veracity of an LLM.
Overall, the multiclass \texttt{sAwMIL} probe outperforms \texttt{MD+CP} and the multiclass SVM.
}
  \label{fig:sup_sota_sawmil_vs_svm}
\end{figure*}

The multiclass \texttt{sAwMIL} probe is a multiple-instance learning version of a support vector machine, designed to operate on bags of token representations.
To assess whether the MIL formulation offers any benefits, we construct a single-instance baseline by training a multiclass SVM on the last token representation only \(h_i(\boldsymbol x)_{[L]}\).
As with multiclass \texttt{sAwMIL}, we first train three one-vs-all probes: \texttt{is-true}, \texttt{is-false}, and \texttt{is-neither}.
Then, we assemble one-vs-all classifiers into a multiclass SVM using the same procedure described in~\Cref{sec:multiclass-sawmil} and~\Cref{alg:sawmil}.

As before, to evaluate performance, we provide all token representations \(h_i(\boldsymbol x)\) (not only the last one), where the final prediction is computed based on 
\begin{equation}
    \hat g_i(\boldsymbol x) \;=\; \max_{1 \,\le\, j \,\le\, L} \;g_i\bigl(h_i( \boldsymbol x)_{[j]}\bigr),
\end{equation}
where \( L=\left|\boldsymbol x \right|\) (number of tokens in \(\boldsymbol{x}\)).

\Cref{fig:sup_sota_sawmil_vs_svm} depicts the performance of multiclass \texttt{sAwMIL} vs.~multiclass SVM. The performance of multiclass SVM is closer to the performance of multiclass \texttt{sAwMIL} (as compared to \texttt{MD+CP} probe in~\Cref{fig:sup_sota_sawmil_vs_svm}.A). However, multiclass \texttt{sAwMIL} still outperforms the multiclass single-instance SVM in 46 out of 48 cases ($=16 \textrm{ LLMs} \times 3 \textrm{ datasets}$). For more results, we refer the reader to~\Cref{sup_tab:mcc_long_sawmil_bag} and~\Cref{sup_tab:mcc_long_svm_bag}. 

In~\Cref{fig:sup_sota_sawmil_vs_svm}, we also observe that the multiclass \texttt{sAwMIL} performs better on the chat models (see bottom right portion of the plot) than the multiclass SVM. 
This supports our claim that veracity signals often emerge at positions other than the final token, and that multiple-instance learning can better isolate the veracity signal.
Recall that the multiclass \texttt{sAwMIL} probe considers all the tokens in the statement and has additional training stages. 

\begin{figure*}
  \includegraphics[width=\linewidth]{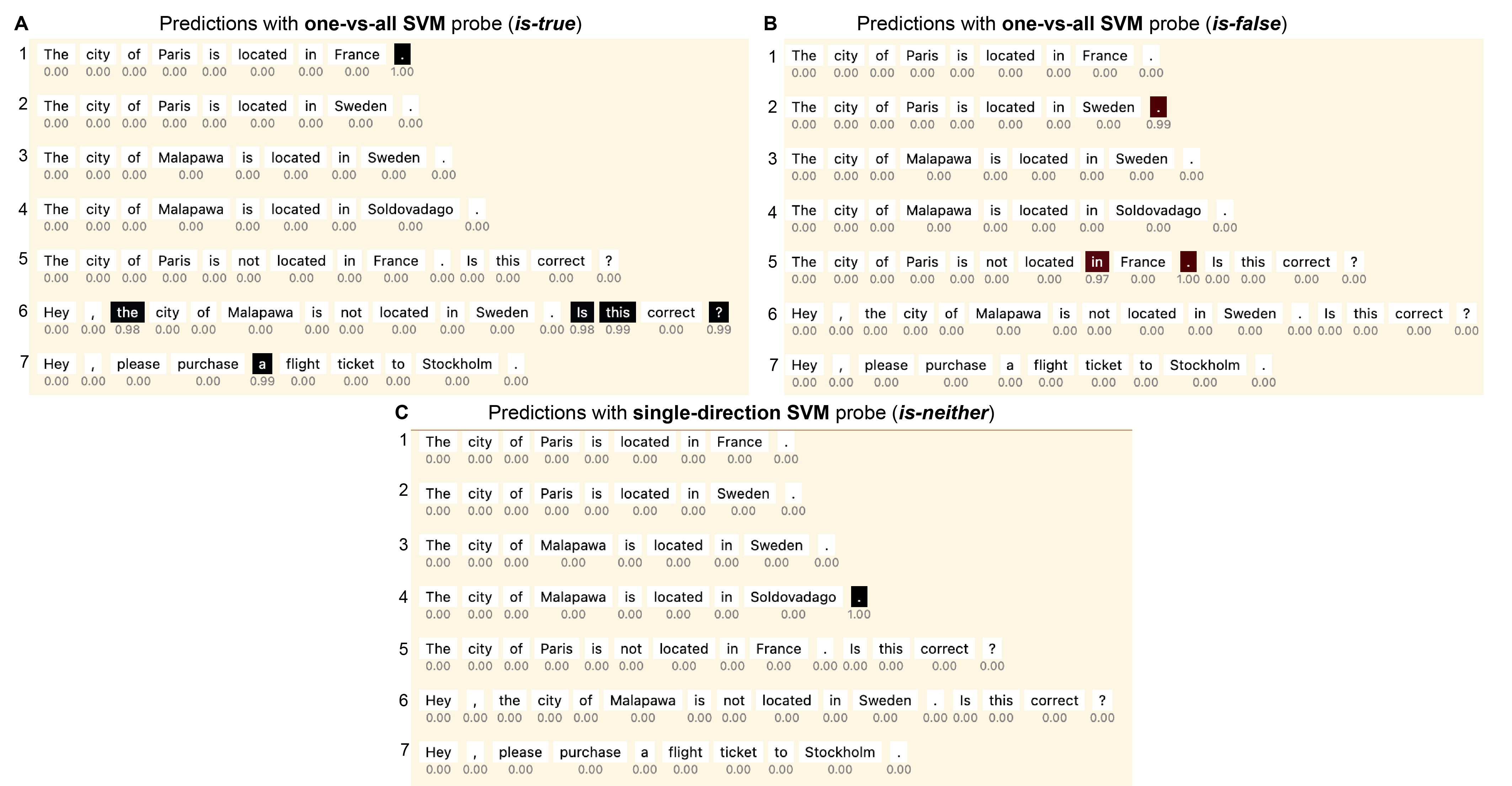}
  \caption{\textbf{Per-token predictions of one-vs-all single-instance SVM on the 13\textsuperscript{th} decoder activations of the default \texttt{Llama-3-8B}}.
Statements are from the \textit{City Locations} dataset. We show per-token probabilities (printed beneath each word), assigned based on the token's representation. Words are shaded based on their predicted probability.
\textbf{Panels A--C} display the one-vs-all \texttt{SVM} probes (\texttt{is-true}, \texttt{is-false}, and \texttt{is-neither}). 
The one-vs-all SVM probe isolates the signal better than the \texttt{MD+CP} probe. (See~\Cref{fig:sup_sota_sanity}.A for the \texttt{MD+CP} results.)
However, in some cases, the one-vs-all SVM probe assigns high-certainty scores to tokens that lack any veracity signal. 
For example, in Panel A (case \(\# 6\)), the probe picks up on tokens including `this,' `is,' and `?,' which do not have inherent truth value.
Overall, the \texttt{sAwMIL} in~\Cref{fig:sup_sota_sanity}.B--D has better selectivity.
}
  \label{fig:sup_sota_sanity_2}
\end{figure*}

\Cref{fig:sup_sota_sanity_2} visualizes the per-token predictions.
The one-vs-all SVM-based probes have better selectivity than the mean-difference probe with  conformal prediction (\texttt{MD+CP}).
However, in some cases, one-vs-all SVM assigns labels to the tokens in the pre-actualized part of the statement (see~\Cref{fig:sup_sota_sanity_2}.B, \(\#5\) statement).
This suggests that one-vs-all SVM probes are capturing spurious correlations with potential proxies.

\clearpage
\twocolumn
\section{Zero-Shot Prompting: Instructions, Veracity Labeling, and Abstention}
\label{sec:sup-zero-shot}

\paragraph{Instructions in Zero-shot Prompting.} In zero-shot prompting, each LLM receives an instruction along with a statement \(\boldsymbol x\) as input.
These instructions outline the task and describe the output format.
We use zero-shot prompts to evaluate how well LLMs can assess the veracity of a given statement and a multiple-choice question.

\begin{figure*}[ht]
    \centering
    \includegraphics[width=\linewidth]{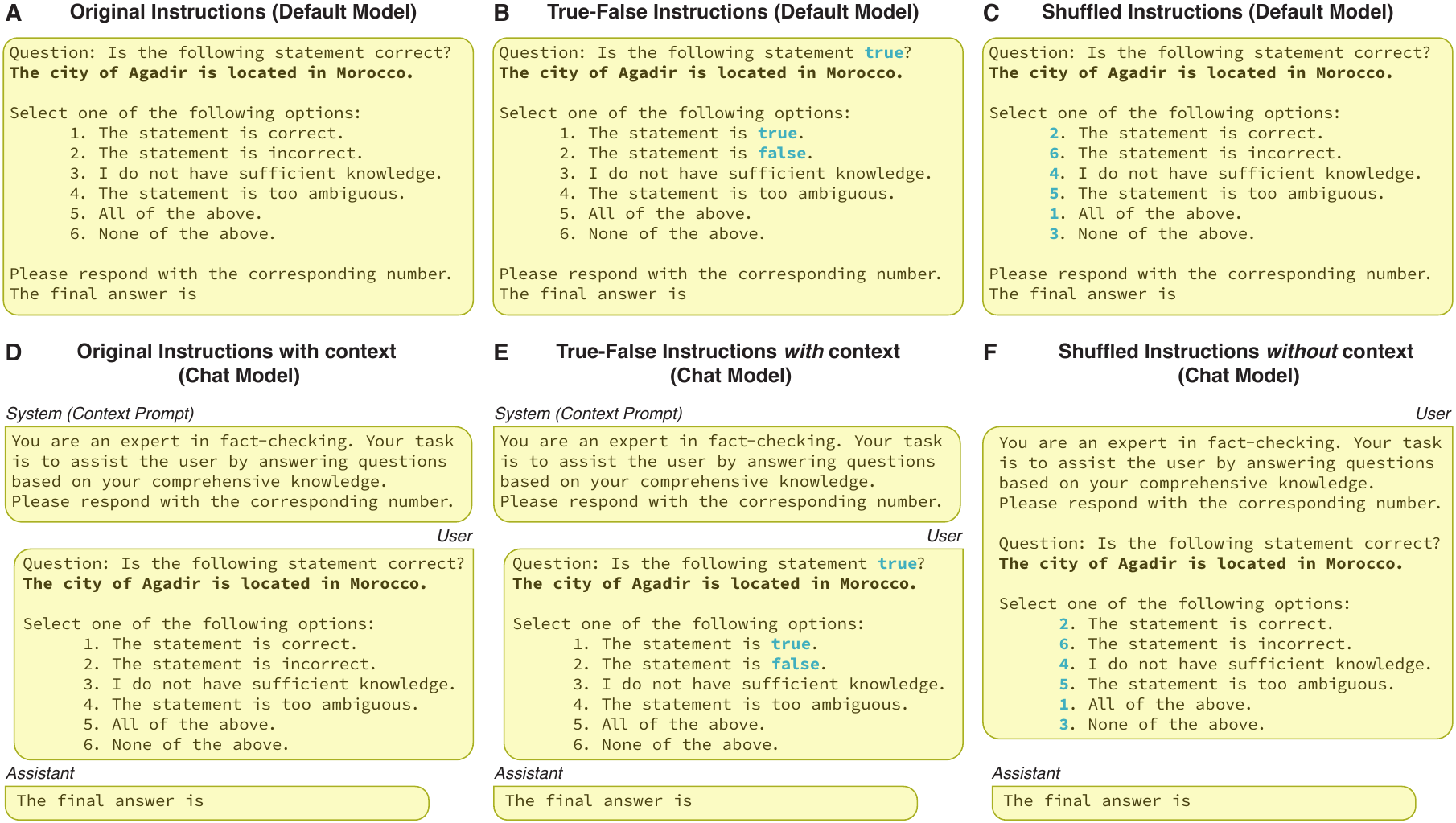}
    \caption{\textbf{Zero-shot prompt templates.} 
    We use these templates in our experiments. 
    \textbf{Panel A--C:} The prompts for the default models.
    \textbf{Panel D--F:} Examples of prompts used for the chat models. 
    Note that chat models like \texttt{Gemma} do not have a context (or system) prompt; hence, we provide instructions in the first message (see Panel F).
    In the main manuscript, we report the performance over the \textit{original instructions}, i.e., instructions in Panels A and D. For the chat LLMs without the context prompt, we apply the two-message template in Panel F, but use the \textit{original instructions}. Note: The statement used in these examples is factually correct.}
    \label{fig:zero-shot-example}
\end{figure*}

In our (zero-shot prompting) experiments, we use three different templates.
All default models share the same template as displayed in~\Cref{fig:zero-shot-example}.A--B.
We do not use this template for chat models, since they support \enquote{turn-based} conversations.
Thus, for most chat models, we use templates that simulate user-assistant dialog. 
We use the template in~\Cref{fig:zero-shot-example}.D--E for the chat models that support context prompts.
We use the template in~\Cref{fig:zero-shot-example}.C and in~\Cref{fig:zero-shot-example}.F for \texttt{Gemma} chat models since they do not support \textit{context prompts}.
Additionally, we use three different phrasings of the instructions:
\squishlisttwo
    \item \textbf{Original instructions} are~\Cref{fig:zero-shot-example}.A and~\Cref{fig:zero-shot-example}.D. 
    In the main paper, we report results based on these instructions.
    \item \textbf{True-False instructions}, where we change the phrasing of the original instructions by replacing the words \enquote{correct} and \enquote{incorrect} with \enquote{true} and \enquote{false}. These are displayed in~\Cref{fig:zero-shot-example}.B and~\Cref{fig:zero-shot-example}.E.
    \item \textbf{Shuffled instructions}, where we change the original instructions by shuffling choice identifiers. These are displayed in~\Cref{fig:zero-shot-example}.C and ~\Cref{fig:zero-shot-example}.F.
\squishend

\begin{figure*}[ht]
    \centering
    \includegraphics[width=\linewidth]{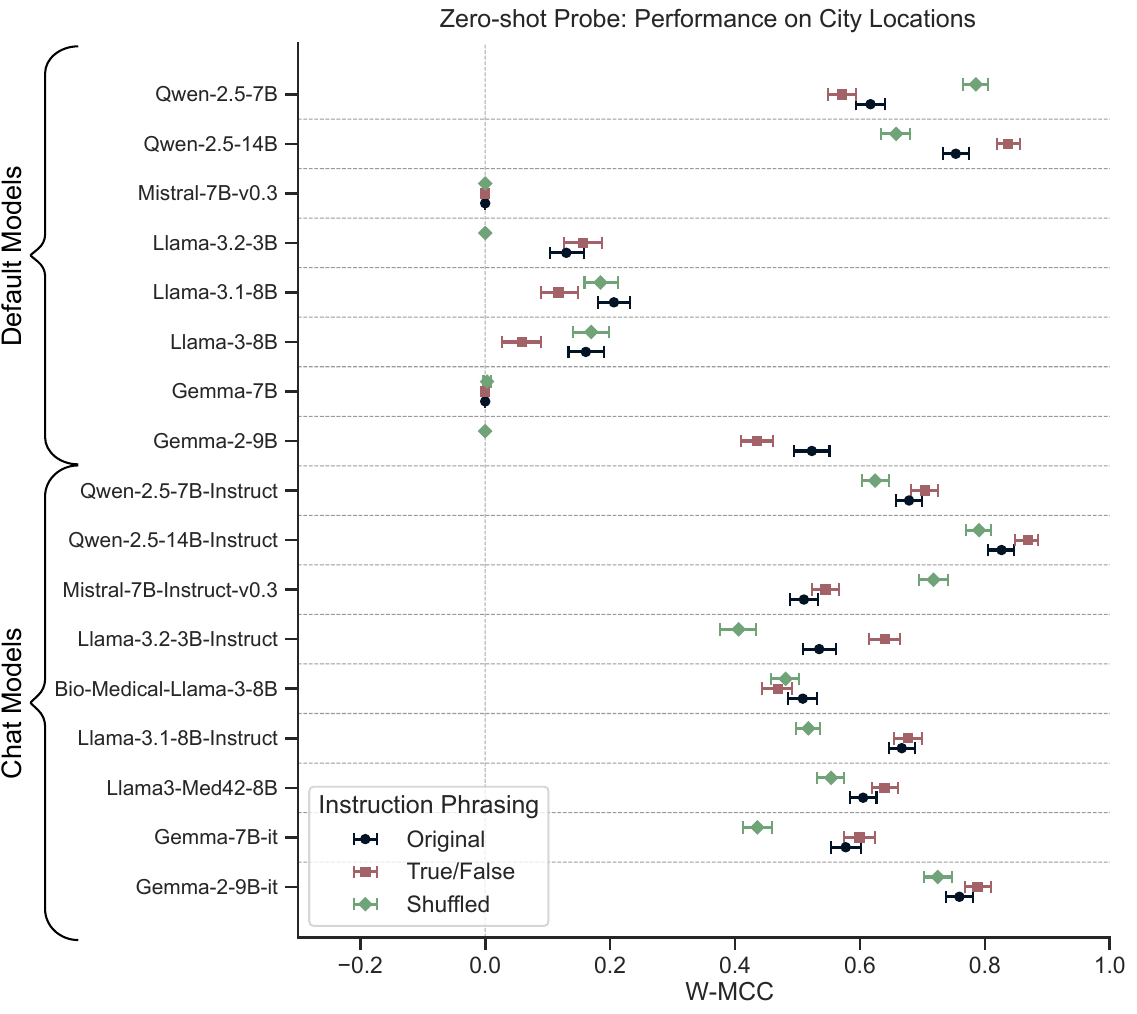}
    \caption{\textbf{Performance of zero-shot prompting on the \textit{City Locations} dataset across different models and instruction phrasings.} 
    We use the Weighted Matthews Correlation Coefficient (W-MCC) to quantify the performance. The marker shows the mean value and the error bars show the \(95\%\) confidence intervals (based on bootstrapping with \(n=\) \(1{,}000\) bootstrap samples).
    Minimal changes to the prompt instructions can skew the performance of zero-shot prompting.
    Chat models exhibit the highest performance across all instruction phrasings.  
    However, the default \texttt{Qwen} models match the performance of other chat-based models. 
    Shuffled instructions appear to lead to worse performance in chat models. We expected that the phrasings would have only a minor effect on their performance.
    The default \texttt{Gemma} and \texttt{Mistral} models seem to fail (their performance is around \(0\)).
    }
    \label{fig:app_zero_shot_cities}
\end{figure*}

\begin{figure*}[ht]
    \centering
\includegraphics[width=\linewidth]{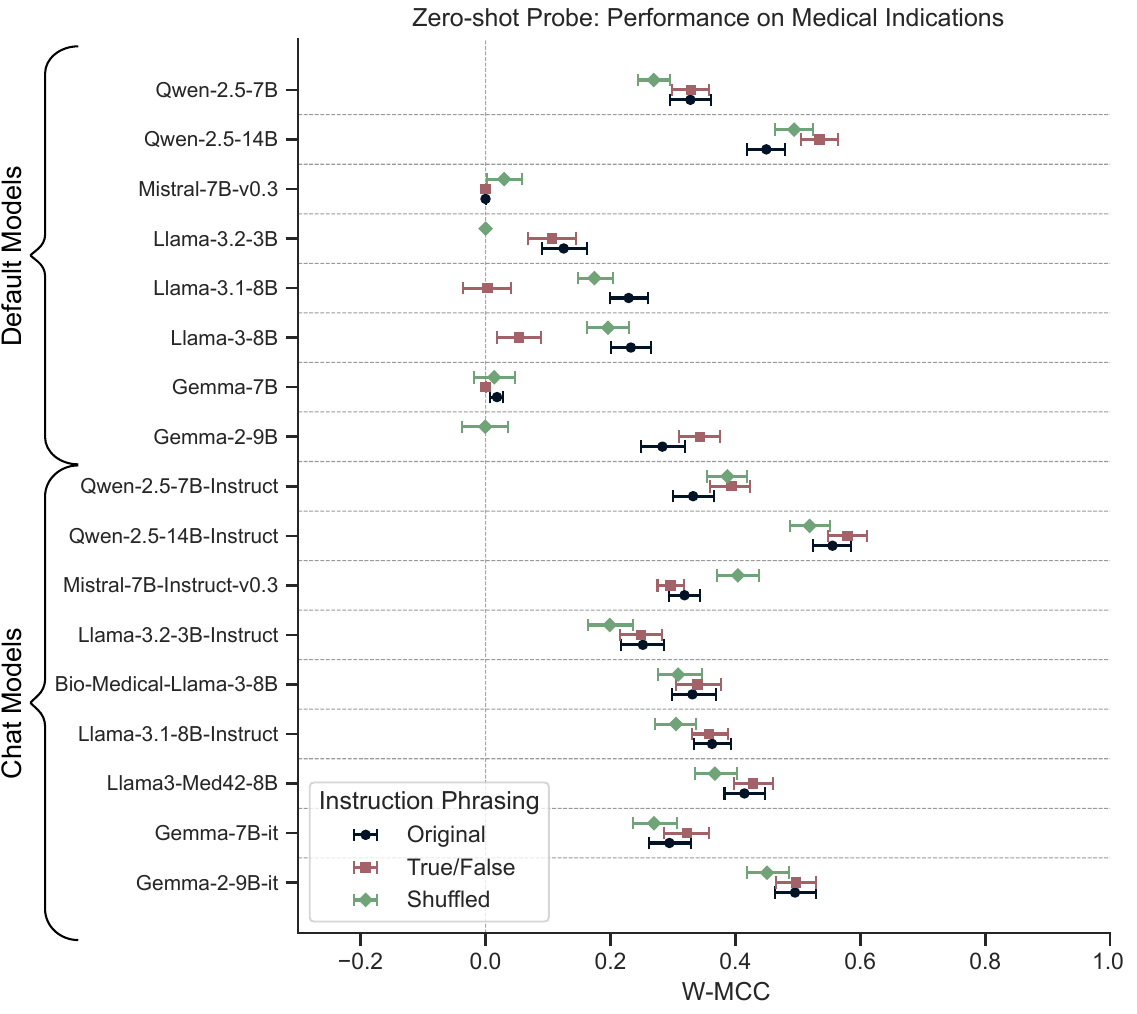}
    \caption{\textbf{Performance of zero-shot prompting on the \textit{Medical Indications} dataset across different models and instruction phrasings.}
    We use the Weighted Matthews Correlation Coefficient (W-MCC) to quantify performance. The marker shows the mean value, and the error bars show the \(95\%\) confidence intervals (based on bootstrapping with \(n=\) 1,000 bootstrap samples).
    Minimal changes to the prompt instructions can skew the performance of zero-shot prompting.
    We observe a slight performance misalignment depending on the phrasing of the instruction.
    The best-performing LLMs are the largest chat models: \texttt{Gemma-2-9b} and \texttt{Qwen-2.5-14b}. 
    We expected the biomedical \texttt{Llama} models to outperform on the medical indications dataset.
    }
\label{fig:app_zero_shot_med_indications}
\end{figure*}

\begin{figure*}[ht]
    \centering
    \includegraphics[width=\linewidth]{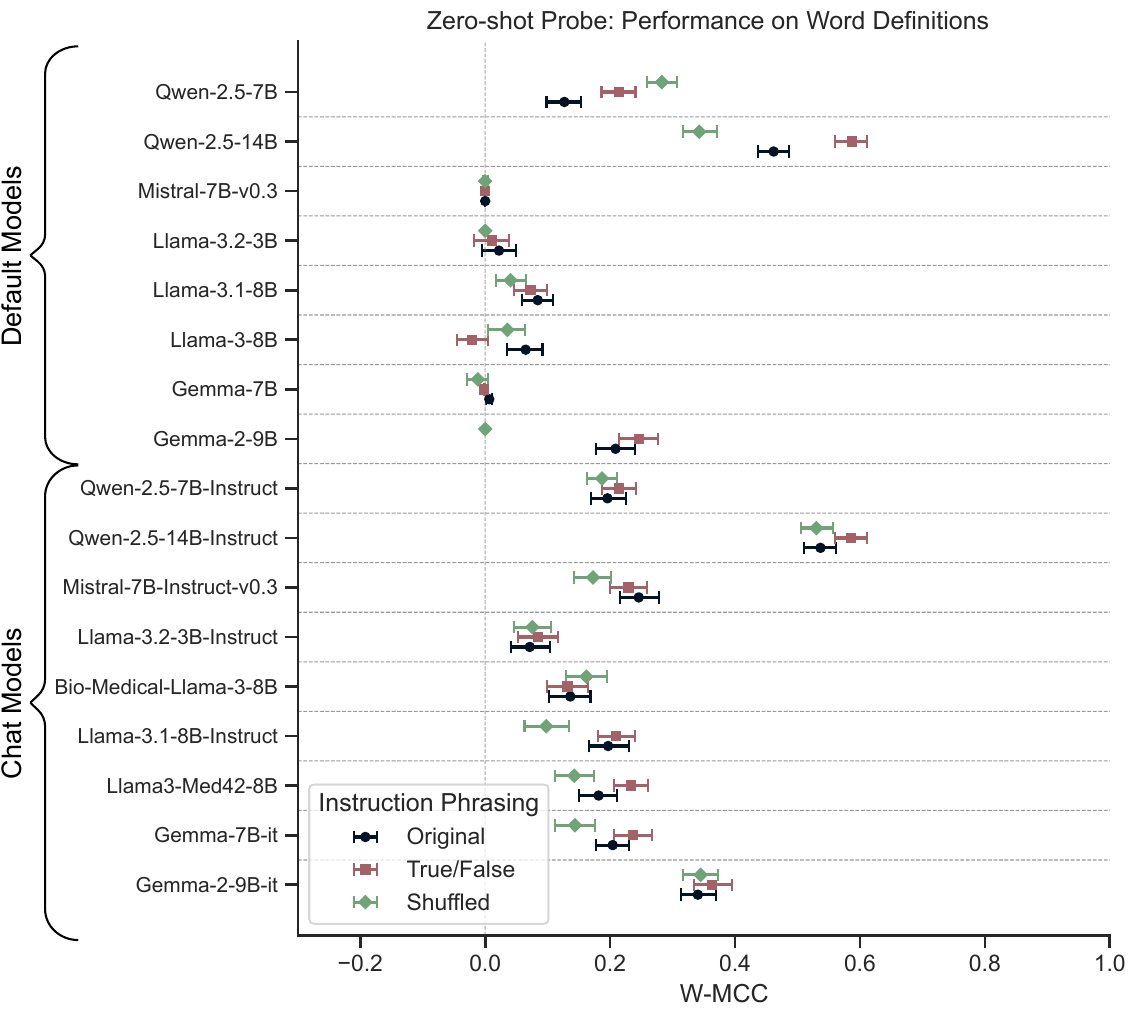}
    \caption{\textbf{Performance of the zero-shot prompting on the \textit{Word Definitions} dataset  across different LLMs and instruction phrasings.} 
    We use the Weighted Matthews Correlation Coefficient (W-MCC) to quantify performance: the marker shows the mean value and the error bars show the \(95\%\) confidence intervals (based on bootstrapping with \(n=\) \(1{,}000\) bootstrap samples).
    Minimal changes to the prompt instructions can skew the performance of zero-shot prompting.
    The overall performance on the \textit{Word Definitions} dataset is much lower than that on the other datasets. 
    Generally, the misalignment in the performance between different instructions is much lower (except for the default \texttt{Qwen} models, where the difference is significant). 
    The largest \texttt{Qwen} models, \texttt{Qwen-2.5-14b} and \texttt{Qwen-2.5-14b-instruct}, are the top-performing models on this task.
    }
    \label{fig:app_zero_shot_defs}
\end{figure*}

\Cref{fig:app_zero_shot_cities,fig:app_zero_shot_med_indications,fig:app_zero_shot_defs} depict the Weighted Matthews Correlation Coefficient (W-MCC) performance of zero-shot prompting\footnote{W-MCC is an MCC score weighted  by the acceptance rate (1-abstention rate).} across three datasets and 16 LLMs for each template type and instruction type.  

In some cases, minor changes to the instructions can significantly affect performance.
Overall, the \texttt{Qwen} and \texttt{Gemma-2} models achieve the best performance among LLMs.
However, \texttt{Qwen} models seem to be more sensitive to the phrasing of the prompt.

\paragraph{From token probabilities to veracity labels in Zero-shot prompting.} 
Given the original instructions and the statement \(\boldsymbol x\), an LLM outputs token-level probabilities over its vocabulary \(\mathcal V\) as
\[
\begin{aligned}
    &P_{\mathcal{M}}(\tau \mid \text{instruction} \wedge \boldsymbol{x}) \\
    &\quad\text{with } \sum_{\tau \in \mathcal{V}} P_{\mathcal{M}}(\tau \mid \text{instruction} \wedge \boldsymbol{x}) = 1
\end{aligned}
\]

We are interested in the probabilities of the tokens that correspond to the multiple choices, i.e., numbers \(1\)--\(6\) in any of the panels in~\Cref{fig:zero-shot-example}.
We denote tokens associated with these numbers as: \texttt{[1]}, \texttt{[2]}, \texttt{[3]}, etc.
We map these token-level probabilities \(P_{\mathcal M}\) into the veracity-label probabilities \(G_{\mathcal M}\) as follows:
\begin{gather*}
    G_{\mathcal M}(\text{true}\mid \boldsymbol{x}) = P_{\mathcal{M}}(\texttt{[1]} \mid \text{instruction} \wedge\boldsymbol x) \\
    G_{\mathcal M}(\text{false}\mid \boldsymbol{x}) = P_{\mathcal{M}}(\texttt{[2]} \mid \text{instruction}\wedge\boldsymbol x) \\
    G_{\mathcal M}(\text{neither}\mid \boldsymbol{x}) = P_{\mathcal{M}}(\texttt{[3]} \mid \text{instruction}\wedge\boldsymbol x) \\ +  P_{\mathcal{M}}(\texttt{[4]} \mid \text{instruction}\wedge\boldsymbol x)
\end{gather*}

\paragraph{Abstention in Zero-shot prompting.}
We include options \texttt{[5]} and \texttt{[6]} to check the ``sanity'' of the model \(\mathcal M\). For example, option \#5 in~\Cref{fig:zero-shot-example}.A suggests that a statement \(\boldsymbol x\) is \textit{true}, \textit{false}, and \textit{neither} -- all at the same time.
If the LLM concentrates most of its probability mass on tokens in \(\mathcal{V}_{\perp} = \mathcal{V} \setminus \{\texttt{[1]},\dots,\texttt{[4]}\}\) (the sanity-check options \(\{\texttt{[5]},\texttt{[6]}\}\) or any token outside the answer set \(\{\texttt{[1]},\dots,\texttt{[4]}\}\)), we assume the LLM did not follow the instructions and \textit{abstains} from prediction:
\begin{multline}
\label{eq:abstain2}
    G_{\mathcal{M}}(\text{abstain} \mid \boldsymbol{x}) = \\ 
    \sum_{\tau \in \mathcal{V}_{\perp}} P_{\mathcal{M}}(\tau \mid \text{ instruction } \wedge \boldsymbol{x})
\end{multline}
Recall that the zero-shot prompting relies solely on the token-level probabilities, i.e., \(\mathcal M \)'s output.
It does not look at the intermediate hidden representation of \(\boldsymbol x\).

\clearpage
\twocolumn
\section{Technical details for \texttt{sAwMIL}}
\label{supsec:sawmil}
Our proposed method, \texttt{sAwMIL} (sparse-aware multiple-instance learning), is a generalization of the standard Single-Instance Support Vector Machine (SVM) to the multiple-instance setting. In this section, we briefly recap the single-instance SVM (SIL-SVM) formulation and describe what changes when we switch to the multiple-instance setting.
To make this section easier to read, \Cref{sup_tab:sawmil_notations} provides an overview of the relevant notations.

\begin{table}[ht]
  \centering
\begin{adjustbox}{width=1\linewidth}
  \begin{tabular}{@{}cl@{}}
    \toprule
    \textbf{Symbol} & \textbf{Description}\\
    \midrule
    \(\boldsymbol{x}\) & Feature vector for a single instance, \(\boldsymbol x  \in \mathbb{R}^{d}\) \\
    \(B\) & Bag consisting of instances \\
    \(\mathcal{X}^+\),     \(\mathcal{X}^-\) & Sets of positive and negative bags \\
    \(\tilde{X}^+\) & Set of instances from all the positive bags   \\
    \(\tilde{X}^-\) & Set of  instances from all the negative bags  \\
   \( \mathcal{D}\) & Dataset \\
   \(\tilde{\mathcal{D}}\) & Relabeled dataset \\
   \(\boldsymbol m\) & Intra-bag labels, \(\boldsymbol m = \{0,1\}^{|B|}\) \\
\bottomrule
\end{tabular}
\end{adjustbox}
\caption{Notations related to the \texttt{sAwMIL} description.}
\label{sup_tab:sawmil_notations}
\end{table}

\subsection{Single-Instance SVM}
\label{supsec:single-instance-svm}
Support Vector Machines are maximum margin classifiers. The general idea behind the single-instance SVM (\texttt{SIL-SVM})\footnote{In most cases, when people mention SVMs, they refer to the single-instance variant.} is to find a separating hyperplane that maximizes the margin between the samples of the two classes.
Thus, in the single-instance setting, our training dataset
\begin{equation}
    \mathcal{D} = \{\langle\boldsymbol{x}_i, y_i\rangle\}_{i=1}^N
\end{equation}
consist of $N$ instance-label pairs, with feature vectors $\boldsymbol{x}_i \in \mathbb{R}^d$ and binary labels $y_i \in \{-1, +1\}$. 

The separating hyperplane is parameterized by a weight vector $\boldsymbol{\theta} \in \mathbb{R}^{1 \times d}$ and bias $\beta \in \mathbb{R}$. The class boundaries are defined by $\boldsymbol{\theta}^{\top} \phi(\boldsymbol{x}) + \beta = \pm 1$, where $\phi(\cdot)$ is a feature transformation. In the case of the linear SVM, $\phi(\cdot)$ is an identity function.

The margin width equals $2/\|\boldsymbol{\theta}\|$, thus, maximizing the margin is equivalent to minimizing $\|\boldsymbol{\theta}\|^2$. At the same time, to allow for misclassifications, we introduce per-instance slack variables $\xi_i \geq 0$, which measure how far each point lies from the correct side of the margin. The resulting (primal) problem is:
\begin{equation}
\begin{aligned}
    \min_{\boldsymbol{\theta},\, \beta,\, \boldsymbol{\xi}} \;\;
    & \frac{\|\boldsymbol{\theta}\|^2}{2} + \frac{C}{N}\sum_{i=1}^N \xi_i \\
    \text{s.t.} \quad 
    & y_i\bigl(\boldsymbol{\theta}^{\top}\phi(\boldsymbol{x}_i) + \beta\bigr) \geq 1 - \xi_i,\\
    & \xi_i \geq 0, \quad \forall\, i
\end{aligned}
\label{eq:sil-svm-primal}
\end{equation}

\noindent where $C > 0$ controls the trade-off between margin maximization and penalizing margin violations\footnote{Larger values penalize margin violations more heavily, yielding a narrower margin but fewer misclassifications on the training set.}. The decision function is $f(\boldsymbol{x}) = \boldsymbol{\theta}^{\top}\phi(\boldsymbol{x}) + \beta$, and the predicted label is  $\hat{y} = \text{sign}(f(\boldsymbol{x}))$.

\subsection{The Multiple-Instance Setting}
In the multiple-instance setting, we no longer have an individual instance per label. Instead, our training data consists of \emph{bags}, such that 
\begin{equation}
    \mathcal D = \{\langle B_i, y_i\rangle\}_{i=1}^N\,.
\end{equation}
Each bag \(B_i\) contains a set \(\{\boldsymbol{x}_i^{(1)}, \dots, \boldsymbol{x}_i^{(L_i)}\}\) of $L_i$ instances that share a single bag-level label $y_i \in \{-1,+1\}$. The important assumptions for the multiple-instance SVMs are as follows:
\squishlisttwo
    \item A \textbf{negative bag} ($y_i = -1$) contains \emph{only} negative instances.
    \item A \textbf{positive bag} ($y_i = +1$) contains \emph{at least one} positive instance.
\squishend

\paragraph{The \texttt{sAwMIL} training consists of three stages.} 
Our framework mirrors the \texttt{sbMIL} framework proposed by \citet{bunescu2007multiple}:
\squishlisttwo
\item We first train a multiple-instance SVM that operates both on bags and instances. \citeauthor{bunescu2007multiple} refer to this SVM as a sparse MIL or \texttt{sMIL}.
\item In the second stage, we use the classifier from Stage 1 to relabel instances in every positive bag.
\item In the third stage, we \enquote{abandon} the bag structure and instead use all the instances together with the new labels from Stage 2. That is, we train a single-instance SVM, refer to~\Cref{supsec:single-instance-svm}.
\squishend

The only difference between \citeauthor{bunescu2007multiple}'s \texttt{sbMIL} and our \texttt{sAwMIL} is how we relabel instances in Stage 2. Below, we provide a detailed description of each stage.

\subsubsection*{Stage 1: The Hybrid Bag--Instance SVM}
Let $\mathcal{X}^-$ denote the set of all negative bags and $\mathcal{X}^+$ the set of all positive bags. Further, we denote
\begin{align}
    \tilde{X}^- &= \bigcup_{B \in \mathcal{X}^-} \{\boldsymbol{x} : \boldsymbol{x} \in B\}, \\
\tilde{X}^+& = \bigcup_{B \in \mathcal{X}^+} \{\boldsymbol{x} : \boldsymbol{x} \in B\}
\end{align}
as the flattened sets of all instances from negative and positive bags, respectively. 
\citeauthor{bunescu2007multiple} design the sparse MIL (\texttt{sMIL}) algorithm specifically for sparse bags. That is, \texttt{sMIL} is suitable for cases in which only a few instances in the positive bags are true positives, while the majority of instances are negative. For that reason, \texttt{sMIL} treats positive and negative bags differently by introducing two types of constraints.

\paragraph{Negative bags: instance-level constraints.} Since every instance in a negative bag is a true negative, we impose a per-instance margin constraint over instances in $\tilde{X}^-$:
\begin{equation}
    \boldsymbol{\theta}^{\top} \phi(\boldsymbol{x}_j) + \beta \;\leq\; -1 + \xi_{\boldsymbol{x}_j}, 
    \, \forall\, \boldsymbol{x}_j \in \tilde{X}^-.
    \label{eq:neg-constraint}
\end{equation}
Each negative instance contributes independently to the optimization, fully exploiting the \textit{unambiguous} negative labels. Note that \cref{eq:neg-constraint} is identical to the negative-class constraints in the single-instance SVM from~\Cref{supsec:single-instance-svm}, i.e., we do not exploit the bag structure.

\paragraph{Positive bags: bag-level constraint.}
For positive bags, we do not know \emph{which} instances are positive, so we cannot impose per-instance constraints. Instead, we impose a single constraint on the representation of each positive bag $B \in \mathcal{X}^+$. Under the assumption that a positive bag has at least one positive instance in $B$ and the remaining $|B|-1$ are likely to be negative, the expected average hidden label is:
\begin{equation}
    \frac{(+1) + (|B|-1)\cdot(-1)}{|B|} \;=\; \frac{2 - |B|}{|B|}.
    \label{eq:positive-bag-scale}
\end{equation}
Using \Cref{eq:positive-bag-scale} as the required margin for the bag mean gives the bag-level constraint
\begin{multline}
        \boldsymbol{\theta}^{\top} \, \frac{\phi(B_j)}{|B_j|} + \beta 
    \;\geq\; \frac{2 - |B_j|}{|B_j|} - \xi_{B_{j}}, \\
    \, \forall\, B_j \in \mathcal{X}^+,
    \label{eq:pos-constraint}
\end{multline}
where $\xi_B \geq 0$ is a bag-level slack and
\begin{equation}
    \phi(B) = \sum_{\boldsymbol{x} \in B} \phi(\boldsymbol{x})
\end{equation}
In the linear case, \(\phi\) is the identity. Note that the right-hand side of \eqref{eq:pos-constraint} becomes more negative as \(|B|\) grows, reflecting that the bag mean is increasingly diluted by negative instances in larger bags, i.e.,  the constraint on the bag mean naturally relaxes accordingly.

Finally, we combine \cref{eq:neg-constraint,eq:pos-constraint} and now solve the following convex problem (can be solved with standard QP solvers):

\begin{equation}
\begin{aligned}
    \min_{\boldsymbol{\theta},\, \beta,\, \boldsymbol{\xi}} \;\;
     \frac{\|\boldsymbol{\theta}\|^2}{2} 
      &+ \frac{C}{|\tilde{X}^-|}\sum_{\boldsymbol{x} \in \tilde{X}^-} \xi_{\boldsymbol{x}} \\
      &+ \frac{C}{|\mathcal{X}^+|} \sum_{B \in \mathcal{X}^+} \xi_B \\
    \text{s.t.} \quad & \text{\cref{eq:neg-constraint,eq:pos-constraint}.}
\end{aligned}
\label{eq:mil-svm-primal}
\end{equation}

\paragraph{Output.} At the end of this stage, we get coefficients \(\boldsymbol{\hat{\theta}} \in \mathbb{R}^{1 \times d}\) and \(\hat{\beta} \in \mathbb{R}\).

\subsubsection*{Stage 2: Pseudo-label assignment}
We use the estimated parameters \(\boldsymbol{\hat{\theta}}\) and \(\hat{\beta}\) to relabel all the instances in the positive bags. That is, for each instance $\boldsymbol{x}_j \in \tilde{X}^+$, we compute a score \(s_j\) according to~\cref{eq:score-positives}:
\begin{equation}
    s_j = \boldsymbol{\hat{\theta}}^{\top} \phi(\boldsymbol{x}_j) + \hat{\beta}, 
    \label{eq:score-positives}
\end{equation}
that further forms
\begin{equation}
    S^+ = \{ s_j \mid \boldsymbol{x}_j \in \tilde{X}^+\}.
\end{equation}
Further, we find the $(1-\eta)$-quantile of the score distribution \(S^+\).
Here, \(\eta\) represents the expected fraction of the positive instances across positive bags. Formally,
\begin{equation}
    q = Q_{1-\eta}(S^+) \, .
\end{equation}
Recall that in our experiments, the statements decompose into pre-actualized and actualized parts: \(\boldsymbol x \leftarrow [ \boldsymbol x^p, \boldsymbol x^a]\). That is, we know where the veracity signal is more likely to be. Thus, in our experiments, each bag \(B\) has a bag label \(y\) \textbf{and} intra-bag labels \(\boldsymbol{m}\). Hence, the dataset is
\begin{equation}
    \mathcal D = \{\langle B_i, y_i, \boldsymbol{m}_i\rangle\}_{i=1}^N\,.
\end{equation}
These intra-bag labels \(\boldsymbol m = \{0,1\}^{|B|}\) specify what instances within the bag are likely to contain the veracity signal and which are not. 

Using these two pieces of information, we can now relabel instances in every positive bag:
\begin{equation}
    \begin{aligned}
    \tilde{y}_j &= \begin{cases}
      +1 & \text{if } s_j \geq q \text{ and } m_j = 1 \\
      -1 & \text{otherwise},
    \end{cases} \\[1pt]
    &\phantom{{}={}} \forall\, \boldsymbol{x}_j \in \tilde{X}^+
    \end{aligned}
\end{equation}

\paragraph{Output.} The result of this stage is the relabeled dataset \(\tilde{\mathcal{D}}\) consisting of instance--pseudolabel pairs:
\begin{equation}
    \tilde{\mathcal{D}} = 
    \{\langle \boldsymbol{x}_j,\, \tilde{y}_j \rangle \mid \boldsymbol{x}_j \in \tilde{X}^+ \cup \tilde{X}^- \},
    \label{eq:relabeled-dataset}
\end{equation}
where $\tilde{y}_j = -1$ for all $\boldsymbol{x}_j \in \tilde{X}^-$.

\subsubsection*{Stage 3: Instance-Level SVM (SIL)}
In the final stage, we discard the bag structure and instead use the relabeled dataset:
\begin{equation}
    \tilde{\mathcal{D}} = \{\langle\boldsymbol{x}_i, \tilde{y}_i\rangle\}_{i=1}^N.
\end{equation}
Here, we fit the single-instance SVM to get the final \(\boldsymbol \theta\) and \(\beta\). The resulting decision function 
\begin{equation}
    f(\boldsymbol{x}) = \boldsymbol{\theta}^{\top}\phi(\boldsymbol{x}) + \beta
\end{equation}
produces per-instance scores.  We provide a pseudocode in~\Cref{alg:sawmil}. Further, we can pass these scores to the conformal prediction algorithm (\Cref{supsec:nonconformity}).

\subsection{Dual Formulation}
In this section, we define SVM algorithms in terms of the primal formulation. However, one can convert the primal problem in~\Cref{eq:sil-svm-primal,eq:mil-svm-primal} into an equivalent dual formulation via Lagrangian multipliers; refer to~\citet{bunescu2007multiple} for details.

\clearpage
\twocolumn
\section{Additional Evaluation Details: Correlation, Selectivity and Generalization}
\label{supsec:evaluation}
\Cref{sec:sup_validity} provides the full list of the validity criteria.
Here, we provide additional details behind the evaluation of the \textbf{Correlation}, \textbf{Selectivity} and \textbf{Generalization}, not covered in~\Cref{sec:results}.

\subsection{Evaluation Setup}
\label{sec:sup_performance}
Here, we describe a pipeline to evaluate our \texttt{sAwMIL} probe over the validity criteria specified in~\Cref{sec:method} and~\Cref{sec:sup_validity}.

\paragraph{Correlation and Selectivity.}
We use the test split of each dataset to evaluate the performance of the probe.
We use Matthews Correlation Coefficient (MCC) to summarize the statistical accuracy of probes.
The multiclass MCC value is calculated using~\Cref{eq:sup_mcc}:
\begin{equation}
\label{eq:sup_mcc}
\operatorname{MCC}=\frac{c\,s-\sum_{k} p_k t_k}{\sqrt{\left(s^2-\sum_{k} p_k^2\right)\left(s^2-\sum_{k} t_k^2\right)}}
\end{equation}
where \(c\) is the number of correct predictions, \(s\) is the total number of samples, \(K\) is the total number of classes, \(t_k\) is the number of \(k\)-class samples in the dataset,  and \(p_k\) is the number of times \(k\)-class was predicted.
\(\operatorname{MCC}=1\) indicates that a classifier predicted every instance correctly. \(\operatorname{MCC}=0\) implies that the predictions are random. 
\(\operatorname{MCC}=-1\) indicates that the predictions are inversely correlated with the ground-truth labels.

Since \textit{neither} statements are included in the test dataset, MCC provides a sense of how well the probe classifies factually \textit{true} or \textit{false} statements and indicates whether the probes can handle \textit{neither}-type cases.

\paragraph{Generalization.}
To test how well a particular probe \(g_i\) trained on a dataset \(\mathcal{D}_i\) generalizes, we evaluate its performance using the test split of another dataset \(\mathcal{D}_j\). That is, \(g_i\) trained on the city locations dataset is evaluated using the test split of the word definitions dataset.

\subsection{Generalization Across Datasets}

\begin{figure*}[h]
\centering
  \includegraphics[width=\linewidth]{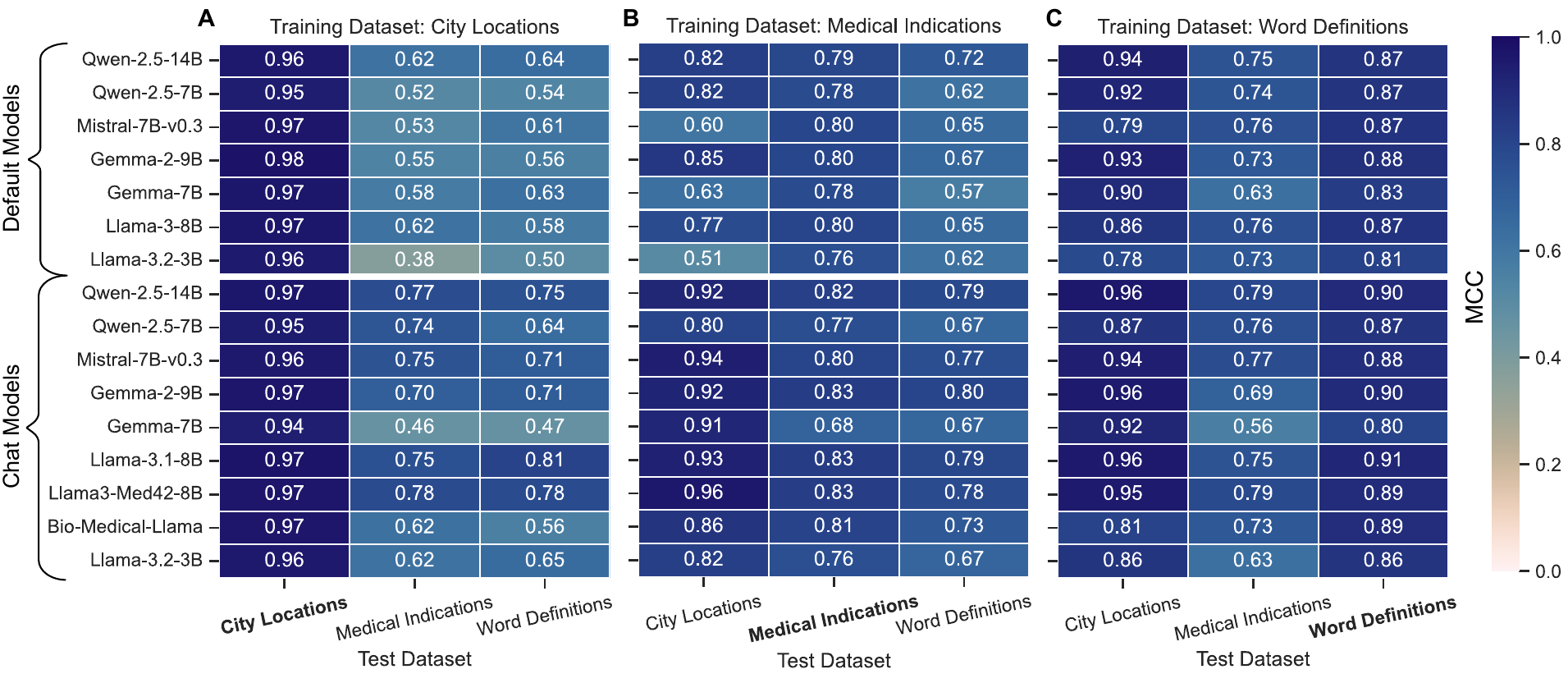}
\caption{\textbf{Generalization performance of the multiclass \texttt{sAwMIL} probe across datasets.}
Each panel corresponds to a different \textit{training} dataset: \textit{City Locations}, \textit{Medical Indications}, and \textit{Word Definitions}. Each column corresponds to a different test dataset.  Each cell displays the MCC value, which quantifies how well the probe generalizes to the test dataset (higher values are better). For each model and dataset pair, we report the maximum MCC achieved across all decoders. In general, probes trained on chat models exhibit better generalization performance than the default models.
\textbf{Panel A:} Generalization performance of multiclass \texttt{sAwMIL} when trained on \textit{City Locations}. 
While the MCC values are significantly higher than a random baseline (with MCC = $0$), the generalization ability is lower than that in Panels B or C.
\textbf{Panel B:} Generalization performance of multiclass \texttt{sAwMIL} when trained on  \textit{Medical Indications}. In Panel A, we observe that training on \textit{City Locations} and testing on \textit{Medical Indications} provides good but not excellent MCC values (average MCC of $0.624$ with standard error of $0.030$). This is not the case in this panel, where \textit{Medical Indications} is the training dataset and  \textit{City Locations} is the test dataset (average MCC of $0.818$ with standard error of $0.033$).
\textbf{Panel C:} Generalization performance of multiclass \texttt{sAwMIL} when trained on \textit{Word Definitions}. This probe has high generalization performance across datasets. When \textit{City Locations} is the test dataset, the average MCC is $0.896$ with a standard error of $0.015$; and when \textit{Medical Indications} is the test dataset, the average MCC is $0.723$ with a standard error of $0.016$.  For aggregated statistics, see~\Cref{tab:general_detailed_averages}.
}
  \label{fig:generalization_performance}
\end{figure*}

To further support the claim that the multiclass \texttt{sAwMIL} captures veracity signals (and not merely a proxy), we demonstrate generalization performance across datasets. \Cref{fig:generalization_performance} provides results for each dataset and LLM.
The columns correspond to three test datasets, and the cells specify the multiclass \texttt{sAwMIL}'s performance for a specific LLM. 

Multiclass \texttt{sAwMIL} provides reasonable generalization performance (see~\Cref{tab:general_detailed_averages}). However, it may overfit to the highly specialized \textit{City Locations} dataset.
Therefore, using more diverse datasets that span a broader range of entities and cover a larger set of topics better isolates the veracity signal and yields better generalization performance. We refer the reader to~\Cref{tab:sup_generalization_cities} through~\Cref{tab:sup_generalization_defs} for detailed statistics. 

In~\Cref{tab:general_detailed_averages}, the values reported are  for \texttt{sAwMIL} computed without conformal prediction, i.e., over all test statements. Averaging the off-diagonal cells of this table therefore gives \(0.63\), \(0.76\), and \(0.81\) for probes trained on City Locations, Medical Indications, and Word Definitions. In~\Cref{sup_tab:aggregated_generalization}, we report \(0.82\), \(0.88\), and \(0.86\) for the \texttt{sAwMIL} with conformal prediction, where MCC is measured only on non-abstained predictions.

\begin{table*}[ht]
  \centering
  \begin{tabular}{lccc}
    \toprule
    & \multicolumn{3}{c}{\textbf{Testing Dataset}} \\
    \cmidrule(lr){2-4}
    \textbf{Training Dataset}
      & City Locations & Medical Indications & Word Definitions \\
    \midrule
    City Locations       & 0.963 (0.003) & 0.624 (0.030) & 0.633 (0.025) \\
    Medical Indications  & 0.818 (0.033) & 0.790 (0.009) & 0.698 (0.018) \\
    Word Definitions     & 0.896 (0.015) & 0.723 (0.016) & 0.868 (0.008) \\
    \bottomrule
  \end{tabular}
  \caption{\textbf{Aggregated generalization performance of the multiclass \texttt{sAwMIL} for each dataset} (the values reported here are computed without conformal prediction, i.e., over all test statements, including the ones where \texttt{sAwMIL} would abstain). Each cell shows an MCC value, which quantifies the performance of the multiclass \texttt{sAwMIL} trained and tested on different combinations of the datasets. The value in the parentheses is the standard error. \textit{Word Definitions} dataset provides better generalization performance because it contains statements covering a diverse set of topics, while the \textit{City Locations} dataset provides lower generalization performance.}
  \label{tab:general_detailed_averages}
\end{table*}

We also observe that generalization performance is higher for chat models, where the average MCC score (on the non-training dataset) is \(77.2\%\) (standard error: \(0.2\%\)), versus \(68.2\%\) (standard error: \(0.2\%\)) for the default models.
This is more noticeable in~\Cref{fig:generalization_performance}.A, where default models have much lower MCC values than their chat model counterparts.
For example, the chat \texttt{Llama-3.2-3B} has an MCC value \(1.6\) times as high as the default \texttt{Llama-3.2-3B}.

Across the three panels, \texttt{Gemma-7B} seems to be an outlier, as its generalization performance drops significantly for the chat version of the model.

Multiclass \texttt{sAwMIL} satisfies the generalization criterion defined in~\Cref{sec:sup_validity}~by transferring veracity probes trained on one dataset to another while maintaining strong performance. This provides further evidence that multiclass \texttt{sAwMIL} captures a veracity signal that is not dataset-specific.

\subsection{Recap: Overall Validity}

In this section and~\Cref{sec:results} of the manuscript, we established that the multiclass \texttt{sAwMIL} probe satisfies the validity criteria. Specifically, we confirmed that it satisfies the correlation and selectivity criteria and outperforms zero-shot prompting and the mean-difference probe using conformal prediction sets. We further demonstrated that multiclass \texttt{sAwMIL} satisfies the generalization criterion, indicating that we can successfully apply probes trained on multiclass \texttt{sAwMIL} to statements from other domains. 

In addition, we showed that the one-vs-all \texttt{sAwMIL} probes satisfy the manipulation and locality criteria (\Cref{supsec:interventions}). In most cases, we can intervene to change the log-probabilities of correct responses. 
Together, these findings provide strong evidence for the overall validity of  \texttt{sAwMIL} probes. \underline{Not} all LLMs have a veracity mechanism that has a linear relationship with the output. Exploring this non-linear relationship is part of our future work.
\clearpage
\twocolumn
\section{Additional Evaluation Details: Causal Validation via Steering}
\label{supsec:interventions}
Our causal validation follows the directional intervention paradigm \cite{li2023inference, todd2024function}, where hidden activations are translated along a learned direction vector \(\boldsymbol \theta_i\). Unlike standard activation patching~\cite{meng2022locating}\footnote{In activation patching, one creates a pair of contrastive prompts (one that elicits the target behavior and one that does not). Activations from specific components (e.g., a feed-forward layer or attention head) of the clean run are then swapped into the corrupted run to determine whether that component is causally responsible for the behavioral difference.}, which swaps activations between specific prompts to localize \emph{where} knowledge is stored, our intervention tests \emph{whether} a learned veracity direction causally influences model \(\mathcal{M}\)'s outputs. To identify this causal effect, we propose \textbf{Contrastive Token Intervention} (CTI) to estimate the causal effect of learned veracity directions on LLM outputs, i.e., the difference-in-differences design.

\paragraph{Activation Steering.} Let \(\boldsymbol{x} = [\boldsymbol{x}^p, \boldsymbol{x}^a]\) be a tokenized statement consisting of a pre-actualized part \(\boldsymbol{x}^p\) and an actualized part \(\boldsymbol{x}^a\). Let \(\boldsymbol{\tau}_c\) denote the correct actualized token sequence (in this case, \(\boldsymbol x^a = \boldsymbol \tau_c\))  and $\boldsymbol{\tau}_r$ a random control token sequence (in this case, \(\boldsymbol x^a = \boldsymbol \tau_r\) and \(| \boldsymbol \tau_c| = |\boldsymbol \tau_r|\)). Let $\boldsymbol \theta_i$ denote the learned veracity direction at layer \(i\), let $\alpha$ denote the intervention dose, and \(\kappa\) indicate the sign of the translation.

We intervene by translating the hidden representation of the \underline{last token} of \(\boldsymbol{x}^p\) at layer \(i\) along \(\boldsymbol \theta_i\) (representations of other tokens stay the same):
\begin{equation}
\label{eq:intervention}
    \tilde{h}^{\kappa}_i(\boldsymbol{x}^p) \leftarrow 
    h_i(\boldsymbol{x}^p) + \kappa \cdot \alpha \cdot 
    \boldsymbol \theta_i, 
\end{equation}
where \(\kappa \in \{-1, +1\}\), \(\tilde{h}^{\kappa}_i(\boldsymbol{x}^p) \in \mathbb{R}^{|\boldsymbol x^p| \times d}\) and \(h_i(\boldsymbol{x}^p) \in \mathbb{R}^{|\boldsymbol x^p| \times d}\).
After we perform the intervention, we measure the 
change in log-probability of a target sequence $\boldsymbol{\tau}$ 
relative to the unperturbed model:
\begin{equation}
\label{eq:delta}
\begin{aligned}
    \Delta_{\boldsymbol{\tau}}^{\kappa} =& 
    \log P_{\mathcal{M}}\!\left(\boldsymbol{\tau} \mid \tilde{h}_i^{\kappa}\left(\boldsymbol{x}^p\right)\right) \\
    &- \log P_{\mathcal{M}}\!\left(\boldsymbol{\tau} \mid h_i\left(\boldsymbol{x}^p\right)\right)
\end{aligned}
\end{equation}
Evaluating~\Cref{eq:delta} for each combination of target sequence 
$\boldsymbol{\tau} \in \{\boldsymbol{\tau}_c, \boldsymbol{\tau}_r\}$ 
and translation sign $\kappa \in \{-1, +1\}$ yields four values per 
statement:
\begin{table}[t]
\centering
\begin{tabular}{lcc}
\toprule
 & ($\kappa=+1$) & ($\kappa=-1$) \\
\midrule
Correct ($\boldsymbol{\tau}_c$) & $\Delta_{\boldsymbol{\tau}_c}^{+}$ 
& $\Delta_{\boldsymbol{\tau}_c}^{-}$ \\
Random ($\boldsymbol{\tau}_r$)  & $\Delta_{\boldsymbol{\tau}_r}^{+}$ 
& $\Delta_{\boldsymbol{\tau}_r}^{-}$ \\
\bottomrule
\end{tabular}
    \caption{Outcome variables used in the difference-in-differences design. Each cell shows the expected log-probability shift $\Delta$ under positive ($\kappa=+1$) and negative ($\kappa=-1$) interventions, for the correct tokens $\boldsymbol{\tau}_c$ and random baseline tokens $\boldsymbol{\tau}_r$. We plug these terms into~\Cref{eq:sup_did} as the outcome variable.}
    \label{tab:did-params}
\end{table}

\paragraph{Intervention Dosage.} 
Our analysis shows that fixing \(\alpha\) at a single value across probes yields intervention effects with substantially different magnitudes. 
This happens because each probe produces scores that have a different scale: for a fixed \(\alpha = 1\), a probe with high variance (of output scores) produces a larger perturbation (translation) than one with low variance. 
To account for this, we calibrate \(\alpha\) for each \(\langle\)probe, LLM, dataset, layer\(\rangle\). That is, we take each \(g_i\) and compute the distribution of scores on the calibration data, \(\mathcal D_{cal}\), and we then estimate the standard deviation of the score distribution~\(\sigma\). The resulting intervention dosage is computed based on~\Cref{eq:dosage}:
\begin{equation} 
    \alpha = d' \cdot \sigma, \quad d' \in \{1, 3\},
    \label{eq:dosage} 
\end{equation} 
where \(d\) controls the intervention strength in units of \(\sigma\).

\paragraph{Difference-in-Differences.} To isolate veracity-specific effects from overall distributional shifts, we employ a \textbf{Difference-in-Differences} analysis.
To reiterate, for each statement, we measure changes in log-probability under positive and negative translations for both the correct-answer tokens and random control tokens. The interaction term \(\beta_3\) from the resulting 2×2 factorial model captures the differential causal effect (\Cref{tab:did-params}):
how much more the intervention shifts the log-probability of correct tokens than of random tokens.
\begin{equation}
\begin{aligned}
    \Delta_\tau^{\kappa} = 
    \beta_0 
    &+ \beta_1 \cdot \mathds{1}[\boldsymbol{\tau}=\boldsymbol{\tau}_c]\\
    &+ \beta_2 \cdot \mathds{1}[\kappa=+] \\
    &+ \beta_3 \cdot \mathds{1}[\boldsymbol{\tau}=\boldsymbol{\tau}_c] \cdot \mathds{1}[\kappa=+] \\
    &+ \varepsilon
\end{aligned}
\label{eq:sup_did}
\end{equation}
In \Cref{eq:sup_did}, standard errors are clustered by statement to account for within-statement correlation across the four cells of the \(2\times2\) model. This design controls for the possibility that a direction merely shifts the entire output distribution without specifically targeting veracity-relevant tokens.
Here we provide the interpretation of each coefficient:
\begin{description}[labelindent=2em]
    \item[\(\beta_1\)]: The \textit{baseline advantage} of correct tokens over random tokens before any shift. A positive \(\beta_1\) indicates that correct tokens have a higher probability than random tokens.
    \item[\(\beta_2\)]: The \textit{overall distribution shift} introduced by the intervention. That is, what is the average effect of the intervention?
    \item[\(\beta_3\)]: The \textit{veracity-specific shift}, aka what is the additional log-probability shift on the correct tokens induced by the intervention.
\end{description}
One can think (abstractly) of \(\beta_3\) as the veracity-specific effect of the intervention, net of the generic distributional shift \(\beta_2\) induced by translating along \(\alpha \cdot \kappa \cdot \boldsymbol{\theta}_i\) as
\begin{equation*}
    \beta_3=\underbrace{\left(\Delta_{\tau_c}^{+}-\Delta_{\tau_c}^{-}\right)}_{\text {effect on correct token }}-\underbrace{\left(\Delta_{\tau_r}^{+}-\Delta_{\tau_r}^{-}\right)}_{\text {effect on random token }}.
\end{equation*}
Therefore, \cref{eq:sup_did} allows us to test the following hypothesis:
\begin{description}[labelindent=2em]
    \item[$H_0$:] \(\beta_3 = 0\): the veracity direction shifts all tokens equally (both \(\boldsymbol \tau_r\) and \(\boldsymbol \tau_c \)).
    \item[$H_1$:] \(\beta_3 \neq 0\): the veracity direction \(\boldsymbol \theta_i\) 
    \textit{specifically} targets the correct-answer token $\boldsymbol \tau_c$.
\end{description}
\paragraph{Cross-comparison.}
While \textbf{intervention dosage} normalization controls for scale differences across $\langle$probe, LLM, dataset, layer$\rangle$ configurations, $\beta_3$ remains correlated with $\sigma$, indicating the interaction effect is not fully scale-independent. To allow unbiased cross-configuration comparisons, we report two complementary metrics:
\squishlisttwo
    \item (Primary) \textbf{Normalized Interaction Effect} in \cref{eq:sup_normalized_interaction}: To summarize the magnitude of the causal effect in a scale-independent manner, we additionally report the normalized interaction effect, pooled across configurations via the DerSimonian--Laird estimator~\cite{kelley2023evolution} described below.
    \item (Secondary) \textbf{Success Rate}: the fraction of $\langle$LLM, dataset$\rangle$ configurations yielding a statistically significant $\beta_3$ after correction for the false discovery rate (per probe).\footnote{We apply the~\citet{benjamini1995controlling} procedure to control the false discovery rate.} This measures how reliably a probe identifies a veracity-specific direction across configurations.
\squishend

\begin{equation}
    \text{Normalized Interaction Effect} = \left| \frac{\beta_3}{\hat{\sigma}} \right|
    \label{eq:sup_normalized_interaction}
\end{equation}
In~\Cref{eq:sup_normalized_interaction}, \(\hat{\sigma}\) is the residual standard deviation of the difference-in-differences fit.

\paragraph{Pooling Across Configurations.}
Even after dosage calibration, $\beta_3$ varies across 
$\langle$probe, LLM, dataset$\rangle$ configurations due to between-configuration heterogeneity. 
That is, different LLMs encode veracity at different layers and with different magnitudes. 
To obtain a single, robust summary per probe, we pool the normalized interaction effects across configurations using the DerSimonian--Laird random-effects estimator~\cite{kelley2023evolution}.

\subsection{Setup}

\paragraph{Collecting results.} For each \(\langle\)~probe, LLM, dataset~\(\rangle\) configuration, we collect the intervention results from the five best-performing layers (according to the MCC values) and fit the difference-in-differences model.  We then select the layer with the median \(\beta_3\). 
This layer is then considered to be the representative of the \(\langle\)~probe, LLM, dataset~\(\rangle\) configuration. 
In total, we have 48 configurations.

\noindent\textbf{Why?} (1) We want to make our analysis robust to the outliers and quantify the effect across \enquote{good} layers; (2) In some cases, the best-performing layer is the last one, and when we apply interventions to the last layer, the \texttt{PyTorch}~\cite{pytorch} package produces \texttt{NaN} values. For instance, that happens with the \texttt{TTPD} probe.

\paragraph{Comparison setting.}
We apply CTI to four probes: the mean-difference (\texttt{MD}), Training of Truth and Polarity Direction (\texttt{TTPD}), SVM, and 
\texttt{sAwMIL} --- the latter two using only the \texttt{is-true} direction. We exclude the supervised PCA (\texttt{sPCA}) probe, as it does not yield a direction in the same representation space. Hence, our goal is to test whether the truthfulness directions identified by each probe causally influence \(\mathcal{M}\) outputs, i.e., whether they can be used to steer LLM behavior toward or away from correct answers.

\subsection{Results}
\begin{figure*}[ht]
\centering
  \includegraphics[width=0.8\linewidth]{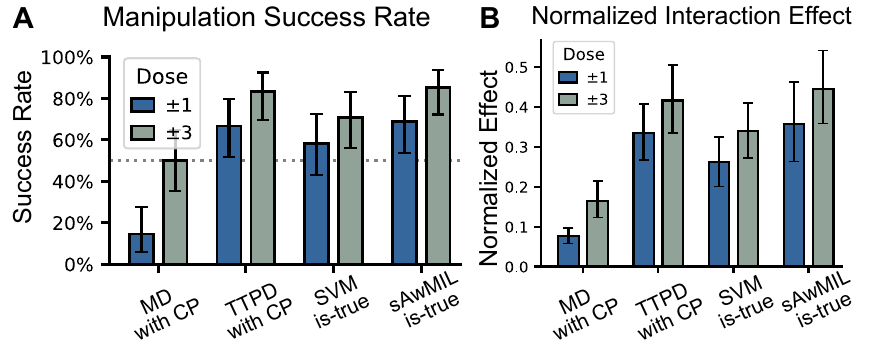}
\caption{Manipulation + locality evaluation pooled over all \(48\) \(\langle\)LLM, dataset\(\rangle\) configurations per each probe. Dose controls the magnitude of the interventions.
\textbf{Panel A:} Intervention success rate per probe across all configurations (total: \(48\)), with binomial proportion confidence intervals, i.e., fraction of configurations with significant \(\beta_3\). Dose controls the magnitude of the translation along the learned veracity direction \(\boldsymbol \theta_i\). A higher success rate indicates that activation steering along the veracity direction produced statistically significant changes in LLM output across more configurations. 
\textbf{Panel B}: Pooled normalized interaction effect, aka normalized \(\beta_3\) adjusted via DerSimonian–Laird random-effects estimator~\cite{kelley2023evolution}.
The success rate denotes the number of interventions with successful manipulation \& locality, whereas the normalized effect size quantifies the magnitude of the effect.
\texttt{MD} has the weakest causal effect, followed by the single-instance \texttt{SVM}. \texttt{TTPD} and the \texttt{sAwMIL}'s \texttt{is-true} probe achieve the highest success rates and similar effect sizes. These results are also displayed in~\Cref{tab:did-pooled}.}
  \label{fig:bar_intervention}
\end{figure*}

To assess whether the learned veracity directions have a causal effect on LLM outputs, we look at the fraction of successful interventions for each probe across all \(\langle\)\texttt{LLM}, \texttt{dataset}\(\rangle\) configurations (\Cref{fig:bar_intervention}.A), as well as the pooled normalized interaction effect in \Cref{fig:bar_intervention}.B; note that the pooled NIE is adjusted via the DerSimonian–Laird
random-effects estimator~\cite{kelley2023evolution}. We also provide these results in~\Cref{tab:did-pooled}.

Success indicates that the translation by the veracity direction (\(\alpha \cdot \kappa \cdot \boldsymbol{\theta}_i\)) yields a statistically significant change in the log-probability of the correct tokens as opposed to the irrelevant tokens (i.e., statistically significant \(\beta_3\)).
The \texttt{MD} probe achieves the lowest success rate, indicating that the direction it identifies has little to no influence on majority of \(\langle\)\texttt{LLM}, \texttt{dataset}\(\rangle\) configurations, while the single-instance \texttt{SVM} yields higher success. The \texttt{sAwMIL} and \texttt{TTPD} achieve the highest success rates, suggesting that both probes reliably identify directions the LLM actively uses when processing veracity-relevant information.

The pooled NIE in~\Cref{fig:bar_intervention} shows a similar trend, where \texttt{sAwMIL} and \texttt{TTPD} yield the highest intervention effects. That is, not only do these probes reliably find the veracity direction, but these directions are also used by the LLMs.

Meanwhile,~\Cref{tab:did-pooled} provides the between-configuration heterogeneity estimates \(\tau^2\). The higher this value is, the greater the variations in effect sizes across \(\langle\)\texttt{LLM}, \texttt{dataset}\(\rangle\) configurations. Notably, \texttt{MD} has the lowest \(\tau^2\) value: the probe consistently finds directions with a weak causal effect. However, \texttt{TTPD} and \texttt{sAwMIL} both have higher \(\tau^2\) values: in some configurations, they identify directions with a high causal effect, whereas in others, they identify a direction with a smaller effect.

Together with findings related to correlation and generalization in~\Cref{sec:results}, we see that \texttt{sAwMIL} yields better performance and also finds a direction with causal connections to LLM outputs.
At the same time, \texttt{TTPD} seems to identify reliable veracity directions but is confounded by noise (since its performance and generalization do not hold as well). 

\begin{table*}[ht]
\centering
\begin{tabular}{@{}lccccccc@{}}
\toprule
& & \multicolumn{3}{c}{\textbf{Normalized Interaction Effect}} & \multicolumn{3}{c}{\textbf{Success Rate}} \\
\cmidrule(lr){3-5} \cmidrule(lr){6-8}
\textbf{Probe} & \textbf{Dose} & NIE& 95\% CI & $\tau^2$ & SR & 95\% CI & $k/n$ \\
\midrule
\multirow{2}{*}{\texttt{MD}} 
    & $1$ & 0.068 & (0.050,\ 0.086) & 0.0017 & 0.146 & (0.061,\ 0.278) & 7/48 \\
    & $3$ & 0.155 & (0.116,\ 0.193) & 0.0152 & 0.500 & (0.352,\ 0.648) & 24/48 \\
\midrule
\multirow{2}{*}{\texttt{TTPD}}
    & $1$ & 0.329 & (0.261,\ 0.397) & 0.0517 & 0.667 & (0.516,\ 0.796) & 32/48 \\
    & $3$ & 0.411 & (0.332,\ 0.491) & 0.0735 & 0.833 & (0.698,\ 0.925) & 40/48 \\
\midrule
\multirow{2}{*}{\texttt{SVM} (\texttt{is-true}) }
    & $1$ & 0.253 & (0.198,\ 0.309) & 0.0334 & 0.583 & (0.432,\ 0.724) & 28/48 \\
    & $3$ & 0.334 & (0.267,\ 0.402) & 0.0511 & 0.708 & (0.559,\ 0.830) & 34/48 \\
\midrule
\multirow{2}{*}{\texttt{sAwMIL} (\texttt{is-true})}
    & $1$ & 0.350 & (0.271,\ 0.428) & 0.0722 & 0.688 & (0.537,\ 0.813) & 33/48 \\
    & $3$ & 0.439 & (0.357,\ 0.520) & 0.0768 & 0.854 & (0.722,\ 0.939) & 41/48 \\
\bottomrule
\end{tabular}
\caption{Evaluation of the manipulation and locality criteria: normalized interaction effect and success rate per probe and intervention dose (across all 48 \(\langle\)\texttt{LLM}, \texttt{dataset}\(\rangle\) configurations). 
Pooled \textbf{normalized interaction effect} (NIE) is estimated via the DerSimonian--Laird random-effects estimator~\cite{kelley2023evolution}. Here, \(\tau^2\) denotes between-configuration heterogeneity. \textbf{Success rate} (SR) is the fraction of configurations with statistically significant $\beta_3$ after BH correction~\cite{benjamini1995controlling}, with $k$ successes out of $n=48$ configurations and binomial proportion 95\% CIs~\cite{wilson1927probable}.}
\label{tab:did-pooled}
\end{table*}

\clearpage
\twocolumn
\section{Algorithms}
\label{supsec:algorithms}
In this section, we provide pseudocodes for several procedures described in the main text.
In~\Cref{alg:sawmil}, we provide pseudocode for the Sparse-Aware Multiple-Instance Learning (\texttt{sAwMIL}) probe. In~\Cref{alg:mean_diff,alg:ttpd,alg:spca}, we show pseudocodes for the mean-difference (\texttt{MD}), the training of truth and polarity directions (\texttt{TTPD}), and the supervised PCA (\texttt{sPCA}) probes, respectively.
In~\Cref{alg:sup_cp_binary,alg:sup_cp_multiclass}, we describe the procedure for the binary and multiclass conformal learning (described in~\Cref{sec:conformal-predictions}).

\onecolumn
\begin{algorithm}[ht]
	\caption{Training a one-vs-all \texttt{sAwMIL} classifier}
	\label{alg:sawmil}
	\renewcommand{\algorithmicrequire}{\textbf{Input:}}
	\renewcommand{\algorithmicensure}{\textbf{Output:}}
\begin{algorithmic}[1]
    \Statex \Require A training dataset \(\{(\boldsymbol{x}_i, y_i, \boldsymbol{m}_i)\}_{i=1}^n\) with binary bag labels \(y_i \in \{0,1\}\), bags \( \boldsymbol{x}_i \in \mathbb{R}^{L_i \times d}\), and intra-bag confidences \(\boldsymbol{m}_i \in \{0,1\}^{L_i}\), where \(L_i\) is the number of items in a bag \(\boldsymbol{x}_i\); also, a balancing parameter \(\eta \in (0,1]\).    
    \Statex \Ensure Parameters  \( \boldsymbol{\theta} \in \mathbb{R}^{1 \times d}\) and \(\beta \in \mathbb{R} \).
    These parameters are subsequently given to a function \(f\) along with \(\boldsymbol z\) to compute \( f(\boldsymbol{z}) = \sigma\big(\boldsymbol{z} \,  \boldsymbol{\theta}^{\top} + \beta)\), where  \(\sigma\) is a sigmoid function.
\vspace*{12pt}
    \State Partition data into positive and negative sets:
        \[
        \mathcal{X}^{+} = \{ \langle\boldsymbol{x}_i, \boldsymbol{m}_i \rangle : y_i = 1 \}, \quad \mathcal{X}^{-} = \{ \langle\boldsymbol{x}_i, \boldsymbol{m}_i \rangle : y_i = 0 \}
        \]
    \State Compute the \textbf{initial} coefficient vector and the intercept 
    \Comment{See \citet{bunescu2007multiple}}
    \[
    (\hat{\boldsymbol{\theta}}, \hat{\beta}) \leftarrow \texttt{solve\_sMIL}(\mathcal{X}^+, \mathcal{X}^-), \text{ where } \hat{\boldsymbol{\theta}} \in \mathbb{R}^{1 \times d} \text{ and } \hat{\beta} \in \mathbb{R}.
    \]
    \State Let \(\tilde{\mathcal{X}}^+\) denote the set of all instances from the positive bags and \(\tilde{\mathcal{X}}^-\) denote all instances from the negative bags.
    \State Compute scores for every instance in a positive set
   \[
   S^+ \leftarrow \tilde{\mathcal{X}}^+\hat{\boldsymbol{\theta}}^{\top} + \hat{\beta}\text{, where } \tilde{\mathcal{X}}^+ \in \mathbb{R}^{|\tilde{\mathcal{X}}^+| \times d}.
   \]
    \State Compute the threshold 
    \[
    q \leftarrow \texttt{quantile}(S^+, 1- \eta).
    \]
    
    \ForAll{positive instances \(\langle\bar{\boldsymbol{x}}_j, \bar{m}_j,\bar{y}_j\rangle \in \tilde{\mathcal{X}}^+\), where \(\bar{\boldsymbol{x}}_j \in \mathbb{R}^{1 \times d}, \, \bar{m}_j \in \{0,1\}\) and \(\bar{y}_j = \emptyset\)}
    \Statex \quad \textbf{if} \( (\bar{\boldsymbol{x}}_j \,\boldsymbol{\hat\theta}^{\top} + \hat{\beta})\geq q\) and \(\bar{m}_j = 1\) \textbf{ then} set \(\bar{y}_j = 1\);
    \Statex \quad \textbf{else} set \(\bar{y}_j=0\).
    \EndFor
    \State Compute the \textbf{final} coefficient vector and the intercept
    \Comment{via a SIL SVM}
    \[ (\boldsymbol{\theta}, \beta) \leftarrow \texttt{solve\_SIL}(\mathcal{\tilde{X}}^+, \mathcal{\tilde{X}}^-).
    \] 
    \Statex \Return  Coefficient vector \( \boldsymbol{\theta} \in \mathbb{R}^{1 \times d}\) and the intercept  \(\beta \in \mathbb{R} \).
\end{algorithmic}
\end{algorithm}

\begin{algorithm}[ht]
\caption{Training a mean-difference (\texttt{MD}) probe~\cite{marks2023geometry}, sometimes referred to as mean-mass/mean-cluster difference classifier or linear discriminant analysis.}
	\label{alg:mean_diff}
	\renewcommand{\algorithmicrequire}{\textbf{Input:}}
	\renewcommand{\algorithmicensure}{\textbf{Output:}}
\begin{algorithmic}[1]
    \Statex \Require A training dataset \(\{\langle \boldsymbol{z}_i, y_i\rangle\}_{i=1}^n\) with binary labels \(y_i \in \{0,1\}\) and \( \boldsymbol{z}_i \in \mathbb{R}^{1 \times d}\). In our experiments, \(\boldsymbol z\) is the embedding of the last token (unless otherwise noted).
    \Statex \Ensure Parameters \(\boldsymbol\theta \in \mathbb{R}^{1 \times d}\),  \(\beta \in \mathbb{R}\), and \(\boldsymbol \Sigma^{-1} \in \mathbb{R}^{d \times d}\). 
    These parameters are subsequently given to a function \(f\) along with \(\boldsymbol z\) to compute \( f(\boldsymbol{z}) = \sigma\big(\boldsymbol{z} \, (\boldsymbol{\Sigma}^{-1} \, \boldsymbol{\theta}^{\top})+ \beta\big) \), where  \(\sigma\) is a sigmoid function.
    \vspace{0.8em} 
    
    \State Partition data into positive and negative sets:
        \[
        \mathcal{X}^{+} = \{ \boldsymbol{z}_i : y_i = 1 \}, \quad \mathcal{X}^{-} = \{ \boldsymbol{z}_i : y_i = 0 \}
        \]
    \State Compute class means \( \boldsymbol{\mu}^+ \) and \( \boldsymbol{\mu}^- \), and covariance matrices \(\boldsymbol{\Sigma}^+\) and \(\boldsymbol{\Sigma}^-\) for \(\mathcal{X}^+\) and \(\mathcal{X}^-\).
    \State Compute pooled covariance matrix (where \(n^+ = |\mathcal{X^+}|\) and \(n^- = |\mathcal{X^-}|\)):
        \[
        \boldsymbol{\Sigma} = \frac{(n^+ - 1)\,\boldsymbol{\Sigma}^+ + (n^- - 1)\,\boldsymbol{\Sigma}^-}{n^+ + n^- - 2}
        \]
    \State Compute the coefficient vector:
        \(
        \boldsymbol{\theta} = \boldsymbol{\mu}^+ - \boldsymbol{\mu}^-
        \), where \(\boldsymbol{\theta} \in \mathbb{R}^{1 \times d}\).
    \State Compute scores for positive and negative sets:
\[
s^+ \leftarrow \mathcal{X}^{+}\, (\boldsymbol{\Sigma}^{-1}\boldsymbol{\theta}^{\top}) \; \text{ and } \; s^- \leftarrow \mathcal{X}^{-}\, (\boldsymbol{\Sigma}^{-1}\boldsymbol{\theta}^{\top}) \text{, where } s^+ \in \mathbb{R}^{n^+} \text{ and } s^- \in \mathbb{R}^{n^-}.
\]
    \State Compute the intercept:
        \[
        \beta = -\frac{1}{2}\left(\operatorname{mean}\left(s^+\right) + \operatorname{mean}\left(s^-\right)\right)
        \]
    \Statex \Return Coefficient vector \(\boldsymbol \theta \), intercept \(\beta\), and  the inverse covariance matrix \(\boldsymbol \Sigma^{-1}\).
\end{algorithmic}
\end{algorithm}

\begin{algorithm}[ht]
\caption{Training of Truth and Polarity Directions, aka \texttt{TTPD} probe~\cite{burger2024truth}}
	\label{alg:ttpd}
	\renewcommand{\algorithmicrequire}{\textbf{Input:}}
	\renewcommand{\algorithmicensure}{\textbf{Output:}}
\begin{algorithmic}[1]
    \Statex \Require A training dataset \( \{\langle \boldsymbol{z}_i, y_i, p_i\rangle\}_{i=1}^n\) with binary truthfulness labels \(y_i \in \{-1,1\}\), binary polarity labels \(p_i \in \{-1,1\}\) that specify affirmative/negated statements and \( \boldsymbol{z}_i \in \mathbb{R}^{1 \times d}\). In our experiments, \(\boldsymbol z\) is the embedding of the last token (unless otherwise noted).
    \Statex \Ensure Parameters \(\Theta  \in \mathbb{R}^{2 \times d}\),  \(\boldsymbol\theta \in \mathbb{R}^{1 \times 2}\) and \(\beta \in \mathbb{R}\). 
    These parameters are subsequently given to a function \(f\) along with \(\boldsymbol z\) to compute \( f(\boldsymbol{z}) = \sigma\big((\boldsymbol{z} \, \Theta^{\top}) \, \boldsymbol{\theta}^{\top}+ \beta \big) \), where  \(\sigma\) is a sigmoid function.
    \vspace{0.8em} 
    
    \State Given data matrix \(X = [\boldsymbol{z}_1, \dots, \boldsymbol{z}_n] \in \mathbb{R}^{n\times d}\) compute the centered data matrix 
    \[
    \bar{X} = X - \operatorname{mean}(X)
    \]
    \State Find the truth direction \(\boldsymbol\theta_t \in \mathbb{R}^{1 \times d}\) via the Ordinary Least Squares:
    \[
    \boldsymbol{\theta}_t =
    (\bar{X}^{\top} \,\bar{X})^{-1} \bar{X}^{\top} \boldsymbol{y} \text{, where } \boldsymbol{y} = [y_1, \dots, y_n]
    \]
    \State Find the polarity direction \(\boldsymbol{\theta}_p \in \mathbb{R}^{1 \times d}\) via the Logistic Regression:
    \[
    \boldsymbol{\theta}_p \leftarrow \operatorname{LogisticRegression}(X, \boldsymbol p) \text{, where } \boldsymbol{p} = [p_1, \dots, p_n]
    \]
    \State Project \(X\) onto \(\boldsymbol{\theta}_t\) and \(\boldsymbol{\theta}_p\):
    \[ \hat{X} \leftarrow \operatorname{stack}(\big[ X\, \boldsymbol \theta_t^{\top}, X\, \boldsymbol \theta_p^{\top}\big]) \text{, where } \hat{X} \in \mathbb{R}^{n \times 2} \]
    \State Use the \textit{projected} data matrix \(\hat{X}\) to get coefficient vector \(\boldsymbol{\theta}\) and the intercept \(\beta\):
    \[
    \boldsymbol{\theta}\, ,\, \beta \leftarrow \operatorname{LogisticRegression}(\hat X, \boldsymbol{y})
    \]
    \Statex \Return Projection matrix \(\Theta = \operatorname{stack}( [ \boldsymbol{\theta}_t, \boldsymbol{\theta}_p])\),  coefficient vector \(\boldsymbol \theta \) and intercept \(\beta\).
\end{algorithmic}
\end{algorithm}

\begin{algorithm}[ht]
\caption{Training a Supervised Principal Component Analysis (\texttt{sPCA}) probe}
	\label{alg:spca}
	\renewcommand{\algorithmicrequire}{\textbf{Input:}}
	\renewcommand{\algorithmicensure}{\textbf{Output:}}
\begin{algorithmic}[1]
    \Statex \Require A training dataset \(\{\langle \boldsymbol{z}_i, y_i\rangle\}_{i=1}^n\) with binary labels \(y_i \in \{0,1\}\) and \( \boldsymbol{z}_i \in \mathbb{R}^{1 \times d}\), \(k \in \mathbb{N}\), \(k \le d\) to specify the number of components and ridge penalty \(\lambda > 0\).
    In our experiments, \(\boldsymbol z\) is the embedding of the last token (unless otherwise noted).
    \Statex \Ensure Parameters \(\tilde\Lambda \in \mathbb{R}^{k \times d}\), \(\boldsymbol\theta \in \mathbb{R}^{1 \times k}\) and \(\beta \in \mathbb{R}\).
    These parameters are subsequently given to a function \(f\) along with \(\boldsymbol z\) to compute \( f(\boldsymbol{z}) = \sigma\big((\boldsymbol{z} \, \tilde\Lambda^{\top}) \, \boldsymbol{\theta}^{\top} + \beta \big) \), where \(\sigma\) is a sigmoid function.
    \vspace{0.8em} 
    
    \State Partition data into positive and negative sets:
        \[
        \mathcal{X}^{+} = \{ \boldsymbol{z}_i : y_i = 1 \}, \quad \mathcal{X}^{-} = \{ \boldsymbol{z}_i : y_i = 0 \}
        \]
    \State
    Compute means and centered matrices (\(\mathcal X = \mathcal X^+ \cup \mathcal X^-\)):
\[
\boldsymbol\mu^+=\operatorname{mean}( \mathcal X^+),\quad \boldsymbol\mu^-=\operatorname{mean}(\mathcal X^-),\quad
\boldsymbol \mu=\operatorname{mean}( \mathcal X),\quad
\mathcal X^+_c= \mathcal X^+- \boldsymbol \mu^+,\ \mathcal X^-_c= \mathcal X^- - \boldsymbol\mu^-.
\]
    \State Compute class covariance matrices \(\boldsymbol{\Sigma}^+\) and \(\boldsymbol{\Sigma}^-\) for \(\mathcal{X}^+_c\) and \(\mathcal{X}_c^-\).
    \State Compute within-class covariance matrix (where \(n^+ = |\mathcal{X}^+_c|\) and \(n^- = |\mathcal{X}^-_c|\)):
        \[
        \boldsymbol{\Sigma}_w = \frac{(n^+ - 1)\,\boldsymbol{\Sigma}^+ + (n^- - 1)\,\boldsymbol{\Sigma}^-}{n^+ + n^- - 2}
        \]
    \State Compute between-class covariance matrix:
\[
\boldsymbol d_p=(\boldsymbol\mu^+-\boldsymbol\mu),\quad \boldsymbol d_n=(\boldsymbol\mu^- - \boldsymbol\mu),\qquad
\boldsymbol \Sigma_b=\frac{n^+\, \boldsymbol d_p^\top \boldsymbol d_p + n^-\, \boldsymbol d_n^\top \boldsymbol d_n}{n}.
\]
\State Build symmetric scatter matrix with ridge penalty:
\[
\boldsymbol M = \boldsymbol \Sigma_b + \boldsymbol \Sigma_w + \lambda \boldsymbol I_d,\qquad \boldsymbol M \leftarrow \tfrac{1}{2}(\boldsymbol M+ \boldsymbol M^\top).
\]

\State Compute top-\(k\) eigenpairs of \(M\) (largest algebraic),  assemble the eigenmatrix \(\Lambda \in \mathbb{R}^{k \times d}\) and corresponding eigenvalue vector \(\boldsymbol{\nu} \in \mathbb{R}^{k}\):
    \[
        (\nu_1,\boldsymbol u_1),\ldots,(\nu_k,\boldsymbol u_k)=\texttt{TopKEigs}(\boldsymbol M,k)
        \;\rightarrow\;
        \Lambda = [\,\boldsymbol u_1, \ldots, \boldsymbol u_k\,]^{\top},
        \quad
        \boldsymbol{\nu} = [\,\nu_1, \ldots, \nu_k\,]^{\top}.
    \] 
    Since \(\boldsymbol M\) is PD (ridge \(\lambda > 0\)), every \(\nu_i \ge \lambda > 0\), so the whitening in the next step is well-defined.
\State Project data matrix \( X = [\boldsymbol{z}_1; \dots; \boldsymbol{z}_n] \in \mathbb{R}^{n \times d}
\) via the whitened eigenmatrix \(\Lambda\):
\[
\hat{X} = X \, \tilde\Lambda^{\top} \text{, where } \tilde\Lambda = \operatorname{diag}(\boldsymbol{\nu})^{-\frac{1}{2}}\,\Lambda \in \mathbb{R}^{k \times d}, \quad \hat{X} \in \mathbb{R}^{n \times k}
\]
    \State Use the \textit{projected} data matrix \(\hat{X}\) to get coefficient vector \(\boldsymbol{\theta}\) and the intercept \(\beta\):
    \[
    \boldsymbol{\theta}\, ,\, \beta \leftarrow \operatorname{LogisticRegression}(\hat{X}, \boldsymbol{y})
    \]
    \Statex \Return Projection matrix \(\tilde\Lambda \),  coefficient vector \(\boldsymbol \theta \) and intercept \(\beta\).
\end{algorithmic}
\end{algorithm}

\begin{algorithm}[ht]
	\caption{Inductive Conformal Predictions with binary nonconformity score.}
	\label{alg:sup_cp_binary}
	\renewcommand{\algorithmicrequire}{\textbf{Input:}}
	\renewcommand{\algorithmicensure}{\textbf{Output:}}
\begin{algorithmic}[1]
    \Statex \Require A calibration dataset 
    \(\{( \boldsymbol{z}_i, y_i)\} _{i=1}^n \subseteq \mathcal{D}_{cal}\), where \(n = |\mathcal{D}_{cal}|\), \(y_i \in \{-1,1\}\) and \(\boldsymbol{z}_i \in \mathbb{R}^{L \times d}\) (\(L=1\) in a single instance setup); a significance level $\alpha \in (0,1)$ (target coverage \(1 -\alpha\)), a pretrained binary classifier \(g\), and a new sample \(\boldsymbol{z}_{new}\). 
    
    \Statex \Ensure  Prediction set \(\mathcal{Y}_{new}\)
    \vspace{0.8em}
    \State Initialize an empty score list $S \leftarrow \emptyset $
    \vspace{0.5em}
    \ForAll{samples \(\langle \boldsymbol z_i, y_i\rangle \in \mathcal{D}_{cal}\)}
        \State \(\tilde z_i = g\left(\boldsymbol{z}_i\right)\) \Comment{\(\tilde z_i \in \mathbb{R}\) is a score}
        \State \(s_i = \exp(-y_i \cdot \tilde z_i)\) \Comment{Nonconformity score, see Eq.~\ref{eq:nc_symmetric}.}
        \State $S \gets S \cup \{\,s_i\}$
    \EndFor
    \vspace{0.5em}
    \State Compute a score for the new samples: \(
    \tilde z_{new} = g\left(\boldsymbol{z}_{new}\right)
    \)
    \State Initialize an empty prediction set \(\mathcal{Y}_{new} \leftarrow \emptyset\)
    \vspace{0.5em}
    \ForAll{$y \in \{-1,1\}$}
    \State $s_{new} = \exp(-y \cdot \tilde z_{new})$
    \State $ \psi_y = \frac{1 + |\{\, s_i \in S : s_i \geq s_{new} \,\}|}{1 + |S|}$     \Comment{$\mathbb{I}(\cdot)$ is an indicator function}
    \vspace{0.5em}
        \If{$\psi_y > \alpha$}
            \State \(\mathcal{Y}_{new} \gets \mathcal{Y}_{new} \cup \{y\}\)
            \vspace{0.3em}
        \EndIf
        \vspace{0.3em}
    \EndFor
    \vspace{0.5em}
    \State \Return Prediction set \(\mathcal{Y}_{new}\)
\end{algorithmic}
\end{algorithm}

\begin{algorithm}[ht]
	\caption{Inductive Conformal Predictions with multiclass nonconformity score.}
	\label{alg:sup_cp_multiclass}
	\renewcommand{\algorithmicrequire}{\textbf{Input:}}
	\renewcommand{\algorithmicensure}{\textbf{Output:}}

\begin{algorithmic}[1]
    \Statex \Require A calibration dataset 
    \(\{( \boldsymbol{z}_i, y_i)\} _{i=1}^n \subseteq \mathcal{D}_{cal}\), where \(n = |\mathcal{D}_{cal}|\), \(y_i \in \{1,\dots, K\}\) (\(K\) is the number of classes) and \(\boldsymbol{z}_i \in \mathbb{R}^{L \times d}\) (\(L=1\) in a single instance setup); a significance level $\alpha \in (0,1)$ (target coverage \(1 -\alpha\)), a pretrained multiclass classifier \(g\), and a new sample \(\boldsymbol{z}_{new}\). 
    
    \Statex \Ensure  Prediction set \(\mathcal{Y}_{new}\)
    \vspace{0.8em}
    \State Initialize an empty score list $S \leftarrow \emptyset $
    \vspace{0.5em}
    \ForAll{samples \(\langle \boldsymbol z_i, y_i\rangle \in \mathcal{D}_{cal}\)}
        \State \(\mathbf p = g\left(\boldsymbol{z}_i\right)\) \Comment{\(\mathbf p \in \Delta^{K-1}\) is a vector of probabilities}
\State \(p_z = \max_{j \not=y_i} p_j\)  \Comment{ Maximum non-target probability, where \(j \in \{1,\dots,K\}\)}
\State \(d_p = p_{y_i} - p_z\) \Comment{Probability margin}
        \State \(s_i = \frac{1 - d_p}{2} \)\Comment{Nonconformity score, see Eq.~\ref{eq:nc_margin}.}
        \vspace{0.2em}
        \State $S \gets S \cup \{\,s_i\}$
    \EndFor
    \vspace{0.5em}
    \State Compute probabilities for the new samples: \(
    \boldsymbol p_{new} = g\left(\boldsymbol{z}_{new}\right)
    \)
    \State Initialize empty prediction set \(\mathcal{Y}_{new} \leftarrow \emptyset\)
    \vspace{0.5em}
    \ForAll{$y \in \{1, \dots, K\}$}
    \State \(p_z = \max_{j \not=y} p_{new,j}\)  \Comment{ where \(j \in \{1,\dots,K\}\) and \(p_{new,j} \in \boldsymbol{p}_{new}\)}
    \State \(d_p = p_{new,y} - p_z\)
    \State \(s_{new} = \frac{1 - d_p}{2} \)
    \State $\psi_y = \frac{1 + |\{\, s_i \in S : s_i \geq s_{new} \,\}|}{1 + |S|}$     \Comment{$\mathbb{I}(\cdot)$ is an indicator function}
    \vspace{0.5em}
        \If{$\psi_y > \alpha$}
            \State \(\mathcal{Y}_{new} \gets \mathcal{Y}_{new} \cup \{y\}\)
        \vspace{0.3em}
        \EndIf
        \vspace{0.3em}
    \EndFor
    \vspace{0.5em}
    \State \Return Prediction set \(\mathcal{Y}_{new}\)
\end{algorithmic}
\end{algorithm}

\clearpage
\twocolumn
\section{More Tables on Classification Performance, Generalization Performance, and Confusion Matrices}
\label{supsec:tables}

In this section, we report detailed tables of the following results.

\begin{itemize}
    \item Classification performance for all $\langle$model, dataset$\rangle$ pairs for the setting when only the last token's representation is available, and the setting when the prediction is made over the full bag (see~\Cref{sup_tab:mcc_long_sawmil_instance,sup_tab:mcc_long_sawmil_bag,sup_tab:mcc_long_zero_shot,sup_tab:mcc_long_mdcp_instance,sup_tab:mcc_long_mdcp_bag,sup_tab:mcc_long_ttpd_instance,sup_tab:mcc_long_ttpd_bag,sup_tab:mcc_long_spca_instance,sup_tab:mcc_long_spca_bag,sup_tab:mcc_long_svm_instance,sup_tab:mcc_long_svm_bag});
    \item Generalization performance aggregated per setting, probe, and training dataset (see~\Cref{sup_tab:aggregated_generalization}) and a detailed overview for the multiclass \texttt{sAwMIL} across all datasets (see~\Cref{tab:sup_generalization_cities,tab:sup_generalization_indications,tab:sup_generalization_defs});
    \item Confusion matrices for each \(\langle\)model, dataset\(\rangle\) pair (see~\Cref{sup_tab:cm_zero_shot,sup_tab:cm_mdcp_instance,sup_tab:cm_mdcp_bag,sup_tab:cm_ttpd_instance,sup_tab:cm_ttpd_bag,sup_tab:cm_spca_instance,sup_tab:cm_spca_bag,sup_tab:cm_sawmil_bag}).
\end{itemize}

\newpage
\onecolumn
\begin{small}
\begin{center}
\begin{longtable}{@{}lllclcccp{0.7cm}c@{}}
  \toprule
   Model Name & Type & Setting & Probe & Dataset & CI$_{.025}$ & MCC & CI$_{.975}$ & Layer & Depth\\
  \midrule
  \endfirsthead

  \multicolumn{10}{c}%
    {{\bfseries \tablename\ \thetable{} — continued from previous page}} \\
  \toprule
    Model Name & Type & Setting & Probe & Dataset & CI$_{.025}$ & MCC & CI$_{.975}$ & L & Depth\\
  \midrule
  \endhead

  \midrule
  \multicolumn{10}{r}{{Continued on next page}} \\ 
  \endfoot

  \bottomrule
   \\[-2pt]
    \caption{\textbf{Classification performance of the multiclass \texttt{sAwMIL} probe across datasets and models} (evaluated on the \underline{\textit{last token's representation}}).
    This probe is trained on bag representations of the statements and evaluated using representations of the last tokens.
    We report Matthews Correlation Coefficient (MCC) with the \(95\% \) confidence intervals.
    Confidence intervals are based on bootstrapping with \(n=1,000\) samples.
    The \textbf{bold} values mark MCC with the confidence intervals that exclude zero.
    The `Setting' column specifies the evaluation setting.
    The `Layer' column specifies the layer at which the probe achieved the highest MCC, and `Depth' denotes the relative depth of that layer within the model.
  \label{sup_tab:mcc_long_sawmil_instance}}
  \endlastfoot

Llama-3-8B & default & instance & \texttt{sAwMIL} & City Locations & 0.99 & \textbf{0.99} & 1.00 & 16 & 0.52 \\
Llama-3.2-3B & default & instance & \texttt{sAwMIL} & City Locations & 0.98 & \textbf{0.99} & 1.00 & 10 & 0.37 \\
Mistral-7B-v0.3 & default & instance & \texttt{sAwMIL} & City Locations & 0.99 & \textbf{0.99} & 1.00 & 17 & 0.55 \\
Qwen-2.5-7B & default & instance & \texttt{sAwMIL} & City Locations & 0.98 & \textbf{0.99} & 0.99 & 19 & 0.70 \\
Qwen-2.5-14B & default & instance & \texttt{sAwMIL} & City Locations & 0.98 & \textbf{0.99} & 0.99 & 28 & 0.60 \\
Gemma-7B & default & instance & \texttt{sAwMIL} & City Locations & 0.99 & \textbf{1.00} & 1.00 & 20 & 0.74 \\
Gemma-2-9B & default & instance & \texttt{sAwMIL} & City Locations & 0.99 & \textbf{1.00} & 1.00 & 21 & 0.51 \\
Llama-3.1-8B & chat & instance & \texttt{sAwMIL} & City Locations & 0.99 & \textbf{1.00} & 1.00 & 13 & 0.42 \\
\bottomrule
\end{longtable}
\end{center}
\end{small}

\newpage
\begin{small}
\begin{center}
\begin{longtable}{@{}lllclcccp{0.7cm}c@{}}
  \toprule
   Model Name & Type & Setting & Probe & Dataset & CI$_{.025}$ & MCC & CI$_{.975}$ & Layer & Depth\\
  \midrule
  \endfirsthead

  \multicolumn{10}{c}%
    {{\bfseries \tablename\ \thetable{} — continued from previous page}} \\
  \toprule
    Model Name & Type & Setting & Probe & Dataset & CI$_{.025}$ & MCC & CI$_{.975}$ & L & Depth\\
  \midrule
  \endhead

  \midrule
  \multicolumn{10}{r}{{Continued on next page}} \\ 
  \endfoot

  \bottomrule
  \\[-2pt]
    \caption{ \textbf{Classification performance of the multiclass \texttt{sAwMIL} probe across datasets and models} (evaluated on the \underline{\textit{full bag}}).
    This probe is trained and evaluated on the bag representation of the statements.
    We report Matthews Correlation Coefficient (MCC) with the \(95\% \) confidence intervals.
    Confidence intervals are based on bootstrapping with \(n=1{,}000\) samples.
    The \textbf{bold} values mark MCC with the confidence intervals that exclude zero.
    The `Setting' column specifies the evaluation setting.
    The `Layer' column specifies the layer at which the probe achieved the highest MCC, and `Depth' denotes the relative depth of that layer within the model.
  \label{sup_tab:mcc_long_sawmil_bag}}\\
  \endlastfoot

Llama-3-8B & default & bag & \texttt{sAwMIL} & City Locations & 0.99 & \textbf{0.99} & 1.00 & 16 & 0.52 \\
Llama-3.2-3B & default & bag & \texttt{sAwMIL} & City Locations & 0.98 & \textbf{0.99} & 1.00 & 10 & 0.37 \\
Mistral-7B-v0.3 & default & bag & \texttt{sAwMIL} & City Locations & 0.99 & \textbf{0.99} & 1.00 & 17 & 0.55 \\
Qwen-2.5-7B & default & bag & \texttt{sAwMIL} & City Locations & 0.98 & \textbf{0.99} & 0.99 & 19 & 0.70 \\
Qwen-2.5-14B & default & bag & \texttt{sAwMIL} & City Locations & 0.98 & \textbf{0.99} & 0.99 & 28 & 0.60 \\
Gemma-7B & default & bag & \texttt{sAwMIL} & City Locations & 0.99 & \textbf{1.00} & 1.00 & 20 & 0.74 \\
Gemma-2-9B & default & bag & \texttt{sAwMIL} & City Locations & 0.99 & \textbf{1.00} & 1.00 & 21 & 0.51 \\
Llama-3.1-8B & chat & bag & \texttt{sAwMIL} & City Locations & 0.99 & \textbf{1.00} & 1.00 & 13 & 0.42 \\
\bottomrule
\end{longtable}
\end{center}
\end{small}

\newpage
\begin{small}
\begin{center}
\begin{longtable}{@{}llcclcccp{0.7cm}c@{}}
  \toprule
   Model Name & Type & Setting & Probe & Dataset & CI$_{.025}$ & MCC & CI$_{.975}$ & Layer & Depth\\
  \midrule
  \endfirsthead

  \multicolumn{10}{c}%
    {{\bfseries \tablename\ \thetable{} — continued from previous page}} \\
  \toprule
    Model Name & Type & Setting & Probe & Dataset & CI$_{.025}$ & MCC & CI$_{.975}$ & L & Depth\\
  \midrule
  \endhead

  \midrule
  \multicolumn{10}{r}{{Continued on next page}} \\ 
  \endfoot

  \bottomrule
  \\[-2pt]
    \caption{ \textbf{Classification performance of zero-shot prompting across datasets and models}.
    We report Matthews Correlation Coefficient (MCC) with the \(95\% \) confidence intervals.
    Confidence intervals are based on bootstrapping with \(n=1{,}000\) samples.
    The \textbf{bold} values mark MCC with the confidence intervals that exclude zero.
    The `Setting' column specifies the evaluation setting.
    The `Layer' column specifies the layer at which the probe achieved the highest MCC, and `Depth' denotes the relative depth of that layer within the model (do not apply to zero-shot prompting).
  \label{sup_tab:mcc_long_zero_shot}}\\
  \endlastfoot

Llama-3-8B & default & - & \texttt{Zero-Shot} & City Locations & 0.05 & \textbf{0.09} & 0.12 & - & - \\
Llama-3.2-3B & default & - & \texttt{Zero-Shot} & City Locations & 0.09 & \textbf{0.12} & 0.15 & - & - \\
Mistral-7B-v0.3 & default & - & \texttt{Zero-Shot} & City Locations & 0.00 & 0.00 & 0.00 & - & - \\
Qwen-2.5-7B & default & - & \texttt{Zero-Shot} & City Locations & 0.53 & \textbf{0.56} & 0.58 & - & - \\
Qwen-2.5-14B & default & - & \texttt{Zero-Shot} & City Locations & 0.61 & \textbf{0.63} & 0.65 & - & - \\
Gemma-7B & default & - & \texttt{Zero-Shot} & City Locations & 0.12 & \textbf{0.15} & 0.18 & - & - \\
Gemma-2-9B & default & - & \texttt{Zero-Shot} & City Locations & 0.50 & \textbf{0.53} & 0.56 & - & - \\
Llama-3.1-8B & chat & - & \texttt{Zero-Shot} & City Locations & 0.54 & \textbf{0.56} & 0.58 & - & - \\
\bottomrule
\end{longtable}
\end{center}
\end{small}

\newpage
\begin{small}
\begin{center}
\begin{longtable}{@{}lllclcccp{0.7cm}c@{}}
  \toprule
   Model Name & Type & Setting & Probe & Dataset & CI$_{.025}$ & MCC & CI$_{.975}$ & Layer & Depth\\
  \midrule
  \endfirsthead

  \multicolumn{10}{c}%
    {{\bfseries \tablename\ \thetable{} — continued from previous page}} \\
  \toprule
    Model Name & Type & Setting & Probe & Dataset & CI$_{.025}$ & MCC & CI$_{.975}$ & L & Depth\\
  \midrule
  \endhead

  \midrule
  \multicolumn{10}{r}{{Continued on next page}} \\ 
  \endfoot

  \bottomrule
  \\[-2pt]
    \caption{ \textbf{Classification performance of the mean-difference probe with conformal prediction (\texttt{MD+CP}) probe across datasets and models} (evaluated on the \underline{\textit{last token's representation}}). This probe is trained and evaluated using the representations of the last tokens.
    We report Matthews Correlation Coefficient (MCC) with the \(95\% \) confidence intervals.
    Confidence intervals are based on bootstrapping with \(n=1{,}000\) samples.
    The \textbf{bold} values mark MCC with the confidence intervals that exclude zero.
    The `Setting' column specifies the evaluation setting.
    The `Layer' column specifies the layer at which the probe achieved the highest MCC, and `Depth' denotes the relative depth of that layer within the model.
    \label{sup_tab:mcc_long_mdcp_instance}}\\
  \endlastfoot
  
Llama-3-8B & default & instance & \texttt{MD+CP} & City Locations & 0.64 & \textbf{0.66} & 0.69 & 19 & 0.61 \\
Llama-3.2-3B & default & instance & \texttt{MD+CP} & City Locations & 0.60 & \textbf{0.62} & 0.64 & 27 & 1.00 \\
Mistral-7B-v0.3 & default & instance & \texttt{MD+CP} & City Locations & 0.65 & \textbf{0.68} & 0.70 & 24 & 0.77 \\
Qwen-2.5-7B & default & instance & \texttt{MD+CP} & City Locations & 0.57 & \textbf{0.60} & 0.62 & 22 & 0.81 \\
Qwen-2.5-14B & default & instance & \texttt{MD+CP} & City Locations & 0.62 & \textbf{0.65} & 0.67 & 32 & 0.68 \\
Gemma-7B & default & instance & \texttt{MD+CP} & City Locations & 0.66 & \textbf{0.68} & 0.70 & 21 & 0.78 \\
Gemma-2-9B & default & instance & \texttt{MD+CP} & City Locations & 0.70 & \textbf{0.73} & 0.75 & 24 & 0.59 \\
Llama-3.1-8B & chat & instance & \texttt{MD+CP} & City Locations & 0.69 & \textbf{0.71} & 0.74 & 27 & 0.87 \\
\bottomrule
\end{longtable}
\end{center}
\end{small}

\newpage
\begin{small}
\begin{center}
\begin{longtable}{@{}lllclcccp{0.7cm}c@{}}
  \toprule
   Model Name & Type & Setting & Probe & Dataset & CI$_{.025}$ & MCC & CI$_{.975}$ & Layer & Depth\\
  \midrule
  \endfirsthead

  \multicolumn{10}{c}%
    {{\bfseries \tablename\ \thetable{} — continued from previous page}} \\
  \toprule
    Model Name & Type & Setting & Probe & Dataset & CI$_{.025}$ & MCC & CI$_{.975}$ & L & Depth\\
  \midrule
  \endhead

  \midrule
  \multicolumn{10}{r}{{Continued on next page}} \\ 
  \endfoot

  \bottomrule
  \\[-2pt]
    \caption{ \textbf{Classification performance of the mean-difference probe with conformal prediction (\texttt{MD+CP}) probe across datasets and models} (evaluated on the \underline{\textit{full bag}}). This probe is trained on the representation of the last tokens and evaluated on the bag representation of the statements.
    We report Matthews Correlation Coefficient (MCC) with the \(95\% \) confidence intervals.
    Confidence intervals are based on bootstrapping with \(n=1{,}000\) samples.
    The \textbf{bold} values mark MCC with the confidence intervals that exclude zero.
    The `Setting' column specifies the evaluation setting.
    The `Layer' column specifies the layer at which the probe achieved the highest MCC, and `Depth' denotes the relative depth of that layer within the model.
  \label{sup_tab:mcc_long_mdcp_bag}}\\
  \endlastfoot
  
Llama-3-8B & default & bag & \texttt{MD+CP} & City Locations & 0.00 & 0.00 & 0.00 & 2 & 0.06 \\
Llama-3.2-3B & default & bag & \texttt{MD+CP} & City Locations & 0.00 & 0.00 & 0.00 & 2 & 0.07 \\
Mistral-7B-v0.3 & default & bag & \texttt{MD+CP} & City Locations & 0.08 & \textbf{0.11} & 0.14 & 9 & 0.29 \\
Qwen-2.5-7B & default & bag & \texttt{MD+CP} & City Locations & 0.06 & \textbf{0.08} & 0.11 & 12 & 0.44 \\
Qwen-2.5-14B & default & bag & \texttt{MD+CP} & City Locations & 0.07 & \textbf{0.10} & 0.11 & 37 & 0.79 \\
Gemma-7B & default & bag & \texttt{MD+CP} & City Locations & 0.29 & \textbf{0.31} & 0.34 & 7 & 0.26 \\
Gemma-2-9B & default & bag & \texttt{MD+CP} & City Locations & 0.01 & \textbf{0.04} & 0.06 & 33 & 0.80 \\
Llama-3.1-8B & chat & bag & \texttt{MD+CP} & City Locations & 0.00 & 0.00 & 0.00 & 2 & 0.06 \\
\bottomrule
\end{longtable}
\end{center}
\end{small}

\newpage
\begin{small}
\begin{center}
\begin{longtable}{@{}lllclcccp{0.7cm}c@{}}
  \toprule
   Model Name & Type & Setting & Probe & Dataset & CI$_{.025}$ & MCC & CI$_{.975}$ & Layer & Depth\\
  \midrule
  \endfirsthead

  \multicolumn{10}{c}%
    {{\bfseries \tablename\ \thetable{} — continued from previous page}} \\
  \toprule
    Model Name & Type & Setting & Probe & Dataset & CI$_{.025}$ & MCC & CI$_{.975}$ & L & Depth\\
  \midrule
  \endhead

  \midrule
  \multicolumn{10}{r}{{Continued on next page}} \\ 
  \endfoot

  \bottomrule
  \\[-2pt]
    \caption{ \textbf{Classification performance of the TTPD with conformal prediction (\texttt{TTPD+CP}) probe across datasets and models} (evaluated on the \underline{\textit{last token's representation}}). This probe is trained and evaluated using the representations of the last tokens.
    We report Matthews Correlation Coefficient (MCC) with the \(95\% \) confidence intervals.
    Confidence intervals are based on bootstrapping with \(n=1{,}000\) samples.
    The \textbf{bold} values mark MCC with the confidence intervals that exclude zero.
    The `Setting' column specifies the evaluation setting.
    The `Layer' column specifies the layer at which the probe achieved the highest MCC, and `Depth' denotes the relative depth of that layer within the model.
  \label{sup_tab:mcc_long_ttpd_instance}}\\
  \endlastfoot

Llama-3-8B & default & instance & \texttt{TTPD+CP} & City Locations & 0.53 & \textbf{0.55} & 0.57 & 13 & 0.42 \\
Llama-3.2-3B & default & instance & \texttt{TTPD+CP} & City Locations & 0.51 & \textbf{0.54} & 0.56 & 12 & 0.44 \\
Mistral-7B-v0.3 & default & instance & \texttt{TTPD+CP} & City Locations & 0.65 & \textbf{0.67} & 0.69 & 11 & 0.35 \\
Qwen-2.5-7B & default & instance & \texttt{TTPD+CP} & City Locations & 0.54 & \textbf{0.57} & 0.59 & 18 & 0.67 \\
Qwen-2.5-14B & default & instance & \texttt{TTPD+CP} & City Locations & 0.55 & \textbf{0.58} & 0.61 & 28 & 0.60 \\
Gemma-7B & default & instance & \texttt{TTPD+CP} & City Locations & 0.56 & \textbf{0.58} & 0.59 & 12 & 0.44 \\
Gemma-2-9B & default & instance & \texttt{TTPD+CP} & City Locations & 0.59 & \textbf{0.61} & 0.63 & 17 & 0.41 \\
Llama-3.1-8B & chat & instance & \texttt{TTPD+CP} & City Locations & 0.83 & \textbf{0.85} & 0.88 & 17 & 0.55 \\
\bottomrule
\end{longtable}
\end{center}
\end{small}

\newpage
\begin{small}
\begin{center}
\begin{longtable}{@{}lllclcccp{0.7cm}c@{}}
  \toprule
   Model Name & Type & Setting & Probe & Dataset & CI$_{.025}$ & MCC & CI$_{.975}$ & Layer & Depth\\
  \midrule
  \endfirsthead

  \multicolumn{10}{c}%
    {{\bfseries \tablename\ \thetable{} — continued from previous page}} \\
  \toprule
    Model Name & Type & Setting & Probe & Dataset & CI$_{.025}$ & MCC & CI$_{.975}$ & L & Depth\\
  \midrule
  \endhead

  \midrule
  \multicolumn{10}{r}{{Continued on next page}} \\ 
  \endfoot

  \bottomrule
  \\[-2pt]
    \caption{ \textbf{Classification performance of the TTPD with conformal prediction (\texttt{TTPD+CP}) probe across datasets and models} (evaluated on the \underline{\textit{full bag}}). This probe is trained on the representation of the last tokens and evaluated on the bag representation of the statements.
    We report Matthews Correlation Coefficient (MCC) with the \(95\% \) confidence intervals.
    Confidence intervals are based on bootstrapping with \(n=1{,}000\) samples. The \textbf{bold} values mark MCC with the confidence intervals that exclude zero.
    The `Setting' column specifies the evaluation setting.
    The `Layer' column specifies the layer at which the probe achieved the highest MCC, and `Depth' denotes the relative depth of that layer within the model.
  \label{sup_tab:mcc_long_ttpd_bag}}\\
  \endlastfoot
  
Llama-3-8B & default & bag & \texttt{TTPD+CP} & City Locations & 0.52 & \textbf{0.54} & 0.56 & 13 & 0.42 \\
Llama-3.2-3B & default & bag & \texttt{TTPD+CP} & City Locations & 0.47 & \textbf{0.54} & 0.62 & 24 & 0.89 \\
Mistral-7B-v0.3 & default & bag & \texttt{TTPD+CP} & City Locations & 0.49 & \textbf{0.51} & 0.52 & 15 & 0.48 \\
Qwen-2.5-7B & default & bag & \texttt{TTPD+CP} & City Locations & 0.49 & \textbf{0.53} & 0.56 & 17 & 0.63 \\
Qwen-2.5-14B & default & bag & \texttt{TTPD+CP} & City Locations & 0.53 & \textbf{0.55} & 0.58 & 20 & 0.43 \\
Gemma-7B & default & bag & \texttt{TTPD+CP} & City Locations & 0.56 & \textbf{0.58} & 0.60 & 16 & 0.59 \\
Gemma-2-9B & default & bag & \texttt{TTPD+CP} & City Locations & 0.50 & \textbf{0.53} & 0.55 & 21 & 0.51 \\
Llama-3.1-8B & chat & bag & \texttt{TTPD+CP} & City Locations & 0.00 & 0.00 & 0.00 & 2 & 0.06 \\
\bottomrule
\end{longtable}
\end{center}
\end{small}

\newpage
\begin{small}
\begin{center}
\begin{longtable}{@{}lllclcccp{0.7cm}c@{}}
  \toprule
   Model Name & Type & Setting & Probe & Dataset & CI$_{.025}$ & MCC & CI$_{.975}$ & Layer & Depth\\
  \midrule
  \endfirsthead

  \multicolumn{10}{c}%
    {{\bfseries \tablename\ \thetable{} — continued from previous page}} \\
  \toprule
    Model Name & Type & Setting & Probe & Dataset & CI$_{.025}$ & MCC & CI$_{.975}$ & L & Depth\\
  \midrule
  \endhead

  \midrule
  \multicolumn{10}{r}{{Continued on next page}} \\ 
  \endfoot

  \bottomrule
  \\[-2pt]
    \caption{ \textbf{Classification performance of the supervised PCA with conformal prediction (\texttt{sPCA+CP}) probe across datasets and models} (evaluated on the \underline{\textit{last token's representation}}). This probe is trained and evaluated using the representations of the last tokens.
    We report Matthews Correlation Coefficient (MCC) with the \(95\% \) confidence intervals.
    Confidence intervals are based on bootstrapping with \(n=1{,}000\) samples.
    The \textbf{bold} values mark MCC with the confidence intervals that exclude zero.
    The `Setting' column specifies the evaluation setting.
    The `Layer' column specifies the layer at which the probe achieved the highest MCC, and `Depth' denotes the relative depth of that layer within the model.
  \label{sup_tab:mcc_long_spca_instance}}\\
  \endlastfoot

Llama-3-8B & default & instance & \texttt{sPCA+CP} & City Locations & 0.72 & \textbf{0.74} & 0.77 & 26 & 0.84 \\
Llama-3.2-3B & default & instance & \texttt{sPCA+CP} & City Locations & 0.65 & \textbf{0.67} & 0.70 & 25 & 0.93 \\
Mistral-7B-v0.3 & default & instance & \texttt{sPCA+CP} & City Locations & 0.70 & \textbf{0.73} & 0.75 & 19 & 0.61 \\
Qwen-2.5-7B & default & instance & \texttt{sPCA+CP} & City Locations & 0.67 & \textbf{0.69} & 0.72 & 21 & 0.78 \\
Qwen-2.5-14B & default & instance & \texttt{sPCA+CP} & City Locations & 0.71 & \textbf{0.73} & 0.76 & 25 & 0.53 \\
Gemma-7B & default & instance & \texttt{sPCA+CP} & City Locations & 0.70 & \textbf{0.73} & 0.75 & 18 & 0.67 \\
Gemma-2-9B & default & instance & \texttt{sPCA+CP} & City Locations & 0.68 & \textbf{0.71} & 0.73 & 21 & 0.51 \\
Llama-3.1-8B & chat & instance & \texttt{sPCA+CP} & City Locations & 0.77 & \textbf{0.80} & 0.82 & 22 & 0.71 \\
\bottomrule
\end{longtable}
\end{center}
\end{small}

\newpage
\begin{small}
\begin{center}
\begin{longtable}{@{}lllclcccp{0.7cm}c@{}}
  \toprule
   Model Name & Type & Setting & Probe & Dataset & CI$_{.025}$ & MCC & CI$_{.975}$ & Layer & Depth\\
  \midrule
  \endfirsthead

  \multicolumn{10}{c}%
    {{\bfseries \tablename\ \thetable{} — continued from previous page}} \\
  \toprule
    Model Name & Type & Setting & Probe & Dataset & CI$_{.025}$ & MCC & CI$_{.975}$ & L & Depth\\
  \midrule
  \endhead

  \midrule
  \multicolumn{10}{r}{{Continued on next page}} \\ 
  \endfoot

  \bottomrule
  \\[-2pt]
    \caption{ \textbf{Classification performance of the supervised PCA with conformal prediction (\texttt{sPCA+CP}) probe across datasets and models} (evaluated on the \underline{\textit{full bag}}). This probe is trained on the representation of the last tokens and evaluated on the bag representation of the statements. 
    We report Matthews Correlation Coefficient (MCC) with the \(95\% \) confidence intervals.
    Confidence intervals are based on bootstrapping with \(n=1{,}000\) samples.
    The \textbf{bold} values mark MCC with the confidence intervals that exclude zero.
    The `Setting' column specifies the evaluation setting.
    The `Layer' column specifies the layer at which the probe achieved the highest MCC, and `Depth' denotes the relative depth of that layer within the model.
  \label{sup_tab:mcc_long_spca_bag}}\\
  \endlastfoot
  
Llama-3-8B & default & bag & \texttt{sPCA+CP} & City Locations & 0.38 & \textbf{0.40} & 0.43 & 8 & 0.26 \\
Llama-3.2-3B & default & bag & \texttt{sPCA+CP} & City Locations & 0.55 & \textbf{0.57} & 0.60 & 27 & 1.00 \\
Mistral-7B-v0.3 & default & bag & \texttt{sPCA+CP} & City Locations & 0.00 & 0.02 & 0.04 & 6 & 0.19 \\
Qwen-2.5-7B & default & bag & \texttt{sPCA+CP} & City Locations & 0.44 & \textbf{0.47} & 0.49 & 15 & 0.56 \\
Qwen-2.5-14B & default & bag & \texttt{sPCA+CP} & City Locations & 0.47 & \textbf{0.51} & 0.55 & 37 & 0.79 \\
Gemma-7B & default & bag & \texttt{sPCA+CP} & City Locations & 0.01 & \textbf{0.04} & 0.07 & 6 & 0.22 \\
Gemma-2-9B & default & bag & \texttt{sPCA+CP} & City Locations & 0.11 & \textbf{0.14} & 0.17 & 32 & 0.78 \\
Llama-3.1-8B & chat & bag & \texttt{sPCA+CP} & City Locations & 0.74 & \textbf{0.78} & 0.82 & 16 & 0.52 \\
\bottomrule
\end{longtable}
\end{center}
\end{small}

\newpage
\begin{small}
\begin{center}
\begin{longtable}{@{}lllclcccp{0.7cm}c@{}}
  \toprule
   Model Name & Type & Setting & Probe & Dataset & CI$_{.025}$ & MCC & CI$_{.975}$ & Layer & Depth\\
  \midrule
  \endfirsthead

  \multicolumn{10}{c}%
    {{\bfseries \tablename\ \thetable{} — continued from previous page}} \\
  \toprule
    Model Name & Type & Setting & Probe & Dataset & CI$_{.025}$ & MCC & CI$_{.975}$ & L & Depth\\
  \midrule
  \endhead

  \midrule
  \multicolumn{10}{r}{{Continued on next page}} \\ 
  \endfoot

  \bottomrule
  \\[-2pt]
    \caption{ \textbf{Classification performance of the multiclass \texttt{SVM} probe across datasets and models} (evaluated on the \underline{\textit{last token's representation}}). This probe is trained and evaluated using the representations of the last tokens. 
    We report Matthews Correlation Coefficient (MCC) with the \(95\% \) confidence intervals.
    Confidence intervals are based on bootstrapping with \(n=1{,}000\) samples.
    The \textbf{bold} values mark MCC with the confidence intervals that exclude zero.
    The `Setting' column specifies the evaluation setting.
    The `Layer' column specifies the layer at which the probe achieved the highest MCC, and `Depth' denotes the relative depth of that layer within the model.
  \label{sup_tab:mcc_long_svm_instance}}\\
  \endlastfoot

Llama-3.2-3B & default & instance & \texttt{SVM} & City Locations & 0.95 & \textbf{0.96} & 0.97 & 9 & 0.33 \\
Mistral-7B-v0.3 & default & instance & \texttt{SVM} & City Locations & 0.96 & \textbf{0.97} & 0.98 & 12 & 0.39 \\
Qwen-2.5-7B & default & instance & \texttt{SVM} & City Locations & 0.96 & \textbf{0.97} & 0.98 & 18 & 0.67 \\
Qwen-2.5-14B & default & instance & \texttt{SVM} & City Locations & 0.97 & \textbf{0.97} & 0.98 & 24 & 0.51 \\
Gemma-7B & default & instance & \texttt{SVM} & City Locations & 0.97 & \textbf{0.98} & 0.99 & 15 & 0.56 \\
Gemma-2-9B & default & instance & \texttt{SVM} & City Locations & 0.97 & \textbf{0.98} & 0.99 & 21 & 0.51 \\
Llama-3.1-8B & chat & instance & \texttt{SVM} & City Locations & 0.97 & \textbf{0.98} & 0.99 & 12 & 0.39 \\
Llama-3.2-3B & chat & instance & \texttt{SVM} & City Locations & 0.94 & \textbf{0.95} & 0.97 & 13 & 0.48 \\
\bottomrule
\end{longtable}
\end{center}
\end{small}

\newpage
\begin{small}
\begin{center}
\begin{longtable}{@{}lllclcccp{0.7cm}c@{}}
  \toprule
   Model Name & Type & Setting & Probe & Dataset & CI$_{.025}$ & MCC & CI$_{.975}$ & Layer & Depth\\
  \midrule
  \endfirsthead

  \multicolumn{10}{c}%
    {{\bfseries \tablename\ \thetable{} — continued from previous page}} \\
  \toprule
    Model Name & Type & Setting & Probe & Dataset & CI$_{.025}$ & MCC & CI$_{.975}$ & L & Depth\\
  \midrule
  \endhead

  \midrule
  \multicolumn{10}{r}{{Continued on next page}} \\ 
  \endfoot

  \bottomrule
  \\[-2pt]
    \caption{ \textbf{Classification performance of the multiclass \texttt{SVM} probe across datasets and models} (evaluated on the \underline{\textit{full bag}}). This probe is trained on the representation of the last tokens and evaluated on the bag representation of the statements.
    We report Matthews Correlation Coefficient (MCC) with the \(95\%\) confidence intervals.
    Confidence intervals are based on bootstrapping with \(n=1{,}000\) samples.
    The \textbf{bold} values mark MCC with the confidence intervals that exclude zero.
    The `Setting' column specifies the evaluation setting.
    The `Layer' column specifies the layer at which the probe achieved the highest MCC, and `Depth' denotes the relative depth of that layer within the model.
  \label{sup_tab:mcc_long_svm_bag}}\\
  \endlastfoot
  
Llama-3-8B & default & bag & \texttt{SVM} & City Locations & 0.52 & \textbf{0.54} & 0.57 & 8 & 0.26 \\
Llama-3.2-3B & default & bag & \texttt{SVM} & City Locations & 0.71 & \textbf{0.73} & 0.75 & 27 & 1.00 \\
Mistral-7B-v0.3 & default & bag & \texttt{SVM} & City Locations & 0.86 & \textbf{0.88} & 0.89 & 10 & 0.32 \\
Qwen-2.5-7B & default & bag & \texttt{SVM} & City Locations & 0.50 & \textbf{0.52} & 0.54 & 26 & 0.96 \\
Qwen-2.5-14B & default & bag & \texttt{SVM} & City Locations & 0.38 & \textbf{0.41} & 0.43 & 40 & 0.85 \\
Gemma-7B & default & bag & \texttt{SVM} & City Locations & 0.38 & \textbf{0.40} & 0.42 & 11 & 0.41 \\
Gemma-2-9B & default & bag & \texttt{SVM} & City Locations & 0.84 & \textbf{0.86} & 0.88 & 17 & 0.41 \\
Llama-3.1-8B & chat & bag & \texttt{SVM} & City Locations & 0.00 & 0.00 & 0.00 & 2 & 0.06 \\
\bottomrule
\end{longtable}
\end{center}
\end{small}

\newpage
\begin{small}
\begin{table}[t]
\centering
\begin{tabular}{llrcc}
\toprule
Evaluation Setting & Probe & Training Dataset & MCC & Stand. Err. \\
\midrule
Bag-Level & \texttt{MD+CP} & City Locations & 0.02 & 0.00 \\
Bag-Level & \texttt{MD+CP} & Medical Indications & 0.03 & 0.01 \\
Bag-Level & \texttt{MD+CP} & Word Definitions & 0.04 & 0.02 \\ [2pt]
Bag-Level & \texttt{SVM} & City Locations & 0.24 & 0.03 \\
Bag-Level & \texttt{SVM} & Medical Indications & 0.37 & 0.04 \\
Bag-Level & \texttt{SVM} & Word Definitions & 0.40 & 0.04 \\ [2pt]
Bag-Level & \texttt{TTPD+CP} & City Locations & 0.11 & 0.03 \\
\bottomrule
\end{tabular}
    \caption{\textbf{Aggregated generalization performance.} Each probe is trained on one dataset and evaluated on the remaining two (e.g., a probe trained on \textit{City Locations} is evaluated on \textit{Medical Indications} and \textit{Word Definitions}). The performance is averaged over all large language models (we pick only the performance from the best-performing layers). The performance is measured by the Matthews Correlation Coefficient (MCC) with the standard error. 
    The `Evaluation Setting' column indicates whether the probe was evaluated using only the representation of the last token in the statement (\textit{Instance-Level}) or the full bag of tokens (\textit{Bag-Level}). 
    Multiclass probes (i.e., SVM and \texttt{sAwMIL}) provide higher generalization performance in both settings, and the multiclass \texttt{sAwMIL} achieves the highest generalization performance.}
\label{sup_tab:aggregated_generalization}
\end{table}
\end{small}

\clearpage

\newpage
\begin{small}
\begin{center}
\begin{longtable}{@{}lclcccc@{}}
  \toprule
  Model Name               & Training Dataset    & Test Dataset    & CI$_{.025}$ & MCC & CI$_{.975}$ & Rel.\ Depth \\
  \midrule
  \endfirsthead

  \multicolumn{7}{c}{\tablename\ \thetable{} (continued)} \\
  \toprule
  Model Name               & Training Dataset    & Test Dataset         &  CI$_{.025}$ & MCC &  CI$_{.975}$ & Rel.\ Depth \\
  \midrule
  \endhead

  \midrule \multicolumn{7}{r}{Continued on next page} \\ 
  \endfoot

  \bottomrule
  \\[-2pt]
    \caption{\small \textbf{Generalization performance of the multiclass \texttt{sAwMIL} trained on the \textit{City Locations} dataset}. The performance is measured by the Matthews Correlation Coefficient (MCC) with \(95\%\) confidence intervals, based on bootstrapping with \(n=1{,}000\) samples.
    The \textbf{bold} values mark MCC with the confidence intervals that exclude zero.
    The `Rel. Depth' column specifies the relative depth of the layer where the multiclass \texttt{sAwMIL} probe achieves the best MCC score.
  }
  \label{tab:sup_generalization_cities}\\
  \endlastfoot

Gemma-7B-it & City Locations & City Locations & 0.93 & \textbf{0.94} & 0.95 & 0.59 \\
Gemma-7B-it & City Locations & Medical Indications & 0.43 & \textbf{0.46} & 0.49 & 0.63 \\
Gemma-7B-it & City Locations & Word Definitions & 0.44 & \textbf{0.47} & 0.50 & 0.63 \\ [2pt]
Gemma-2-9B-it & City Locations & City Locations & 0.96 & \textbf{0.97} & 0.98 & 0.56 \\
Gemma-2-9B-it & City Locations & Medical Indications & 0.67 & \textbf{0.70} & 0.73 & 0.44 \\
Gemma-2-9B-it & City Locations & Word Definitions & 0.69 & \textbf{0.71} & 0.74 & 0.49 \\ [2pt]
Llama-3.2-3B-Instruct & City Locations & City Locations & 0.95 & \textbf{0.96} & 0.97 & 0.52 \\
Llama-3.2-3B-Instruct & City Locations & Medical Indications & 0.59 & \textbf{0.62} & 0.65 & 0.56 \\
Llama-3.2-3B-Instruct & City Locations & Word Definitions & 0.62 & \textbf{0.65} & 0.67 & 0.48 \\ [2pt]
Llama3-Med42-8B & City Locations & City Locations & 0.96 & \textbf{0.97} & 0.98 & 0.42 \\
Llama3-Med42-8B & City Locations & Medical Indications & 0.75 & \textbf{0.78} & 0.81 & 0.90 \\
Llama3-Med42-8B & City Locations & Word Definitions & 0.75 & \textbf{0.78} & 0.80 & 0.45 \\ [2pt]
Llama-3.1-8B-Instruct & City Locations & City Locations & 0.96 & \textbf{0.97} & 0.98 & 0.48 \\
Llama-3.1-8B-Instruct & City Locations & Medical Indications & 0.73 & \textbf{0.75} & 0.78 & 0.52 \\
Llama-3.1-8B-Instruct & City Locations & Word Definitions & 0.79 & \textbf{0.81} & 0.83 & 0.42 \\ [2pt]
Bio-Medical-Llama-3-8B & City Locations & City Locations & 0.96 & \textbf{0.97} & 0.98 & 0.97 \\
Bio-Medical-Llama-3-8B & City Locations & Medical Indications & 0.59 & \textbf{0.62} & 0.65 & 0.42 \\
Bio-Medical-Llama-3-8B & City Locations & Word Definitions & 0.53 & \textbf{0.56} & 0.59 & 0.26 \\ [2pt]
Mistral-7B-Instruct-v0.3 & City Locations & City Locations & 0.95 & \textbf{0.96} & 0.97 & 0.48 \\
Mistral-7B-Instruct-v0.3 & City Locations & Medical Indications & 0.72 & \textbf{0.75} & 0.78 & 0.55 \\
Mistral-7B-Instruct-v0.3 & City Locations & Word Definitions & 0.69 & \textbf{0.71} & 0.74 & 0.35 \\ [2pt]
Qwen-2.5-7B-Instruct & City Locations & City Locations & 0.94 & \textbf{0.95} & 0.96 & 0.67 \\
Qwen-2.5-7B-Instruct & City Locations & Medical Indications & 0.71 & \textbf{0.74} & 0.76 & 0.70 \\
Qwen-2.5-7B-Instruct & City Locations & Word Definitions & 0.62 & \textbf{0.64} & 0.67 & 0.70 \\ [2pt]
Qwen-2.5-14B-Instruct & City Locations & City Locations & 0.96 & \textbf{0.97} & 0.98 & 0.62 \\
Qwen-2.5-14B-Instruct & City Locations & Medical Indications & 0.74 & \textbf{0.77} & 0.80 & 0.64 \\ 
Qwen-2.5-14B-Instruct & City Locations & Word Definitions & 0.73 & \textbf{0.75} & 0.77 & 0.60 \\ [2pt]
Gemma-7B & City Locations & City Locations & 0.96 & \textbf{0.97} & 0.98 & 0.74 \\
Gemma-7B & City Locations & Medical Indications & 0.55 & \textbf{0.58} & 0.60 & 0.41 \\
Gemma-7B & City Locations & Word Definitions & 0.61 & \textbf{0.63} & 0.66 & 0.52 \\ [2pt]
Gemma-2-9B & City Locations & City Locations & 0.97 & \textbf{0.98} & 0.99 & 0.63 \\
Gemma-2-9B & City Locations & Medical Indications & 0.52 & \textbf{0.55} & 0.58 & 0.44 \\
Gemma-2-9B & City Locations & Word Definitions & 0.54 & \textbf{0.56} & 0.58 & 0.27 \\ [2pt]
Llama-3.2-3B & City Locations & City Locations & 0.95 & \textbf{0.96} & 0.97 & 0.37 \\
Llama-3.2-3B & City Locations & Medical Indications & 0.35 & \textbf{0.38} & 0.41 & 0.33 \\
Llama-3.2-3B & City Locations & Word Definitions & 0.48 & \textbf{0.50} & 0.52 & 0.30 \\ [2pt]
Llama-3-8B & City Locations & City Locations & 0.96 & \textbf{0.97} & 0.98 & 0.32 \\
Llama-3-8B & City Locations & Medical Indications & 0.59 & \textbf{0.62} & 0.65 & 0.35 \\
Llama-3-8B & City Locations & Word Definitions & 0.56 & \textbf{0.58} & 0.61 & 0.26 \\ [2pt]
Mistral-7B-v0.3 & City Locations & City Locations & 0.96 & \textbf{0.97} & 0.98 & 0.42 \\
Mistral-7B-v0.3 & City Locations & Medical Indications & 0.50 & \textbf{0.53} & 0.55 & 0.39 \\
Mistral-7B-v0.3 & City Locations & Word Definitions & 0.58 & \textbf{0.61} & 0.63 & 0.39 \\ [2pt]
Qwen-2.5-7B & City Locations & City Locations & 0.94 & \textbf{0.95} & 0.96 & 0.70 \\
Qwen-2.5-7B & City Locations & Medical Indications & 0.48 & \textbf{0.52} & 0.55 & 0.59 \\
Qwen-2.5-7B & City Locations & Word Definitions & 0.51 & \textbf{0.54} & 0.57 & 0.59 \\ [2pt]
Qwen-2.5-14B & City Locations & City Locations & 0.95 & \textbf{0.96} & 0.97 & 0.79 \\
Qwen-2.5-14B & City Locations & Medical Indications & 0.59 & \textbf{0.62} & 0.65 & 0.45 \\
Qwen-2.5-14B & City Locations & Word Definitions & 0.62 & \textbf{0.64} & 0.67 & 0.43 \\
\end{longtable}
\end{center}
\end{small}

\newpage
\begin{small}
\begin{center}
\begin{longtable}{@{}lclcccc@{}}
  \toprule
  Model Name               & Training Dataset    & Test Dataset         &   CI$_{.025}$ & MCC & CI$_{.975}$ & Rel.\ Depth \\
  \midrule
  \endfirsthead

  \multicolumn{7}{c}{\tablename\ \thetable{} (continued)} \\
  \toprule
  Model Name               & Training Dataset    & Test Dataset         &   CI$_{.025}$ & MCC & CI$_{.975}$ & Rel.\ Depth \\
  \midrule
  \endhead

  \midrule \multicolumn{7}{r}{Continued on next page} \\ 
  \endfoot

  \bottomrule
  \\[-2pt]
  \caption{\small \textbf{Generalization performance of the multiclass \texttt{sAwMIL} trained on the \textit{Medical Indications} dataset.} The performance is measured by the Matthews Correlation Coefficient (MCC) with \(95\%\) confidence intervals, based on bootstrapping with \(n=1{,}000\) samples.
  The \textbf{bold} values mark MCC with the confidence intervals that exclude zero.
  The `Rel. Depth' column specifies the relative depth of the layer where a multiclass \texttt{sAwMIL} probe achieves the best MCC score.
  }
  \label{tab:sup_generalization_indications}\\
  \endlastfoot

Gemma-7B-it & Medical Indications & City Locations & 0.90 & \textbf{0.91} & 0.93 & 0.70 \\
Gemma-7B-it & Medical Indications & Medical Indications & 0.65 & \textbf{0.68} & 0.72 & 0.59 \\
Gemma-7B-it & Medical Indications & Word Definitions & 0.64 & \textbf{0.67} & 0.69 & 0.59 \\ [2pt]
Gemma-2-9B-it & Medical Indications & City Locations & 0.91 & \textbf{0.92} & 0.94 & 0.63 \\
Gemma-2-9B-it & Medical Indications & Medical Indications & 0.81 & \textbf{0.83} & 0.85 & 0.49 \\
Gemma-2-9B-it & Medical Indications & Word Definitions & 0.78 & \textbf{0.80} & 0.82 & 0.61 \\ [2pt]
Llama-3.2-3B-Instruct & Medical Indications & City Locations & 0.80 & \textbf{0.82} & 0.85 & 0.41 \\
Llama-3.2-3B-Instruct & Medical Indications & Medical Indications & 0.73 & \textbf{0.76} & 0.79 & 0.48 \\
Llama-3.2-3B-Instruct & Medical Indications & Word Definitions & 0.65 & \textbf{0.67} & 0.70 & 0.48 \\ [2pt]
Llama3-Med42-8B & Medical Indications & City Locations & 0.95 & \textbf{0.96} & 0.97 & 0.52 \\
Llama3-Med42-8B & Medical Indications & Medical Indications & 0.80 & \textbf{0.83} & 0.85 & 0.45 \\
Llama3-Med42-8B & Medical Indications & Word Definitions & 0.76 & \textbf{0.78} & 0.80 & 0.45 \\ [2pt]
Llama-3.1-8B-Instruct & Medical Indications & City Locations & 0.92 & \textbf{0.93} & 0.94 & 0.55 \\
Llama-3.1-8B-Instruct & Medical Indications & Medical Indications & 0.81 & \textbf{0.83} & 0.85 & 0.55 \\
Llama-3.1-8B-Instruct & Medical Indications & Word Definitions & 0.77 & \textbf{0.79} & 0.81 & 0.42 \\ [2pt]
Bio-Medical-Llama-3-8B & Medical Indications & City Locations & 0.85 & \textbf{0.86} & 0.88 & 0.39 \\
Bio-Medical-Llama-3-8B & Medical Indications & Medical Indications & 0.78 & \textbf{0.81} & 0.83 & 0.81 \\
Bio-Medical-Llama-3-8B & Medical Indications & Word Definitions & 0.70 & \textbf{0.73} & 0.75 & 0.32 \\ [2pt]
Mistral-7B-Instruct-v0.3 & Medical Indications & City Locations & 0.93 & \textbf{0.94} & 0.95 & 0.39 \\
Mistral-7B-Instruct-v0.3 & Medical Indications & Medical Indications & 0.78 & \textbf{0.80} & 0.83 & 0.45 \\
Mistral-7B-Instruct-v0.3 & Medical Indications & Word Definitions & 0.74 & \textbf{0.77} & 0.79 & 0.45 \\ [2pt]
Qwen-2.5-7B-Instruct & Medical Indications & City Locations & 0.78 & \textbf{0.80} & 0.83 & 0.52 \\
Qwen-2.5-7B-Instruct & Medical Indications & Medical Indications & 0.74 & \textbf{0.77} & 0.79 & 0.67 \\
Qwen-2.5-7B-Instruct & Medical Indications & Word Definitions & 0.64 & \textbf{0.67} & 0.69 & 0.63 \\ [2pt]
Qwen-2.5-14B-Instruct & Medical Indications & City Locations & 0.90 & \textbf{0.92} & 0.93 & 0.49 \\
Qwen-2.5-14B-Instruct & Medical Indications & Medical Indications & 0.79 & \textbf{0.82} & 0.84 & 0.57 \\
Qwen-2.5-14B-Instruct & Medical Indications & Word Definitions & 0.76 & \textbf{0.79} & 0.81 & 0.57 \\ [2pt]
Gemma-7B & Medical Indications & City Locations & 0.60 & \textbf{0.63} & 0.66 & 0.56 \\
Gemma-7B & Medical Indications & Medical Indications & 0.75 & \textbf{0.78} & 0.80 & 0.63 \\
Gemma-7B & Medical Indications & Word Definitions & 0.55 & \textbf{0.57} & 0.59 & 0.70 \\ [2pt]
Gemma-2-9B & Medical Indications & City Locations & 0.83 & \textbf{0.85} & 0.87 & 0.56 \\
Gemma-2-9B & Medical Indications & Medical Indications & 0.77 & \textbf{0.80} & 0.82 & 0.44 \\
Gemma-2-9B & Medical Indications & Word Definitions & 0.64 & \textbf{0.67} & 0.69 & 0.39 \\ [2pt]
Llama-3.2-3B & Medical Indications & City Locations & 0.49 & \textbf{0.51} & 0.53 & 0.41 \\
Llama-3.2-3B & Medical Indications & Medical Indications & 0.74 & \textbf{0.76} & 0.79 & 0.44 \\
Llama-3.2-3B & Medical Indications & Word Definitions & 0.59 & \textbf{0.62} & 0.64 & 0.52 \\ [2pt]
Llama-3-8B & Medical Indications & City Locations & 0.75 & \textbf{0.77} & 0.80 & 0.45 \\
Llama-3-8B & Medical Indications & Medical Indications & 0.77 & \textbf{0.80} & 0.83 & 0.39 \\
Llama-3-8B & Medical Indications & Word Definitions & 0.63 & \textbf{0.65} & 0.68 & 0.26 \\ [2pt]
Mistral-7B-v0.3 & Medical Indications & City Locations & 0.58 & \textbf{0.60} & 0.63 & 0.42 \\
Mistral-7B-v0.3 & Medical Indications & Medical Indications & 0.77 & \textbf{0.80} & 0.82 & 0.42 \\
Mistral-7B-v0.3 & Medical Indications & Word Definitions & 0.63 & \textbf{0.65} & 0.68 & 0.42 \\ [2pt]
Qwen-2.5-7B & Medical Indications & City Locations & 0.80 & \textbf{0.82} & 0.84 & 0.67 \\
Qwen-2.5-7B & Medical Indications & Medical Indications & 0.76 & \textbf{0.78} & 0.81 & 0.63 \\
Qwen-2.5-7B & Medical Indications & Word Definitions & 0.59 & \textbf{0.62} & 0.65 & 0.74 \\ [2pt]
Qwen-2.5-14B & Medical Indications & City Locations & 0.80 & \textbf{0.82} & 0.84 & 0.70 \\
Qwen-2.5-14B & Medical Indications & Medical Indications & 0.76 & \textbf{0.79} & 0.82 & 0.60 \\
Qwen-2.5-14B & Medical Indications & Word Definitions & 0.70 & \textbf{0.72} & 0.75 & 0.60 \\
\end{longtable}
\end{center}
\end{small}

\newpage
\begin{small}
\begin{center}
\begin{longtable}{@{}lllcccc@{}}
  \toprule
  Model Name               & Training Dataset    & Test Dataset         &   CI$_{.025}$ & MCC & CI$_{.975}$ & Rel.\ Depth \\
  \midrule
  \endfirsthead

  \multicolumn{7}{c}{\tablename\ \thetable{} (continued)} \\
  \toprule
  Model Name               & Training Dataset    & Test Dataset         &  CI$_{.025}$ &  MCC & CI$_{.975}$ & Rel.\ Depth \\
  \midrule
  \endhead

  \midrule \multicolumn{7}{r}{Continued on next page} \\ 
  \endfoot

  \bottomrule
  \\[-2pt]
  \caption{\small \textbf{Generalization performance of the multiclass \texttt{sAwMIL} trained on the \textit{Word Definitions} dataset.} 
  The performance is measured by the Matthews Correlation Coefficient (MCC) with \(95\%\) confidence intervals, based on bootstrapping with \(n=1{,}000\) samples. 
  The \textbf{bold} values mark MCC with the confidence intervals that exclude zero. 
  The `Rel. Depth' column specifies the relative depth of the layer where a multiclass \texttt{sAwMIL} probe achieves the best MCC score.
  }
  \label{tab:sup_generalization_defs}\\
  \endlastfoot

Gemma-7B-it & Word Definitions & City Locations & 0.90 & \textbf{0.92} & 0.93 & 0.70 \\
Gemma-7B-it & Word Definitions & Medical Indications & 0.53 & \textbf{0.56} & 0.60 & 0.67 \\
Gemma-7B-it & Word Definitions & Word Definitions & 0.78 & \textbf{0.80} & 0.82 & 0.67 \\ [2pt]
Gemma-2-9B-it & Word Definitions & City Locations & 0.94 & \textbf{0.96} & 0.97 & 0.56 \\
Gemma-2-9B-it & Word Definitions & Medical Indications & 0.65 & \textbf{0.69} & 0.71 & 0.41 \\
Gemma-2-9B-it & Word Definitions & Word Definitions & 0.88 & \textbf{0.90} & 0.91 & 0.54 \\ [2pt]
Llama-3.2-3B-Instruct & Word Definitions & City Locations & 0.84 & \textbf{0.86} & 0.88 & 0.48 \\
Llama-3.2-3B-Instruct & Word Definitions & Medical Indications & 0.60 & \textbf{0.63} & 0.66 & 0.44 \\ 
Llama-3.2-3B-Instruct & Word Definitions & Word Definitions & 0.85 & \textbf{0.86} & 0.88 & 0.44 \\ [2pt]
Llama3-Med42-8B & Word Definitions & City Locations & 0.94 & \textbf{0.95} & 0.97 & 0.35 \\
Llama3-Med42-8B & Word Definitions & Medical Indications & 0.76 & \textbf{0.79} & 0.81 & 0.45 \\
Llama3-Med42-8B & Word Definitions & Word Definitions & 0.87 & \textbf{0.89} & 0.91 & 0.45 \\ [2pt]
Llama-3.1-8B-Instruct & Word Definitions & City Locations & 0.95 & \textbf{0.96} & 0.97 & 0.45 \\
Llama-3.1-8B-Instruct & Word Definitions & Medical Indications & 0.72 & \textbf{0.75} & 0.77 & 0.32 \\
Llama-3.1-8B-Instruct & Word Definitions & Word Definitions & 0.90 & \textbf{0.91} & 0.93 & 0.45 \\ [2pt]
Bio-Medical-Llama-3-8B & Word Definitions & City Locations & 0.78 & \textbf{0.81} & 0.83 & 0.35 \\
Bio-Medical-Llama-3-8B & Word Definitions & Medical Indications & 0.70 & \textbf{0.73} & 0.76 & 0.32 \\
Bio-Medical-Llama-3-8B & Word Definitions & Word Definitions & 0.87 & \textbf{0.89} & 0.90 & 0.39 \\ [2pt]
Mistral-7B-Instruct-v0.3 & Word Definitions & City Locations & 0.93 & \textbf{0.94} & 0.96 & 0.48 \\
Mistral-7B-Instruct-v0.3 & Word Definitions & Medical Indications & 0.74 & \textbf{0.77} & 0.80 & 0.52 \\
Mistral-7B-Instruct-v0.3 & Word Definitions & Word Definitions & 0.87 & \textbf{0.88} & 0.90 & 0.35 \\ [2pt]
Qwen-2.5-7B-Instruct & Word Definitions & City Locations & 0.85 & \textbf{0.87} & 0.88 & 0.63 \\
Qwen-2.5-7B-Instruct & Word Definitions & Medical Indications & 0.74 & \textbf{0.76} & 0.79 & 0.67 \\
Qwen-2.5-7B-Instruct & Word Definitions & Word Definitions & 0.85 & \textbf{0.87} & 0.88 & 0.63 \\ [2pt]
Qwen-2.5-14B-Instruct & Word Definitions & City Locations & 0.95 & \textbf{0.96} & 0.97 & 0.66 \\
Qwen-2.5-14B-Instruct & Word Definitions & Medical Indications & 0.77 & \textbf{0.79} & 0.82 & 0.66 \\
Qwen-2.5-14B-Instruct & Word Definitions & Word Definitions & 0.88 & \textbf{0.90} & 0.92 & 0.51 \\ [2pt]
Gemma-7B & Word Definitions & City Locations & 0.88 & \textbf{0.90} & 0.91 & 0.56 \\
Gemma-7B & Word Definitions & Medical Indications & 0.59 & \textbf{0.63} & 0.66 & 0.48 \\
Gemma-7B & Word Definitions & Word Definitions & 0.81 & \textbf{0.83} & 0.85 & 0.56 \\ [2pt]
Gemma-2-9B & Word Definitions & City Locations & 0.92 & \textbf{0.93} & 0.94 & 0.56 \\
Gemma-2-9B & Word Definitions & Medical Indications & 0.70 & \textbf{0.73} & 0.75 & 0.46 \\
Gemma-2-9B & Word Definitions & Word Definitions & 0.86 & \textbf{0.88} & 0.90 & 0.41 \\ [2pt]
Llama-3.2-3B & Word Definitions & City Locations & 0.76 & \textbf{0.78} & 0.81 & 0.41 \\
Llama-3.2-3B & Word Definitions & Medical Indications & 0.70 & \textbf{0.73} & 0.76 & 0.41 \\
Llama-3.2-3B & Word Definitions & Word Definitions & 0.79 & \textbf{0.81} & 0.83 & 0.41 \\ [2pt]
Llama-3-8B & Word Definitions & City Locations & 0.84 & \textbf{0.86} & 0.88 & 0.39 \\
Llama-3-8B & Word Definitions & Medical Indications & 0.74 & \textbf{0.76} & 0.79 & 0.39 \\
Llama-3-8B & Word Definitions & Word Definitions & 0.85 & \textbf{0.87} & 0.89 & 0.39 \\ [2pt]
Mistral-7B-v0.3 & Word Definitions & City Locations & 0.77 & \textbf{0.79} & 0.81 & 0.52 \\
Mistral-7B-v0.3 & Word Definitions & Medical Indications & 0.73 & \textbf{0.76} & 0.78 & 0.39 \\
Mistral-7B-v0.3 & Word Definitions & Word Definitions & 0.85 & \textbf{0.87} & 0.89 & 0.45 \\ [2pt]
Qwen-2.5-7B & Word Definitions & City Locations & 0.90 & \textbf{0.92} & 0.93 & 0.67 \\ 
Qwen-2.5-7B & Word Definitions & Medical Indications & 0.72 & \textbf{0.74} & 0.77 & 0.59 \\
Qwen-2.5-7B & Word Definitions & Word Definitions & 0.85 & \textbf{0.87} & 0.88 & 0.59 \\ [2pt]
Qwen-2.5-14B & Word Definitions & City Locations & 0.92 & \textbf{0.94} & 0.95 & 0.45 \\ 
Qwen-2.5-14B & Word Definitions & Medical Indications & 0.72 & \textbf{0.75} & 0.78 & 0.66 \\
Qwen-2.5-14B & Word Definitions & Word Definitions & 0.85 & \textbf{0.87} & 0.88 & 0.47 \\
\end{longtable}
\end{center}
\end{small}

\newpage
\begin{table}
\centering
\begin{adjustbox}{width=1\textwidth}
\scriptsize
\begin{tabular}{ll|cccc|cccc|cccc}
\toprule
 & Ground-truth label \(\rightarrow\)
   & \multicolumn{4}{c|}{True } 
   & \multicolumn{4}{c|}{False} 
   & \multicolumn{4}{c}{Neither} \\
 & Predicted  \(\rightarrow\)
   & True & False & Neither & Abstain 
   & True & False & Neither & Abstain 
   & True & False & Neither & Abstain 
\\
Model  \(\downarrow\)& Dataset \(\downarrow\)
   &  &  &  &  
   &  &  &  &  
   &  &  &  &   \\
\midrule
\multirow[c]{3}{*}{Bio-Medical-Llama-3-8B} & City Locations & 0.89 & 0.11 & 0.00 & 0.00 & 0.06 & 0.94 & 0.00 & 0.00 & 0.18 & 0.82 & 0.00 & 0.00 \\
 & Medical Indications & 0.79 & 0.21 & 0.00 & 0.00 & 0.32 & 0.68 & 0.00 & 0.00 & 0.27 & 0.73 & 0.00 & 0.00 \\
 & Word Definitions & 0.61 & 0.39 & 0.00 & 0.00 & 0.37 & 0.63 & 0.00 & 0.00 & 0.05 & 0.95 & 0.00 & 0.00 \\
\cline{1-14}
\multirow[c]{3}{*}{Gemma-2-9B} & City Locations & 0.50 & 0.10 & 0.39 & 0.01 & 0.03 & 0.53 & 0.43 & 0.01 & 0.05 & 0.00 & 0.93 & 0.02 \\
 & Medical Indications & 0.70 & 0.12 & 0.18 & 0.00 & 0.34 & 0.51 & 0.14 & 0.00 & 0.57 & 0.19 & 0.24 & 0.00 \\
 & Word Definitions & 0.36 & 0.20 & 0.40 & 0.05 & 0.17 & 0.26 & 0.53 & 0.04 & 0.14 & 0.12 & 0.72 & 0.01 \\
\cline{1-14}
\multirow[c]{3}{*}{Gemma-2-9B-it} & City Locations & 0.98 & 0.02 & 0.00 & 0.00 & 0.03 & 0.97 & 0.00 & 0.00 & 0.06 & 0.45 & 0.49 & 0.00 \\
 & Medical Indications & 0.87 & 0.12 & 0.01 & 0.00 & 0.25 & 0.75 & 0.00 & 0.00 & 0.23 & 0.59 & 0.18 & 0.00 \\
 & Word Definitions & 0.76 & 0.16 & 0.09 & 0.00 & 0.25 & 0.70 & 0.06 & 0.00 & 0.10 & 0.65 & 0.25 & 0.00 \\
\cline{1-14}
\multirow[c]{3}{*}{Gemma-7B} & City Locations & 0.00 & 0.00 & 0.00 & 1.00 & 0.00 & 0.00 & 0.00 & 1.00 & 0.00 & 0.00 & 0.00 & 1.00 \\
 & Medical Indications & 0.16 & 0.00 & 0.00 & 0.84 & 0.09 & 0.00 & 0.00 & 0.91 & 0.03 & 0.00 & 0.00 & 0.96 \\
 & Word Definitions & 0.03 & 0.00 & 0.00 & 0.97 & 0.03 & 0.01 & 0.00 & 0.96 & 0.00 & 0.00 & 0.00 & 1.00 \\
\cline{1-14}
\multirow[c]{3}{*}{Gemma-7B-it} & City Locations & 0.76 & 0.23 & 0.01 & 0.00 & 0.05 & 0.95 & 0.00 & 0.00 & 0.04 & 0.64 & 0.32 & 0.00 \\
 & Medical Indications & 0.69 & 0.30 & 0.01 & 0.00 & 0.27 & 0.73 & 0.00 & 0.00 & 0.37 & 0.61 & 0.02 & 0.00 \\
 & Word Definitions & 0.27 & 0.63 & 0.09 & 0.01 & 0.09 & 0.89 & 0.02 & 0.00 & 0.04 & 0.77 & 0.15 & 0.04 \\
\cline{1-14}
\multirow[c]{3}{*}{Llama-3-8B} & City Locations & 0.35 & 0.65 & 0.00 & 0.00 & 0.22 & 0.78 & 0.00 & 0.00 & 0.47 & 0.52 & 0.00 & 0.01 \\
 & Medical Indications & 0.33 & 0.67 & 0.00 & 0.00 & 0.08 & 0.92 & 0.00 & 0.00 & 0.19 & 0.81 & 0.00 & 0.00 \\
 & Word Definitions & 0.45 & 0.55 & 0.00 & 0.00 & 0.33 & 0.67 & 0.00 & 0.00 & 0.37 & 0.63 & 0.00 & 0.00 \\
\cline{1-14}
\multirow[c]{3}{*}{Llama-3.1-8B-Instruct} & City Locations & 0.95 & 0.05 & 0.00 & 0.00 & 0.03 & 0.97 & 0.00 & 0.00 & 0.08 & 0.65 & 0.27 & 0.00 \\
 & Medical Indications & 0.54 & 0.46 & 0.00 & 0.00 & 0.07 & 0.93 & 0.00 & 0.00 & 0.13 & 0.86 & 0.01 & 0.00 \\
 & Word Definitions & 0.54 & 0.46 & 0.00 & 0.00 & 0.21 & 0.79 & 0.00 & 0.00 & 0.06 & 0.94 & 0.00 & 0.00 \\
\cline{1-14}
\multirow[c]{3}{*}{Llama-3.2-3B} & City Locations & 0.29 & 0.71 & 0.00 & 0.00 & 0.11 & 0.89 & 0.00 & 0.00 & 0.43 & 0.57 & 0.00 & 0.00 \\
 & Medical Indications & 0.46 & 0.54 & 0.00 & 0.00 & 0.34 & 0.66 & 0.00 & 0.00 & 0.50 & 0.50 & 0.00 & 0.00 \\
 & Word Definitions & 0.48 & 0.52 & 0.00 & 0.00 & 0.44 & 0.56 & 0.00 & 0.00 & 0.46 & 0.54 & 0.00 & 0.00 \\
\cline{1-14}
\multirow[c]{3}{*}{Llama-3.2-3B-Instruct} & City Locations & 0.93 & 0.07 & 0.00 & 0.00 & 0.15 & 0.85 & 0.00 & 0.00 & 0.04 & 0.84 & 0.13 & 0.00 \\
 & Medical Indications & 0.35 & 0.65 & 0.00 & 0.00 & 0.07 & 0.93 & 0.00 & 0.00 & 0.15 & 0.85 & 0.00 & 0.00 \\
 & Word Definitions & 0.60 & 0.40 & 0.00 & 0.00 & 0.46 & 0.54 & 0.00 & 0.00 & 0.06 & 0.93 & 0.01 & 0.00 \\
\cline{1-14}
\multirow[c]{3}{*}{Llama3-Med42-8B} & City Locations & 0.96 & 0.04 & 0.00 & 0.00 & 0.04 & 0.96 & 0.00 & 0.00 & 0.22 & 0.64 & 0.14 & 0.00 \\
 & Medical Indications & 0.69 & 0.31 & 0.00 & 0.00 & 0.13 & 0.87 & 0.00 & 0.00 & 0.17 & 0.82 & 0.00 & 0.00 \\
 & Word Definitions & 0.47 & 0.52 & 0.01 & 0.00 & 0.18 & 0.81 & 0.01 & 0.00 & 0.08 & 0.89 & 0.03 & 0.00 \\
\cline{1-14}
\multirow[c]{3}{*}{Mistral-7B-Instruct-v0.3} & City Locations & 0.88 & 0.00 & 0.09 & 0.03 & 0.04 & 0.09 & 0.50 & 0.37 & 0.08 & 0.00 & 0.90 & 0.02 \\
 & Medical Indications & 0.47 & 0.04 & 0.49 & 0.00 & 0.07 & 0.09 & 0.85 & 0.00 & 0.03 & 0.00 & 0.97 & 0.00 \\
 & Word Definitions & 0.63 & 0.02 & 0.33 & 0.02 & 0.40 & 0.01 & 0.55 & 0.04 & 0.25 & 0.00 & 0.73 & 0.02 \\
\cline{1-14}
\multirow[c]{3}{*}{Mistral-7B-v0.3} & City Locations & 0.00 & 0.00 & 0.00 & 1.00 & 0.00 & 0.00 & 0.00 & 1.00 & 0.00 & 0.00 & 0.00 & 1.00 \\
 & Medical Indications & 0.00 & 0.00 & 0.00 & 1.00 & 0.00 & 0.00 & 0.00 & 1.00 & 0.00 & 0.00 & 0.00 & 1.00 \\
 & Word Definitions & 0.00 & 0.00 & 0.00 & 1.00 & 0.00 & 0.00 & 0.00 & 1.00 & 0.00 & 0.00 & 0.00 & 1.00 \\
\cline{1-14}
\multirow[c]{3}{*}{Qwen-2.5-14B} & City Locations & 0.95 & 0.05 & 0.00 & 0.00 & 0.02 & 0.98 & 0.00 & 0.00 & 0.04 & 0.47 & 0.49 & 0.00 \\
 & Medical Indications & 0.53 & 0.46 & 0.01 & 0.00 & 0.04 & 0.95 & 0.00 & 0.00 & 0.00 & 0.77 & 0.23 & 0.00 \\
 & Word Definitions & 0.50 & 0.41 & 0.09 & 0.00 & 0.05 & 0.90 & 0.05 & 0.00 & 0.00 & 0.53 & 0.47 & 0.00 \\
\cline{1-14}
\multirow[c]{3}{*}{Qwen-2.5-14B-Instruct} & City Locations & 0.93 & 0.07 & 0.00 & 0.00 & 0.02 & 0.98 & 0.00 & 0.00 & 0.00 & 0.28 & 0.71 & 0.00 \\
 & Medical Indications & 0.63 & 0.23 & 0.14 & 0.00 & 0.04 & 0.87 & 0.09 & 0.00 & 0.00 & 0.46 & 0.54 & 0.00 \\
 & Word Definitions & 0.55 & 0.32 & 0.13 & 0.00 & 0.06 & 0.89 & 0.06 & 0.00 & 0.01 & 0.37 & 0.62 & 0.00 \\
\cline{1-14}
\multirow[c]{3}{*}{Qwen-2.5-7B} & City Locations & 0.92 & 0.07 & 0.01 & 0.00 & 0.03 & 0.96 & 0.01 & 0.00 & 0.09 & 0.68 & 0.23 & 0.00 \\
 & Medical Indications & 0.56 & 0.44 & 0.00 & 0.00 & 0.11 & 0.89 & 0.00 & 0.00 & 0.25 & 0.75 & 0.00 & 0.00 \\
 & Word Definitions & 0.60 & 0.39 & 0.01 & 0.00 & 0.31 & 0.67 & 0.02 & 0.00 & 0.35 & 0.64 & 0.01 & 0.00 \\
\cline{1-14}
\multirow[c]{3}{*}{Qwen-2.5-7B-Instruct} & City Locations & 0.89 & 0.11 & 0.00 & 0.00 & 0.02 & 0.98 & 0.00 & 0.00 & 0.02 & 0.57 & 0.37 & 0.03 \\
 & Medical Indications & 0.41 & 0.56 & 0.02 & 0.00 & 0.04 & 0.94 & 0.02 & 0.00 & 0.01 & 0.92 & 0.06 & 0.01 \\
 & Word Definitions & 0.56 & 0.39 & 0.04 & 0.01 & 0.27 & 0.70 & 0.03 & 0.01 & 0.25 & 0.61 & 0.12 & 0.03 \\
\bottomrule
\end{tabular}
\end{adjustbox}
\caption{\textbf{Row-wise confusion matrices for zero-shot prompting across all $\langle$model, dataset$\rangle$ pairs} (evaluated on the full bag). 
Each row corresponds to a specific model and a dataset. 
Columns are grouped by the ground-truth labels (\textit{True}, \textit{False}, \textit{Neither}) with groups of subcolumns that specify the distribution of predictions (\textit{true}, \textit{false}, \textit{neither}, \textit{abstain}).  
For each statement in a dataset, the predicted class is the class with the highest probability (as estimated by zero-shot prompting).
The values in each group of four subcolumns sum to \(1\) because they are normalized counts.
For example, in the first row under the \textit{True} ground-truth label, we see that \textit{true} predictions have the value of \(0.89\); that is, \(89\%\) of all the true statements are classified as true.
}
\label{sup_tab:cm_zero_shot}
\end{table}

\begin{small}
\centering
\begin{table}
\centering
\begin{adjustbox}{width=1\textwidth}
\begin{tabular}{ll|cccc|cccc|cccc}
\toprule
 & True Labels \(\rightarrow\)
   & \multicolumn{4}{c|}{True } 
   & \multicolumn{4}{c|}{False} 
   & \multicolumn{4}{c}{Neither} \\
 & Predicted  \(\rightarrow\)
   & True & False & Neither & Abstain 
   & True & False & Neither & Abstain 
   & True & False & Neither & Abstain 
\\
Model  \(\downarrow\)& Dataset \(\downarrow\)
   &  &  &  &  
   &  &  &  &  
   &  &  &  &   \\
\midrule
\multirow[c]{3}{*}{Bio-Medical-Llama-3-8B} & City Locations & 0.89 & 0.01 & 0.10 & 0.00 & 0.01 & 0.92 & 0.07 & 0.00 & 0.34 & 0.23 & 0.42 & 0.00 \\
 & Medical Indications & 0.68 & 0.13 & 0.19 & 0.00 & 0.09 & 0.75 & 0.16 & 0.00 & 0.21 & 0.58 & 0.20 & 0.00 \\
 & Word Definitions & 0.73 & 0.09 & 0.18 & 0.00 & 0.09 & 0.72 & 0.18 & 0.00 & 0.42 & 0.36 & 0.22 & 0.00 \\
\cline{1-14}
\multirow[c]{3}{*}{Gemma-2-9B} & City Locations & 0.90 & 0.02 & 0.09 & 0.00 & 0.02 & 0.91 & 0.07 & 0.00 & 0.07 & 0.66 & 0.27 & 0.00 \\
 & Medical Indications & 0.78 & 0.12 & 0.10 & 0.00 & 0.13 & 0.77 & 0.10 & 0.00 & 0.15 & 0.77 & 0.09 & 0.00 \\
 & Word Definitions & 0.80 & 0.11 & 0.10 & 0.00 & 0.10 & 0.81 & 0.09 & 0.00 & 0.47 & 0.39 & 0.14 & 0.00 \\
\cline{1-14}
\multirow[c]{3}{*}{Gemma-2-9B-it} & City Locations & 0.89 & 0.09 & 0.02 & 0.00 & 0.05 & 0.93 & 0.02 & 0.00 & 0.79 & 0.19 & 0.02 & 0.00 \\
 & Medical Indications & 0.70 & 0.11 & 0.19 & 0.00 & 0.11 & 0.73 & 0.17 & 0.00 & 0.42 & 0.38 & 0.20 & 0.00 \\
 & Word Definitions & 0.75 & 0.11 & 0.14 & 0.00 & 0.11 & 0.75 & 0.14 & 0.00 & 0.36 & 0.42 & 0.23 & 0.00 \\
\cline{1-14}
\multirow[c]{3}{*}{Gemma-7B} & City Locations & 0.90 & 0.01 & 0.08 & 0.00 & 0.02 & 0.92 & 0.06 & 0.00 & 0.04 & 0.79 & 0.16 & 0.00 \\
 & Medical Indications & 0.71 & 0.10 & 0.18 & 0.00 & 0.09 & 0.72 & 0.19 & 0.00 & 0.19 & 0.53 & 0.28 & 0.00 \\
 & Word Definitions & 0.70 & 0.10 & 0.20 & 0.00 & 0.08 & 0.75 & 0.18 & 0.00 & 0.37 & 0.36 & 0.28 & 0.00 \\
\cline{1-14}
\multirow[c]{3}{*}{Gemma-7B-it} & City Locations & 0.94 & 0.03 & 0.03 & 0.00 & 0.06 & 0.90 & 0.04 & 0.00 & 0.32 & 0.54 & 0.14 & 0.00 \\
 & Medical Indications & 0.53 & 0.11 & 0.36 & 0.00 & 0.12 & 0.61 & 0.27 & 0.00 & 0.19 & 0.47 & 0.34 & 0.00 \\
 & Word Definitions & 0.75 & 0.11 & 0.13 & 0.00 & 0.11 & 0.78 & 0.11 & 0.00 & 0.31 & 0.48 & 0.22 & 0.00 \\
\cline{1-14}
\multirow[c]{3}{*}{Llama-3-8B} & City Locations & 0.90 & 0.01 & 0.08 & 0.00 & 0.00 & 0.92 & 0.08 & 0.00 & 0.66 & 0.17 & 0.17 & 0.00 \\
 & Medical Indications & 0.71 & 0.12 & 0.17 & 0.00 & 0.11 & 0.75 & 0.14 & 0.00 & 0.40 & 0.39 & 0.21 & 0.00 \\
 & Word Definitions & 0.73 & 0.09 & 0.17 & 0.00 & 0.07 & 0.74 & 0.19 & 0.00 & 0.48 & 0.31 & 0.22 & 0.00 \\
\cline{1-14}
\multirow[c]{3}{*}{Llama-3.1-8B-Instruct} & City Locations & 0.89 & 0.03 & 0.08 & 0.00 & 0.04 & 0.89 & 0.06 & 0.00 & 0.68 & 0.27 & 0.05 & 0.00 \\
 & Medical Indications & 0.74 & 0.14 & 0.13 & 0.00 & 0.10 & 0.79 & 0.12 & 0.00 & 0.22 & 0.67 & 0.11 & 0.00 \\
 & Word Definitions & 0.80 & 0.10 & 0.10 & 0.00 & 0.07 & 0.83 & 0.09 & 0.00 & 0.45 & 0.46 & 0.09 & 0.00 \\
\cline{1-14}
\multirow[c]{3}{*}{Llama-3.2-3B} & City Locations & 0.89 & 0.03 & 0.08 & 0.00 & 0.03 & 0.90 & 0.07 & 0.00 & 0.25 & 0.39 & 0.35 & 0.00 \\
 & Medical Indications & 0.72 & 0.12 & 0.16 & 0.00 & 0.12 & 0.71 & 0.18 & 0.00 & 0.19 & 0.64 & 0.17 & 0.00 \\
 & Word Definitions & 0.72 & 0.09 & 0.18 & 0.00 & 0.11 & 0.71 & 0.18 & 0.00 & 0.31 & 0.44 & 0.26 & 0.00 \\
\cline{1-14}
\multirow[c]{3}{*}{Llama-3.2-3B-Instruct} & City Locations & 0.87 & 0.05 & 0.08 & 0.00 & 0.04 & 0.90 & 0.06 & 0.00 & 0.85 & 0.06 & 0.09 & 0.00 \\
 & Medical Indications & 0.59 & 0.11 & 0.30 & 0.00 & 0.11 & 0.62 & 0.27 & 0.00 & 0.33 & 0.38 & 0.30 & 0.00 \\
 & Word Definitions & 0.59 & 0.10 & 0.31 & 0.00 & 0.09 & 0.62 & 0.29 & 0.00 & 0.37 & 0.40 & 0.24 & 0.00 \\
\cline{1-14}
\multirow[c]{3}{*}{Llama3-Med42-8B} & City Locations & 0.91 & 0.01 & 0.08 & 0.00 & 0.01 & 0.91 & 0.09 & 0.00 & 0.04 & 0.73 & 0.23 & 0.00 \\
 & Medical Indications & 0.77 & 0.14 & 0.09 & 0.00 & 0.10 & 0.79 & 0.11 & 0.00 & 0.21 & 0.70 & 0.09 & 0.00 \\
 & Word Definitions & 0.83 & 0.09 & 0.08 & 0.00 & 0.08 & 0.88 & 0.05 & 0.00 & 0.59 & 0.32 & 0.09 & 0.00 \\
\cline{1-14}
\multirow[c]{3}{*}{Mistral-7B-Instruct-v0.3} & City Locations & 0.88 & 0.01 & 0.11 & 0.00 & 0.01 & 0.90 & 0.09 & 0.00 & 0.45 & 0.40 & 0.16 & 0.00 \\
 & Medical Indications & 0.68 & 0.11 & 0.21 & 0.00 & 0.13 & 0.71 & 0.16 & 0.00 & 0.29 & 0.50 & 0.20 & 0.00 \\
 & Word Definitions & 0.77 & 0.13 & 0.10 & 0.00 & 0.11 & 0.81 & 0.08 & 0.00 & 0.24 & 0.60 & 0.15 & 0.00 \\
\cline{1-14}
\multirow[c]{3}{*}{Mistral-7B-v0.3} & City Locations & 0.88 & 0.02 & 0.10 & 0.00 & 0.02 & 0.91 & 0.07 & 0.00 & 0.31 & 0.45 & 0.24 & 0.00 \\
 & Medical Indications & 0.69 & 0.12 & 0.19 & 0.00 & 0.10 & 0.73 & 0.17 & 0.00 & 0.23 & 0.54 & 0.22 & 0.00 \\
 & Word Definitions & 0.72 & 0.08 & 0.20 & 0.00 & 0.08 & 0.77 & 0.15 & 0.00 & 0.42 & 0.35 & 0.24 & 0.00 \\
\cline{1-14}
\multirow[c]{3}{*}{Qwen-2.5-14B} & City Locations & 0.89 & 0.04 & 0.07 & 0.00 & 0.02 & 0.93 & 0.06 & 0.00 & 0.21 & 0.48 & 0.30 & 0.00 \\
 & Medical Indications & 0.72 & 0.15 & 0.13 & 0.00 & 0.11 & 0.76 & 0.14 & 0.00 & 0.54 & 0.25 & 0.21 & 0.00 \\
 & Word Definitions & 0.80 & 0.09 & 0.11 & 0.00 & 0.11 & 0.79 & 0.10 & 0.00 & 0.39 & 0.46 & 0.15 & 0.00 \\
\cline{1-14}
\multirow[c]{3}{*}{Qwen-2.5-14B-Instruct} & City Locations & 0.88 & 0.09 & 0.03 & 0.00 & 0.06 & 0.92 & 0.02 & 0.00 & 0.38 & 0.59 & 0.03 & 0.00 \\
 & Medical Indications & 0.77 & 0.12 & 0.11 & 0.00 & 0.11 & 0.79 & 0.10 & 0.00 & 0.46 & 0.44 & 0.10 & 0.00 \\
 & Word Definitions & 0.82 & 0.08 & 0.10 & 0.00 & 0.08 & 0.85 & 0.07 & 0.00 & 0.30 & 0.62 & 0.08 & 0.00 \\
\cline{1-14}
\multirow[c]{3}{*}{Qwen-2.5-7B} & City Locations & 0.91 & 0.03 & 0.06 & 0.00 & 0.03 & 0.91 & 0.06 & 0.00 & 0.34 & 0.55 & 0.12 & 0.00 \\
 & Medical Indications & 0.71 & 0.11 & 0.17 & 0.00 & 0.10 & 0.77 & 0.14 & 0.00 & 0.34 & 0.43 & 0.23 & 0.00 \\
 & Word Definitions & 0.74 & 0.09 & 0.16 & 0.00 & 0.10 & 0.74 & 0.16 & 0.00 & 0.32 & 0.43 & 0.25 & 0.00 \\
\cline{1-14}
\multirow[c]{3}{*}{Qwen-2.5-7B-Instruct} & City Locations & 0.90 & 0.03 & 0.07 & 0.00 & 0.02 & 0.92 & 0.06 & 0.00 & 0.62 & 0.24 & 0.14 & 0.00 \\
 & Medical Indications & 0.75 & 0.09 & 0.15 & 0.00 & 0.10 & 0.78 & 0.12 & 0.00 & 0.67 & 0.20 & 0.13 & 0.00 \\
 & Word Definitions & 0.78 & 0.10 & 0.12 & 0.00 & 0.09 & 0.83 & 0.08 & 0.00 & 0.43 & 0.44 & 0.13 & 0.00 \\
\cline{1-14}
\bottomrule
\end{tabular}
\end{adjustbox}
\caption{\textbf{Row-wise confusion matrices for mean-difference probe with conformal prediction (\texttt{MD+CP}) across all $\langle$model, dataset$\rangle$ pairs} (evaluated on the \underline{\textit{last token representation}}).  
Each row corresponds to a specific model and a dataset. 
Columns are grouped by the ground-truth labels (\textit{True}, \textit{False}, \textit{Neither}) with groups of subcolumns that specify the distribution of predictions (\textit{true}, \textit{false}, \textit{neither}, \textit{abstain}).  
For each statement in a dataset, the predicted class is the class with the highest probability (as estimated by \texttt{MD+CP}).
The values in each group of four subcolumns sum to \(1\) because they are normalized counts.
For example, in the first row under the \textit{True} ground-truth label, we see that \textit{true} predictions have the value of \(0.89\); that is, \(89\%\) of all the true statements are classified as true.}
\label{sup_tab:cm_mdcp_instance}
\end{table}
\end{small}

\begin{small}
\centering
\begin{table}
\begin{adjustbox}{width=\textwidth}
\begin{tabular}{ll|cccc|cccc|cccc}
\toprule
 & True Labels \(\rightarrow\)
   & \multicolumn{4}{c|}{True } 
   & \multicolumn{4}{c|}{False} 
   & \multicolumn{4}{c}{Neither} \\
 & Predicted  \(\rightarrow\)
   & True & False & Neither & Abstain 
   & True & False & Neither & Abstain 
   & True & False & Neither & Abstain 
\\
Model  \(\downarrow\)& Dataset \(\downarrow\)
   &  &  &  &  
   &  &  &  &  
   &  &  &  &   \\
\midrule
\multirow[t]{3}{*}{Bio-Medical-Llama-3-8B} & City Locations & 1.00 & 0.00 & 0.00 & 0.00 & 1.00 & 0.00 & 0.00 & 0.00 & 1.00 & 0.00 & 0.00 & 0.00 \\
 & Medical Indications & 0.99 & 0.00 & 0.01 & 0.00 & 0.95 & 0.03 & 0.02 & 0.00 & 0.98 & 0.01 & 0.01 & 0.00 \\
 & Word Definitions & 1.00 & 0.00 & 0.00 & 0.00 & 1.00 & 0.00 & 0.00 & 0.00 & 1.00 & 0.00 & 0.00 & 0.00 \\
\cline{1-14}
\multirow[t]{3}{*}{Gemma-2-9B} & City Locations & 1.00 & 0.00 & 0.00 & 0.00 & 0.97 & 0.01 & 0.02 & 0.00 & 0.99 & 0.00 & 0.01 & 0.00 \\
 & Medical Indications & 0.95 & 0.01 & 0.04 & 0.00 & 0.88 & 0.04 & 0.09 & 0.00 & 0.98 & 0.01 & 0.01 & 0.00 \\
 & Word Definitions & 0.80 & 0.02 & 0.18 & 0.00 & 0.65 & 0.05 & 0.29 & 0.00 & 0.63 & 0.06 & 0.31 & 0.00 \\
\cline{1-14}
\multirow[t]{3}{*}{Gemma-2-9B-it} & City Locations & 0.87 & 0.01 & 0.12 & 0.00 & 0.30 & 0.15 & 0.55 & 0.00 & 0.42 & 0.07 & 0.51 & 0.00 \\
 & Medical Indications & 1.00 & 0.00 & 0.00 & 0.00 & 0.98 & 0.00 & 0.01 & 0.00 & 0.99 & 0.00 & 0.01 & 0.00 \\
 & Word Definitions & 0.57 & 0.02 & 0.41 & 0.00 & 0.32 & 0.13 & 0.55 & 0.00 & 0.20 & 0.12 & 0.68 & 0.00 \\
\cline{1-14}
\multirow[t]{3}{*}{Gemma-7B} & City Locations & 0.95 & 0.03 & 0.02 & 0.00 & 0.48 & 0.46 & 0.06 & 0.00 & 0.76 & 0.21 & 0.02 & 0.00 \\
 & Medical Indications & 1.00 & 0.00 & 0.00 & 0.00 & 1.00 & 0.00 & 0.00 & 0.00 & 1.00 & 0.00 & 0.00 & 0.00 \\
 & Word Definitions & 0.93 & 0.00 & 0.06 & 0.00 & 0.90 & 0.01 & 0.09 & 0.00 & 0.94 & 0.00 & 0.05 & 0.00 \\
\cline{1-14}
\multirow[t]{3}{*}{Gemma-7B-it} & City Locations & 0.83 & 0.06 & 0.11 & 0.00 & 0.59 & 0.31 & 0.11 & 0.00 & 0.81 & 0.12 & 0.07 & 0.00 \\
 & Medical Indications & 1.00 & 0.00 & 0.00 & 0.00 & 1.00 & 0.00 & 0.00 & 0.00 & 1.00 & 0.00 & 0.00 & 0.00 \\
 & Word Definitions & 0.93 & 0.00 & 0.07 & 0.00 & 0.88 & 0.01 & 0.11 & 0.00 & 0.96 & 0.00 & 0.04 & 0.00 \\
\cline{1-14}
\multirow[t]{3}{*}{Llama-3-8B} & City Locations & 1.00 & 0.00 & 0.00 & 0.00 & 1.00 & 0.00 & 0.00 & 0.00 & 1.00 & 0.00 & 0.00 & 0.00 \\
 & Medical Indications & 0.68 & 0.02 & 0.30 & 0.00 & 0.59 & 0.05 & 0.36 & 0.00 & 0.69 & 0.04 & 0.27 & 0.00 \\
 & Word Definitions & 0.96 & 0.00 & 0.03 & 0.00 & 0.94 & 0.01 & 0.05 & 0.00 & 0.91 & 0.01 & 0.08 & 0.00 \\
\cline{1-14}
\multirow[t]{3}{*}{Llama-3.1-8B-Instruct} & City Locations & 1.00 & 0.00 & 0.00 & 0.00 & 1.00 & 0.00 & 0.00 & 0.00 & 1.00 & 0.00 & 0.00 & 0.00 \\
 & Medical Indications & 1.00 & 0.00 & 0.00 & 0.00 & 1.00 & 0.00 & 0.00 & 0.00 & 1.00 & 0.00 & 0.00 & 0.00 \\
 & Word Definitions & 1.00 & 0.00 & 0.00 & 0.00 & 1.00 & 0.00 & 0.00 & 0.00 & 1.00 & 0.00 & 0.00 & 0.00 \\
\cline{1-14}
\multirow[t]{3}{*}{Llama-3.2-3B} & City Locations & 1.00 & 0.00 & 0.00 & 0.00 & 1.00 & 0.00 & 0.00 & 0.00 & 1.00 & 0.00 & 0.00 & 0.00 \\
 & Medical Indications & 0.83 & 0.03 & 0.14 & 0.00 & 0.56 & 0.22 & 0.23 & 0.00 & 0.95 & 0.02 & 0.03 & 0.00 \\
 & Word Definitions & 0.92 & 0.01 & 0.07 & 0.00 & 0.75 & 0.09 & 0.16 & 0.00 & 0.92 & 0.02 & 0.07 & 0.00 \\
\cline{1-14}
\multirow[t]{3}{*}{Llama-3.2-3B-Instruct} & City Locations & 1.00 & 0.00 & 0.00 & 0.00 & 1.00 & 0.00 & 0.00 & 0.00 & 1.00 & 0.00 & 0.00 & 0.00 \\
 & Medical Indications & 1.00 & 0.00 & 0.00 & 0.00 & 1.00 & 0.00 & 0.00 & 0.00 & 1.00 & 0.00 & 0.00 & 0.00 \\
 & Word Definitions & 1.00 & 0.00 & 0.00 & 0.00 & 1.00 & 0.00 & 0.00 & 0.00 & 1.00 & 0.00 & 0.00 & 0.00 \\
\cline{1-14}
\multirow[t]{3}{*}{Llama3-Med42-8B} & City Locations & 1.00 & 0.00 & 0.00 & 0.00 & 1.00 & 0.00 & 0.00 & 0.00 & 1.00 & 0.00 & 0.00 & 0.00 \\
 & Medical Indications & 0.93 & 0.00 & 0.07 & 0.00 & 0.93 & 0.00 & 0.06 & 0.00 & 0.94 & 0.00 & 0.06 & 0.00 \\
 & Word Definitions & 1.00 & 0.00 & 0.00 & 0.00 & 1.00 & 0.00 & 0.00 & 0.00 & 1.00 & 0.00 & 0.00 & 0.00 \\
\cline{1-14}
\multirow[t]{3}{*}{Mistral-7B-Instruct-v0.3} & City Locations & 0.97 & 0.00 & 0.03 & 0.00 & 0.88 & 0.02 & 0.10 & 0.00 & 0.97 & 0.00 & 0.03 & 0.00 \\
 & Medical Indications & 0.96 & 0.01 & 0.03 & 0.00 & 0.91 & 0.04 & 0.05 & 0.00 & 0.90 & 0.04 & 0.06 & 0.00 \\
 & Word Definitions & 1.00 & 0.00 & 0.00 & 0.00 & 0.97 & 0.01 & 0.02 & 0.00 & 1.00 & 0.00 & 0.00 & 0.00 \\
\cline{1-14}
\multirow[t]{3}{*}{Mistral-7B-v0.3} & City Locations & 0.97 & 0.01 & 0.02 & 0.00 & 0.86 & 0.08 & 0.06 & 0.00 & 0.90 & 0.05 & 0.06 & 0.00 \\
 & Medical Indications & 0.97 & 0.00 & 0.03 & 0.00 & 0.96 & 0.00 & 0.04 & 0.00 & 0.97 & 0.00 & 0.03 & 0.00 \\
 & Word Definitions & 0.94 & 0.00 & 0.06 & 0.00 & 0.86 & 0.02 & 0.12 & 0.00 & 0.92 & 0.02 & 0.06 & 0.00 \\
\cline{1-14}
\multirow[t]{3}{*}{Qwen-2.5-14B} & City Locations & 1.00 & 0.00 & 0.00 & 0.00 & 0.89 & 0.03 & 0.07 & 0.00 & 0.99 & 0.00 & 0.01 & 0.00 \\
 & Medical Indications & 0.91 & 0.01 & 0.07 & 0.00 & 0.88 & 0.04 & 0.08 & 0.00 & 0.96 & 0.01 & 0.03 & 0.00 \\
 & Word Definitions & 0.98 & 0.00 & 0.01 & 0.00 & 0.93 & 0.03 & 0.04 & 0.00 & 0.97 & 0.01 & 0.02 & 0.00 \\
\cline{1-14}
\multirow[t]{3}{*}{Qwen-2.5-14B-Instruct} & City Locations & 1.00 & 0.00 & 0.00 & 0.00 & 1.00 & 0.00 & 0.00 & 0.00 & 1.00 & 0.00 & 0.00 & 0.00 \\
 & Medical Indications & 0.90 & 0.01 & 0.09 & 0.00 & 0.66 & 0.10 & 0.24 & 0.00 & 0.60 & 0.04 & 0.36 & 0.00 \\
 & Word Definitions & 1.00 & 0.00 & 0.00 & 0.00 & 1.00 & 0.00 & 0.00 & 0.00 & 1.00 & 0.00 & 0.00 & 0.00 \\
\cline{1-14}
\multirow[t]{3}{*}{Qwen-2.5-7B} & City Locations & 0.99 & 0.00 & 0.01 & 0.00 & 0.95 & 0.03 & 0.02 & 0.00 & 1.00 & 0.00 & 0.00 & 0.00 \\
 & Medical Indications & 0.93 & 0.01 & 0.06 & 0.00 & 0.76 & 0.08 & 0.16 & 0.00 & 0.87 & 0.05 & 0.09 & 0.00 \\
 & Word Definitions & 0.92 & 0.01 & 0.07 & 0.00 & 0.90 & 0.01 & 0.09 & 0.00 & 0.90 & 0.01 & 0.08 & 0.00 \\
\cline{1-14}
\multirow[t]{3}{*}{Qwen-2.5-7B-Instruct} & City Locations & 1.00 & 0.00 & 0.00 & 0.00 & 1.00 & 0.00 & 0.00 & 0.00 & 1.00 & 0.00 & 0.00 & 0.00 \\
 & Medical Indications & 1.00 & 0.00 & 0.00 & 0.00 & 1.00 & 0.00 & 0.00 & 0.00 & 1.00 & 0.00 & 0.00 & 0.00 \\
 & Word Definitions & 1.00 & 0.00 & 0.00 & 0.00 & 1.00 & 0.00 & 0.00 & 0.00 & 1.00 & 0.00 & 0.00 & 0.00 \\
\cline{1-14}
\bottomrule
\end{tabular}
\end{adjustbox}
\caption{\textbf{Row-wise confusion matrices for mean-difference probe with conformal prediction (\texttt{MD+CP}) across all $\langle$model, dataset$\rangle$ pairs} (evaluated on the \underline{\textit{bag}}).  
Each row corresponds to a specific model and a dataset. 
Columns are grouped by the ground-truth labels (\textit{True}, \textit{False}, \textit{Neither}) with groups of subcolumns that specify the distribution of predictions (\textit{true}, \textit{false}, \textit{neither}, \textit{abstain}).  
For each statement in a dataset, the predicted class is the class with the highest probability (as estimated by \texttt{MD+CP}).
The values in each group of four subcolumns sum to \(1\) because they are normalized counts.
For example, in the first row under the \textit{True} ground-truth label, we see that \textit{true} predictions have the value of \(1.00\); that is, \(100\%\) of all the true statements are classified as true.}
\label{sup_tab:cm_mdcp_bag}
\end{table}
\end{small}

\begin{small}
\begin{table}
\centering
\begin{adjustbox}{width=\textwidth}
\begin{tabular}{ll|cccc|cccc|cccc}
\toprule
 & True Labels \(\rightarrow\)
   & \multicolumn{4}{c|}{True } 
   & \multicolumn{4}{c|}{False} 
   & \multicolumn{4}{c}{Neither} \\
 & Predicted  \(\rightarrow\)
   & True & False & Neither & Abstain 
   & True & False & Neither & Abstain 
   & True & False & Neither & Abstain 
\\
Model  \(\downarrow\)& Dataset \(\downarrow\)
   &  &  &  &  
   &  &  &  &  
   &  &  &  &   \\
\midrule
\multirow[c]{3}{*}{Bio-Medical-Llama-3-8B} & City Locations & 0.87 & 0.08 & 0.05 & 0.00 & 0.04 & 0.90 & 0.05 & 0.00 & 0.09 & 0.78 & 0.14 & 0.00 \\
 & Medical Indications & 0.59 & 0.08 & 0.33 & 0.00 & 0.09 & 0.64 & 0.27 & 0.00 & 0.03 & 0.61 & 0.36 & 0.00 \\
 & Word Definitions & 0.44 & 0.11 & 0.46 & 0.00 & 0.08 & 0.50 & 0.41 & 0.00 & 0.00 & 0.69 & 0.31 & 0.00 \\
\cline{1-14}
\multirow[c]{3}{*}{Gemma-2-9B} & City Locations & 0.91 & 0.09 & 0.00 & 0.00 & 0.14 & 0.85 & 0.00 & 0.00 & 0.00 & 1.00 & 0.00 & 0.00 \\
 & Medical Indications & 0.71 & 0.09 & 0.20 & 0.00 & 0.11 & 0.70 & 0.19 & 0.00 & 0.11 & 0.73 & 0.16 & 0.00 \\
 & Word Definitions & 0.68 & 0.08 & 0.23 & 0.00 & 0.08 & 0.74 & 0.18 & 0.00 & 0.08 & 0.55 & 0.36 & 0.00 \\
\cline{1-14}
\multirow[c]{3}{*}{Gemma-2-9B-it} & City Locations & 0.86 & 0.12 & 0.02 & 0.00 & 0.04 & 0.94 & 0.01 & 0.00 & 0.98 & 0.02 & 0.01 & 0.00 \\
 & Medical Indications & 0.70 & 0.07 & 0.23 & 0.00 & 0.11 & 0.70 & 0.19 & 0.00 & 0.14 & 0.34 & 0.51 & 0.00 \\
 & Word Definitions & 0.53 & 0.11 & 0.36 & 0.00 & 0.07 & 0.59 & 0.34 & 0.00 & 0.03 & 0.58 & 0.39 & 0.00 \\
\cline{1-14}
\multirow[c]{3}{*}{Gemma-7B} & City Locations & 0.90 & 0.05 & 0.06 & 0.00 & 0.08 & 0.88 & 0.03 & 0.00 & 0.03 & 0.95 & 0.02 & 0.00 \\
 & Medical Indications & 0.68 & 0.08 & 0.25 & 0.00 & 0.09 & 0.68 & 0.23 & 0.00 & 0.10 & 0.52 & 0.38 & 0.00 \\
 & Word Definitions & 0.75 & 0.09 & 0.16 & 0.00 & 0.09 & 0.79 & 0.12 & 0.00 & 0.13 & 0.66 & 0.21 & 0.00 \\
\cline{1-14}
\multirow[c]{3}{*}{Gemma-7B-it} & City Locations & 0.88 & 0.04 & 0.08 & 0.00 & 0.05 & 0.91 & 0.03 & 0.00 & 0.90 & 0.01 & 0.09 & 0.00 \\
 & Medical Indications & 0.54 & 0.07 & 0.39 & 0.00 & 0.13 & 0.52 & 0.35 & 0.00 & 0.45 & 0.13 & 0.41 & 0.00 \\
 & Word Definitions & 0.60 & 0.12 & 0.28 & 0.00 & 0.09 & 0.64 & 0.26 & 0.00 & 0.17 & 0.37 & 0.45 & 0.00 \\
\cline{1-14}
\multirow[c]{3}{*}{Llama-3-8B} & City Locations & 0.88 & 0.10 & 0.02 & 0.00 & 0.07 & 0.89 & 0.04 & 0.00 & 0.02 & 0.98 & 0.01 & 0.00 \\
 & Medical Indications & 0.65 & 0.10 & 0.25 & 0.00 & 0.11 & 0.70 & 0.19 & 0.00 & 0.02 & 0.62 & 0.36 & 0.00 \\
 & Word Definitions & 0.72 & 0.11 & 0.17 & 0.00 & 0.09 & 0.79 & 0.13 & 0.00 & 0.41 & 0.42 & 0.17 & 0.00 \\
\cline{1-14}
\multirow[c]{3}{*}{Llama-3.1-8B-Instruct} & City Locations & 0.84 & 0.03 & 0.13 & 0.00 & 0.03 & 0.92 & 0.05 & 0.00 & 0.00 & 0.84 & 0.16 & 0.00 \\
 & Medical Indications & 0.68 & 0.06 & 0.26 & 0.00 & 0.10 & 0.65 & 0.25 & 0.00 & 0.00 & 0.74 & 0.26 & 0.00 \\
 & Word Definitions & 0.80 & 0.08 & 0.11 & 0.00 & 0.09 & 0.81 & 0.11 & 0.00 & 0.15 & 0.55 & 0.30 & 0.00 \\
\cline{1-14}
\multirow[c]{3}{*}{Llama-3.2-3B} & City Locations & 0.88 & 0.10 & 0.02 & 0.00 & 0.07 & 0.90 & 0.02 & 0.00 & 0.02 & 0.98 & 0.01 & 0.00 \\
 & Medical Indications & 0.71 & 0.08 & 0.21 & 0.00 & 0.10 & 0.69 & 0.21 & 0.00 & 0.01 & 0.86 & 0.12 & 0.00 \\
 & Word Definitions & 0.64 & 0.09 & 0.27 & 0.00 & 0.08 & 0.66 & 0.25 & 0.00 & 0.10 & 0.55 & 0.34 & 0.00 \\
\cline{1-14}
\multirow[c]{3}{*}{Llama-3.2-3B-Instruct} & City Locations & 0.85 & 0.11 & 0.04 & 0.00 & 0.10 & 0.87 & 0.02 & 0.00 & 0.73 & 0.23 & 0.04 & 0.00 \\
 & Medical Indications & 0.51 & 0.13 & 0.36 & 0.00 & 0.08 & 0.60 & 0.32 & 0.00 & 0.06 & 0.41 & 0.53 & 0.00 \\
 & Word Definitions & 0.44 & 0.12 & 0.43 & 0.00 & 0.07 & 0.49 & 0.44 & 0.00 & 0.13 & 0.48 & 0.40 & 0.00 \\
\cline{1-14}
\multirow[c]{3}{*}{Llama3-Med42-8B} & City Locations & 0.89 & 0.04 & 0.07 & 0.00 & 0.05 & 0.92 & 0.02 & 0.00 & 0.22 & 0.66 & 0.12 & 0.00 \\
 & Medical Indications & 0.69 & 0.09 & 0.22 & 0.00 & 0.10 & 0.74 & 0.16 & 0.00 & 0.06 & 0.07 & 0.87 & 0.00 \\
 & Word Definitions & 0.71 & 0.10 & 0.19 & 0.00 & 0.07 & 0.78 & 0.15 & 0.00 & 0.02 & 0.82 & 0.17 & 0.00 \\
\cline{1-14}
\multirow[c]{3}{*}{Mistral-7B-Instruct-v0.3} & City Locations & 0.88 & 0.11 & 0.00 & 0.00 & 0.07 & 0.93 & 0.00 & 0.00 & 0.30 & 0.70 & 0.00 & 0.00 \\
 & Medical Indications & 0.73 & 0.08 & 0.19 & 0.00 & 0.10 & 0.72 & 0.19 & 0.00 & 0.06 & 0.54 & 0.40 & 0.00 \\
 & Word Definitions & 0.71 & 0.10 & 0.19 & 0.00 & 0.09 & 0.72 & 0.19 & 0.00 & 0.16 & 0.57 & 0.27 & 0.00 \\
\cline{1-14}
\multirow[c]{3}{*}{Mistral-7B-v0.3} & City Locations & 0.88 & 0.08 & 0.04 & 0.00 & 0.06 & 0.92 & 0.03 & 0.00 & 0.31 & 0.59 & 0.10 & 0.00 \\
 & Medical Indications & 0.72 & 0.07 & 0.21 & 0.00 & 0.10 & 0.73 & 0.17 & 0.00 & 0.15 & 0.36 & 0.49 & 0.00 \\
 & Word Definitions & 0.72 & 0.10 & 0.17 & 0.00 & 0.11 & 0.75 & 0.15 & 0.00 & 0.20 & 0.50 & 0.31 & 0.00 \\
\cline{1-14}
\multirow[c]{3}{*}{Qwen-2.5-14B} & City Locations & 0.91 & 0.07 & 0.01 & 0.00 & 0.08 & 0.91 & 0.01 & 0.00 & 0.10 & 0.88 & 0.02 & 0.00 \\
 & Medical Indications & 0.69 & 0.06 & 0.25 & 0.00 & 0.15 & 0.65 & 0.20 & 0.00 & 0.36 & 0.14 & 0.50 & 0.00 \\
 & Word Definitions & 0.60 & 0.12 & 0.28 & 0.00 & 0.09 & 0.61 & 0.30 & 0.00 & 0.13 & 0.49 & 0.37 & 0.00 \\
\cline{1-14}
\multirow[c]{3}{*}{Qwen-2.5-14B-Instruct} & City Locations & 0.88 & 0.10 & 0.03 & 0.00 & 0.07 & 0.91 & 0.02 & 0.00 & 0.39 & 0.54 & 0.07 & 0.00 \\
 & Medical Indications & 0.80 & 0.08 & 0.13 & 0.00 & 0.13 & 0.75 & 0.11 & 0.00 & 0.03 & 0.55 & 0.42 & 0.00 \\
 & Word Definitions & 0.72 & 0.10 & 0.19 & 0.00 & 0.11 & 0.75 & 0.15 & 0.00 & 0.04 & 0.71 & 0.25 & 0.00 \\
\cline{1-14}
\multirow[c]{3}{*}{Qwen-2.5-7B} & City Locations & 0.92 & 0.03 & 0.05 & 0.00 & 0.06 & 0.89 & 0.05 & 0.00 & 0.01 & 0.90 & 0.09 & 0.00 \\
 & Medical Indications & 0.67 & 0.07 & 0.25 & 0.00 & 0.10 & 0.70 & 0.20 & 0.00 & 0.33 & 0.19 & 0.48 & 0.00 \\
 & Word Definitions & 0.52 & 0.11 & 0.37 & 0.00 & 0.07 & 0.56 & 0.37 & 0.00 & 0.11 & 0.52 & 0.37 & 0.00 \\
\cline{1-14}
\multirow[c]{3}{*}{Qwen-2.5-7B-Instruct} & City Locations & 0.90 & 0.03 & 0.07 & 0.00 & 0.07 & 0.87 & 0.06 & 0.00 & 0.33 & 0.10 & 0.57 & 0.00 \\
 & Medical Indications & 0.73 & 0.09 & 0.17 & 0.00 & 0.15 & 0.72 & 0.13 & 0.00 & 0.02 & 0.88 & 0.10 & 0.00 \\
 & Word Definitions & 0.71 & 0.11 & 0.18 & 0.00 & 0.10 & 0.71 & 0.19 & 0.00 & 0.11 & 0.60 & 0.30 & 0.00 \\
\cline{1-14}
\bottomrule
\end{tabular}
\end{adjustbox}
\caption{\textbf{Row-wise confusion matrices for \texttt{TTPD} probe with conformal prediction (\texttt{TTPD+CP}) across all $\langle$model, dataset$\rangle$ pairs} (evaluated on the \underline{\textit{last token representation}}).  
Each row corresponds to a specific model and a dataset. 
Columns are grouped by the ground-truth labels (\textit{True}, \textit{False}, \textit{Neither}) with groups of subcolumns that specify the distribution of predictions (\textit{true}, \textit{false}, \textit{neither}, \textit{abstain}).  
For each statement in a dataset, the predicted class is the class with the highest probability (as estimated by \texttt{TTPD+CP}).
The values in each group of four subcolumns sum to \(1\) because they are normalized counts.
For example, in the first row under the \textit{True} ground-truth label, we see that \textit{true} predictions have the value of \(0.87\); that is, \(87\%\) of all the true statements are classified as true.}
\label{sup_tab:cm_ttpd_instance}
\end{table}
\end{small}

\begin{small}
\begin{table}
\centering
\begin{adjustbox}{width=\textwidth}
\begin{tabular}{ll|cccc|cccc|cccc}
\toprule
 & True Labels \(\rightarrow\)
   & \multicolumn{4}{c|}{True } 
   & \multicolumn{4}{c|}{False} 
   & \multicolumn{4}{c}{Neither} \\
 & Predicted  \(\rightarrow\)
   & True & False & Neither & Abstain 
   & True & False & Neither & Abstain 
   & True & False & Neither & Abstain 
\\
Model  \(\downarrow\)& Dataset \(\downarrow\)
   &  &  &  &  
   &  &  &  &  
   &  &  &  &   \\
\midrule
\multirow[c]{3}{*}{Bio-Medical-Llama-3-8B} & City Locations & 0.19 & 0.00 & 0.81 & 0.00 & 0.16 & 0.01 & 0.83 & 0.00 & 0.03 & 0.24 & 0.72 & 0.00 \\
 & Medical Indications & 0.08 & 0.08 & 0.84 & 0.00 & 0.09 & 0.05 & 0.86 & 0.00 & 0.08 & 0.02 & 0.91 & 0.00 \\
 & Word Definitions & 1.00 & 0.00 & 0.00 & 0.00 & 1.00 & 0.00 & 0.00 & 0.00 & 1.00 & 0.00 & 0.00 & 0.00 \\
\cline{1-14}
\multirow[c]{3}{*}{Gemma-2-9B} & City Locations & 0.92 & 0.08 & 0.00 & 0.00 & 0.14 & 0.85 & 0.00 & 0.00 & 0.00 & 1.00 & 0.00 & 0.00 \\
 & Medical Indications & 0.12 & 0.03 & 0.85 & 0.00 & 0.08 & 0.03 & 0.89 & 0.00 & 0.00 & 0.02 & 0.98 & 0.00 \\
 & Word Definitions & 0.30 & 0.08 & 0.62 & 0.00 & 0.11 & 0.16 & 0.72 & 0.00 & 0.00 & 0.79 & 0.21 & 0.00 \\
\cline{1-14}
\multirow[c]{3}{*}{Gemma-2-9B-it} & City Locations & 1.00 & 0.00 & 0.00 & 0.00 & 1.00 & 0.00 & 0.00 & 0.00 & 1.00 & 0.00 & 0.00 & 0.00 \\
 & Medical Indications & 1.00 & 0.00 & 0.00 & 0.00 & 1.00 & 0.00 & 0.00 & 0.00 & 1.00 & 0.00 & 0.00 & 0.00 \\
 & Word Definitions & 0.48 & 0.01 & 0.51 & 0.00 & 0.11 & 0.20 & 0.69 & 0.00 & 0.01 & 0.40 & 0.59 & 0.00 \\
\cline{1-14}
\multirow[c]{3}{*}{Gemma-7B} & City Locations & 0.93 & 0.04 & 0.04 & 0.00 & 0.05 & 0.88 & 0.07 & 0.00 & 0.01 & 0.98 & 0.01 & 0.00 \\
 & Medical Indications & 0.10 & 0.08 & 0.82 & 0.00 & 0.07 & 0.13 & 0.80 & 0.00 & 0.00 & 0.26 & 0.74 & 0.00 \\
 & Word Definitions & 0.38 & 0.10 & 0.53 & 0.00 & 0.09 & 0.27 & 0.64 & 0.00 & 0.01 & 0.59 & 0.40 & 0.00 \\
\cline{1-14}
\multirow[c]{3}{*}{Gemma-7B-it} & City Locations & 0.23 & 0.05 & 0.72 & 0.00 & 0.19 & 0.13 & 0.67 & 0.00 & 0.01 & 0.51 & 0.48 & 0.00 \\
 & Medical Indications & 1.00 & 0.00 & 0.00 & 0.00 & 1.00 & 0.00 & 0.00 & 0.00 & 1.00 & 0.00 & 0.00 & 0.00 \\
 & Word Definitions & 1.00 & 0.00 & 0.00 & 0.00 & 1.00 & 0.00 & 0.00 & 0.00 & 1.00 & 0.00 & 0.00 & 0.00 \\
\cline{1-14}
\multirow[c]{3}{*}{Llama-3-8B} & City Locations & 0.88 & 0.11 & 0.00 & 0.00 & 0.11 & 0.88 & 0.01 & 0.00 & 0.02 & 0.98 & 0.00 & 0.00 \\
 & Medical Indications & 0.54 & 0.08 & 0.37 & 0.00 & 0.10 & 0.54 & 0.36 & 0.00 & 0.01 & 0.66 & 0.32 & 0.00 \\
 & Word Definitions & 1.00 & 0.00 & 0.00 & 0.00 & 1.00 & 0.00 & 0.00 & 0.00 & 1.00 & 0.00 & 0.00 & 0.00 \\
\cline{1-14}
\multirow[c]{3}{*}{Llama-3.1-8B-Instruct} & City Locations & 1.00 & 0.00 & 0.00 & 0.00 & 1.00 & 0.00 & 0.00 & 0.00 & 1.00 & 0.00 & 0.00 & 0.00 \\
 & Medical Indications & 0.18 & 0.10 & 0.73 & 0.00 & 0.08 & 0.14 & 0.78 & 0.00 & 0.20 & 0.32 & 0.48 & 0.00 \\
 & Word Definitions & 1.00 & 0.00 & 0.00 & 0.00 & 1.00 & 0.00 & 0.00 & 0.00 & 1.00 & 0.00 & 0.00 & 0.00 \\
\cline{1-14}
\multirow[c]{3}{*}{Llama-3.2-3B} & City Locations & 0.91 & 0.07 & 0.02 & 0.00 & 0.07 & 0.90 & 0.03 & 0.00 & 0.02 & 0.98 & 0.01 & 0.00 \\
 & Medical Indications & 0.64 & 0.05 & 0.31 & 0.00 & 0.10 & 0.18 & 0.72 & 0.00 & 0.01 & 0.56 & 0.43 & 0.00 \\
 & Word Definitions & 0.38 & 0.03 & 0.59 & 0.00 & 0.09 & 0.08 & 0.82 & 0.00 & 0.01 & 0.16 & 0.83 & 0.00 \\
\cline{1-14}
\multirow[c]{3}{*}{Llama-3.2-3B-Instruct} & City Locations & 1.00 & 0.00 & 0.00 & 0.00 & 1.00 & 0.00 & 0.00 & 0.00 & 1.00 & 0.00 & 0.00 & 0.00 \\
 & Medical Indications & 1.00 & 0.00 & 0.00 & 0.00 & 1.00 & 0.00 & 0.00 & 0.00 & 1.00 & 0.00 & 0.00 & 0.00 \\
 & Word Definitions & 1.00 & 0.00 & 0.00 & 0.00 & 1.00 & 0.00 & 0.00 & 0.00 & 1.00 & 0.00 & 0.00 & 0.00 \\
\cline{1-14}
\multirow[c]{3}{*}{Llama3-Med42-8B} & City Locations & 0.82 & 0.01 & 0.17 & 0.00 & 0.41 & 0.07 & 0.52 & 0.00 & 0.51 & 0.47 & 0.02 & 0.00 \\
 & Medical Indications & 1.00 & 0.00 & 0.00 & 0.00 & 1.00 & 0.00 & 0.00 & 0.00 & 1.00 & 0.00 & 0.00 & 0.00 \\
 & Word Definitions & 1.00 & 0.00 & 0.00 & 0.00 & 1.00 & 0.00 & 0.00 & 0.00 & 1.00 & 0.00 & 0.00 & 0.00 \\
\cline{1-14}
\multirow[c]{3}{*}{Mistral-7B-Instruct-v0.3} & City Locations & 1.00 & 0.00 & 0.00 & 0.00 & 1.00 & 0.00 & 0.00 & 0.00 & 1.00 & 0.00 & 0.00 & 0.00 \\
 & Medical Indications & 0.35 & 0.05 & 0.60 & 0.00 & 0.24 & 0.11 & 0.65 & 0.00 & 0.23 & 0.32 & 0.45 & 0.00 \\
 & Word Definitions & 1.00 & 0.00 & 0.00 & 0.00 & 1.00 & 0.00 & 0.00 & 0.00 & 1.00 & 0.00 & 0.00 & 0.00 \\
\cline{1-14}
\multirow[c]{3}{*}{Mistral-7B-v0.3} & City Locations & 0.88 & 0.06 & 0.06 & 0.00 & 0.24 & 0.56 & 0.20 & 0.00 & 0.34 & 0.50 & 0.16 & 0.00 \\
 & Medical Indications & 0.11 & 0.14 & 0.75 & 0.00 & 0.06 & 0.13 & 0.81 & 0.00 & 0.00 & 0.39 & 0.61 & 0.00 \\
 & Word Definitions & 0.76 & 0.00 & 0.24 & 0.00 & 0.31 & 0.01 & 0.68 & 0.00 & 0.34 & 0.08 & 0.58 & 0.00 \\
\cline{1-14}
\multirow[c]{3}{*}{Qwen-2.5-14B} & City Locations & 0.90 & 0.04 & 0.06 & 0.00 & 0.05 & 0.87 & 0.08 & 0.00 & 0.05 & 0.88 & 0.08 & 0.00 \\
 & Medical Indications & 0.11 & 0.02 & 0.87 & 0.00 & 0.09 & 0.02 & 0.89 & 0.00 & 0.01 & 0.07 & 0.93 & 0.00 \\
 & Word Definitions & 0.52 & 0.00 & 0.47 & 0.00 & 0.29 & 0.01 & 0.70 & 0.00 & 0.29 & 0.12 & 0.59 & 0.00 \\
\cline{1-14}
\multirow[c]{3}{*}{Qwen-2.5-14B-Instruct} & City Locations & 0.87 & 0.00 & 0.13 & 0.00 & 0.02 & 0.17 & 0.81 & 0.00 & 0.26 & 0.01 & 0.73 & 0.00 \\
 & Medical Indications & 1.00 & 0.00 & 0.00 & 0.00 & 1.00 & 0.00 & 0.00 & 0.00 & 1.00 & 0.00 & 0.00 & 0.00 \\
 & Word Definitions & 1.00 & 0.00 & 0.00 & 0.00 & 1.00 & 0.00 & 0.00 & 0.00 & 1.00 & 0.00 & 0.00 & 0.00 \\
\cline{1-14}
\multirow[c]{3}{*}{Qwen-2.5-7B} & City Locations & 0.91 & 0.06 & 0.03 & 0.00 & 0.11 & 0.85 & 0.05 & 0.00 & 0.28 & 0.66 & 0.06 & 0.00 \\
 & Medical Indications & 0.10 & 0.09 & 0.81 & 0.00 & 0.07 & 0.08 & 0.85 & 0.00 & 0.00 & 0.13 & 0.87 & 0.00 \\
 & Word Definitions & 0.39 & 0.01 & 0.60 & 0.00 & 0.11 & 0.03 & 0.86 & 0.00 & 0.09 & 0.11 & 0.81 & 0.00 \\
\cline{1-14}
\multirow[c]{3}{*}{Qwen-2.5-7B-Instruct} & City Locations & 1.00 & 0.00 & 0.00 & 0.00 & 1.00 & 0.00 & 0.00 & 0.00 & 1.00 & 0.00 & 0.00 & 0.00 \\
 & Medical Indications & 1.00 & 0.00 & 0.00 & 0.00 & 1.00 & 0.00 & 0.00 & 0.00 & 1.00 & 0.00 & 0.00 & 0.00 \\
 & Word Definitions & 1.00 & 0.00 & 0.00 & 0.00 & 1.00 & 0.00 & 0.00 & 0.00 & 1.00 & 0.00 & 0.00 & 0.00 \\
\cline{1-14}
\bottomrule
\end{tabular}
\end{adjustbox}
\caption{\textbf{Row-wise confusion matrices for \texttt{TTPD} probe with conformal prediction (\texttt{TTPD+CP}) across all $\langle$model-dataset pairs$\rangle$} (evaluated on the \underline{\textit{bag}}).  
Each row corresponds to a specific model and a dataset. 
Columns are grouped by the ground-truth labels (\textit{True}, \textit{False}, \textit{Neither}) with groups of subcolumns that specify the distribution of predictions (\textit{true}, \textit{false}, \textit{neither}, \textit{abstain}).  
For each statement in a dataset, the predicted class is the class with the highest probability (as estimated by \texttt{TTPD+CP}).
The values in each group of four subcolumns sum to \(1\) because they are normalized counts.
For example, in the first row under the \textit{True} ground-truth label, we see that \textit{true} predictions have the value of \(0.19\); that is, \(19\%\) of all the true statements are classified as true.}
\label{sup_tab:cm_ttpd_bag}
\end{table}
\end{small}

\begin{small}
\begin{table}
\centering
\begin{adjustbox}{width=\textwidth}
\begin{tabular}{ll|cccc|cccc|cccc}
\toprule
 & True Labels \(\rightarrow\)
   & \multicolumn{4}{c|}{True } 
   & \multicolumn{4}{c|}{False} 
   & \multicolumn{4}{c}{Neither} \\
 & Predicted  \(\rightarrow\)
   & True & False & Neither & Abstain 
   & True & False & Neither & Abstain 
   & True & False & Neither & Abstain 
\\
Model  \(\downarrow\)& Dataset \(\downarrow\)
   &  &  &  &  
   &  &  &  &  
   &  &  &  &   \\
\midrule
\multirow[c]{3}{*}{Bio-Medical-Llama-3-8B} & City Locations & 0.90 & 0.02 & 0.09 & 0.00 & 0.01 & 0.92 & 0.07 & 0.00 & 0.04 & 0.75 & 0.21 & 0.00 \\
 & Medical Indications & 0.74 & 0.10 & 0.16 & 0.00 & 0.09 & 0.79 & 0.13 & 0.00 & 0.06 & 0.79 & 0.15 & 0.00 \\
 & Word Definitions & 0.80 & 0.11 & 0.09 & 0.00 & 0.10 & 0.83 & 0.07 & 0.00 & 0.39 & 0.54 & 0.07 & 0.00 \\
\cline{1-14}
\multirow[c]{3}{*}{Gemma-2-9B} & City Locations & 0.89 & 0.03 & 0.08 & 0.00 & 0.01 & 0.93 & 0.06 & 0.00 & 0.00 & 0.98 & 0.02 & 0.00 \\
 & Medical Indications & 0.83 & 0.11 & 0.06 & 0.00 & 0.09 & 0.84 & 0.07 & 0.00 & 0.03 & 0.86 & 0.11 & 0.00 \\
 & Word Definitions & 0.87 & 0.10 & 0.03 & 0.00 & 0.08 & 0.89 & 0.04 & 0.00 & 0.42 & 0.50 & 0.08 & 0.00 \\
\cline{1-14}
\multirow[c]{3}{*}{Gemma-2-9B-it} & City Locations & 0.86 & 0.10 & 0.03 & 0.00 & 0.05 & 0.93 & 0.02 & 0.00 & 0.47 & 0.49 & 0.04 & 0.00 \\
 & Medical Indications & 0.79 & 0.10 & 0.10 & 0.00 & 0.07 & 0.81 & 0.12 & 0.00 & 0.41 & 0.28 & 0.31 & 0.00 \\
 & Word Definitions & 0.79 & 0.13 & 0.08 & 0.00 & 0.09 & 0.86 & 0.05 & 0.00 & 0.38 & 0.52 & 0.10 & 0.00 \\
\cline{1-14}
\multirow[c]{3}{*}{Gemma-7B} & City Locations & 0.89 & 0.00 & 0.11 & 0.00 & 0.00 & 0.91 & 0.09 & 0.00 & 0.14 & 0.29 & 0.57 & 0.00 \\
 & Medical Indications & 0.78 & 0.09 & 0.13 & 0.00 & 0.08 & 0.81 & 0.11 & 0.00 & 0.23 & 0.49 & 0.27 & 0.00 \\
 & Word Definitions & 0.84 & 0.08 & 0.08 & 0.00 & 0.09 & 0.83 & 0.07 & 0.00 & 0.42 & 0.42 & 0.16 & 0.00 \\
\cline{1-14}
\multirow[c]{3}{*}{Gemma-7B-it} & City Locations & 0.86 & 0.04 & 0.10 & 0.00 & 0.02 & 0.92 & 0.07 & 0.00 & 0.10 & 0.60 & 0.30 & 0.00 \\
 & Medical Indications & 0.62 & 0.09 & 0.28 & 0.00 & 0.10 & 0.66 & 0.24 & 0.00 & 0.46 & 0.46 & 0.07 & 0.00 \\
 & Word Definitions & 0.80 & 0.08 & 0.12 & 0.00 & 0.07 & 0.84 & 0.10 & 0.00 & 0.33 & 0.49 & 0.18 & 0.00 \\
\cline{1-14}
\multirow[c]{3}{*}{Llama-3-8B} & City Locations & 0.88 & 0.02 & 0.10 & 0.00 & 0.01 & 0.91 & 0.08 & 0.00 & 0.70 & 0.12 & 0.19 & 0.00 \\
 & Medical Indications & 0.84 & 0.12 & 0.05 & 0.00 & 0.11 & 0.85 & 0.04 & 0.00 & 0.35 & 0.56 & 0.09 & 0.00 \\
 & Word Definitions & 0.85 & 0.09 & 0.07 & 0.00 & 0.08 & 0.86 & 0.07 & 0.00 & 0.49 & 0.43 & 0.08 & 0.00 \\
\cline{1-14}
\multirow[c]{3}{*}{Llama-3.1-8B-Instruct} & City Locations & 0.88 & 0.12 & 0.01 & 0.00 & 0.07 & 0.93 & 0.00 & 0.00 & 0.52 & 0.48 & 0.00 & 0.00 \\
 & Medical Indications & 0.83 & 0.11 & 0.06 & 0.00 & 0.08 & 0.84 & 0.08 & 0.00 & 0.25 & 0.59 & 0.15 & 0.00 \\
 & Word Definitions & 0.89 & 0.07 & 0.03 & 0.00 & 0.07 & 0.91 & 0.02 & 0.00 & 0.48 & 0.48 & 0.04 & 0.00 \\
\cline{1-14}
\multirow[c]{3}{*}{Llama-3.2-3B} & City Locations & 0.92 & 0.04 & 0.04 & 0.00 & 0.03 & 0.93 & 0.04 & 0.00 & 0.39 & 0.58 & 0.03 & 0.00 \\
 & Medical Indications & 0.73 & 0.11 & 0.16 & 0.00 & 0.09 & 0.79 & 0.13 & 0.00 & 0.07 & 0.72 & 0.21 & 0.00 \\
 & Word Definitions & 0.77 & 0.07 & 0.16 & 0.00 & 0.09 & 0.81 & 0.10 & 0.00 & 0.28 & 0.49 & 0.22 & 0.00 \\
\cline{1-14}
\multirow[c]{3}{*}{Llama-3.2-3B-Instruct} & City Locations & 0.88 & 0.12 & 0.00 & 0.00 & 0.09 & 0.91 & 0.00 & 0.00 & 0.89 & 0.11 & 0.00 & 0.00 \\
 & Medical Indications & 0.67 & 0.10 & 0.23 & 0.00 & 0.09 & 0.70 & 0.20 & 0.00 & 0.26 & 0.49 & 0.25 & 0.00 \\
 & Word Definitions & 0.69 & 0.11 & 0.20 & 0.00 & 0.07 & 0.76 & 0.17 & 0.00 & 0.37 & 0.43 & 0.21 & 0.00 \\
\cline{1-14}
\multirow[c]{3}{*}{Llama3-Med42-8B} & City Locations & 0.91 & 0.01 & 0.08 & 0.00 & 0.00 & 0.92 & 0.08 & 0.00 & 0.26 & 0.16 & 0.57 & 0.00 \\
 & Medical Indications & 0.80 & 0.12 & 0.07 & 0.00 & 0.08 & 0.87 & 0.05 & 0.00 & 0.24 & 0.65 & 0.11 & 0.00 \\
 & Word Definitions & 0.91 & 0.09 & 0.00 & 0.00 & 0.07 & 0.92 & 0.00 & 0.00 & 0.51 & 0.49 & 0.00 & 0.00 \\
\cline{1-14}
\multirow[c]{3}{*}{Mistral-7B-Instruct-v0.3} & City Locations & 0.90 & 0.04 & 0.06 & 0.00 & 0.04 & 0.89 & 0.06 & 0.00 & 0.46 & 0.50 & 0.03 & 0.00 \\
 & Medical Indications & 0.79 & 0.11 & 0.10 & 0.00 & 0.09 & 0.81 & 0.10 & 0.00 & 0.48 & 0.46 & 0.07 & 0.00 \\
 & Word Definitions & 0.86 & 0.13 & 0.01 & 0.00 & 0.11 & 0.88 & 0.01 & 0.00 & 0.43 & 0.56 & 0.02 & 0.00 \\
\cline{1-14}
\multirow[c]{3}{*}{Mistral-7B-v0.3} & City Locations & 0.86 & 0.00 & 0.14 & 0.00 & 0.00 & 0.91 & 0.08 & 0.00 & 0.17 & 0.45 & 0.38 & 0.00 \\
 & Medical Indications & 0.80 & 0.10 & 0.10 & 0.00 & 0.10 & 0.84 & 0.07 & 0.00 & 0.42 & 0.46 & 0.12 & 0.00 \\
 & Word Definitions & 0.86 & 0.09 & 0.05 & 0.00 & 0.09 & 0.87 & 0.03 & 0.00 & 0.43 & 0.48 & 0.09 & 0.00 \\
\cline{1-14}
\multirow[c]{3}{*}{Qwen-2.5-14B} & City Locations & 0.92 & 0.01 & 0.07 & 0.00 & 0.01 & 0.93 & 0.06 & 0.00 & 0.13 & 0.47 & 0.40 & 0.00 \\
 & Medical Indications & 0.85 & 0.11 & 0.04 & 0.00 & 0.08 & 0.88 & 0.05 & 0.00 & 0.37 & 0.56 & 0.07 & 0.00 \\
 & Word Definitions & 0.89 & 0.10 & 0.01 & 0.00 & 0.10 & 0.89 & 0.01 & 0.00 & 0.48 & 0.49 & 0.03 & 0.00 \\
\cline{1-14}
\multirow[c]{3}{*}{Qwen-2.5-14B-Instruct} & City Locations & 0.93 & 0.04 & 0.03 & 0.00 & 0.05 & 0.92 & 0.03 & 0.00 & 0.51 & 0.49 & 0.01 & 0.00 \\
 & Medical Indications & 0.87 & 0.11 & 0.02 & 0.00 & 0.09 & 0.89 & 0.02 & 0.00 & 0.23 & 0.71 & 0.06 & 0.00 \\
 & Word Definitions & 0.90 & 0.10 & 0.00 & 0.00 & 0.09 & 0.91 & 0.00 & 0.00 & 0.35 & 0.65 & 0.00 & 0.00 \\
\cline{1-14}
\multirow[c]{3}{*}{Qwen-2.5-7B} & City Locations & 0.89 & 0.06 & 0.06 & 0.00 & 0.03 & 0.92 & 0.04 & 0.00 & 0.52 & 0.46 & 0.02 & 0.00 \\
 & Medical Indications & 0.80 & 0.11 & 0.10 & 0.00 & 0.08 & 0.83 & 0.09 & 0.00 & 0.39 & 0.47 & 0.14 & 0.00 \\
 & Word Definitions & 0.87 & 0.10 & 0.03 & 0.00 & 0.08 & 0.89 & 0.03 & 0.00 & 0.40 & 0.53 & 0.07 & 0.00 \\
\cline{1-14}
\multirow[c]{3}{*}{Qwen-2.5-7B-Instruct} & City Locations & 0.89 & 0.02 & 0.09 & 0.00 & 0.02 & 0.91 & 0.07 & 0.00 & 0.50 & 0.43 & 0.06 & 0.00 \\
 & Medical Indications & 0.80 & 0.11 & 0.08 & 0.00 & 0.12 & 0.82 & 0.06 & 0.00 & 0.44 & 0.49 & 0.06 & 0.00 \\
 & Word Definitions & 0.85 & 0.10 & 0.05 & 0.00 & 0.10 & 0.88 & 0.03 & 0.00 & 0.45 & 0.51 & 0.04 & 0.00 \\
\cline{1-14}
\bottomrule
\end{tabular}
\end{adjustbox}
\caption{\textbf{Row-wise confusion matrices for supervised PCA probe with conformal prediction (\texttt{sPCA+CP}) across all $\langle$model-dataset pairs$\rangle$} (evaluated on the \underline{\textit{last token representation}}).  
Each row corresponds to a specific model and a dataset. 
Columns are grouped by the ground-truth labels (\textit{True}, \textit{False}, \textit{Neither}) with groups of subcolumns that specify the distribution of predictions (\textit{true}, \textit{false}, \textit{neither}, \textit{abstain}).  
For each statement in a dataset, the predicted class is the class with the highest probability (as estimated by \texttt{sPCA+CP}).
The values in each group of four subcolumns sum to \(1\) because they are normalized counts.
For example, in the first row under the \textit{True} ground-truth label, we see that \textit{true} predictions have the value of \(0.90\); that is, \(90\%\) of all the true statements are classified as true.}
\label{sup_tab:cm_spca_instance}
\end{table}
\end{small}

\begin{small}
\begin{table}
\centering
\begin{adjustbox}{width=\textwidth}
\begin{tabular}{ll|cccc|cccc|cccc}
\toprule
 & True Labels \(\rightarrow\)
   & \multicolumn{4}{c|}{True } 
   & \multicolumn{4}{c|}{False} 
   & \multicolumn{4}{c}{Neither} \\
 & Predicted  \(\rightarrow\)
   & True & False & Neither & Abstain 
   & True & False & Neither & Abstain 
   & True & False & Neither & Abstain 
\\
Model  \(\downarrow\)& Dataset \(\downarrow\)
   &  &  &  &  
   &  &  &  &  
   &  &  &  &   \\
\midrule
\multirow[c]{3}{*}{Bio-Medical-Llama-3-8B} & City Locations & 1.00 & 0.00 & 0.00 & 0.00 & 1.00 & 0.00 & 0.00 & 0.00 & 1.00 & 0.00 & 0.00 & 0.00 \\
 & Medical Indications & 1.00 & 0.00 & 0.00 & 0.00 & 1.00 & 0.00 & 0.00 & 0.00 & 1.00 & 0.00 & 0.00 & 0.00 \\
 & Word Definitions & 0.81 & 0.03 & 0.17 & 0.00 & 0.72 & 0.07 & 0.21 & 0.00 & 0.68 & 0.07 & 0.25 & 0.00 \\
\cline{1-14}
\multirow[c]{3}{*}{Gemma-2-9B} & City Locations & 0.98 & 0.00 & 0.02 & 0.00 & 0.54 & 0.04 & 0.41 & 0.00 & 0.06 & 0.24 & 0.70 & 0.00 \\
 & Medical Indications & 0.86 & 0.03 & 0.11 & 0.00 & 0.44 & 0.33 & 0.23 & 0.00 & 0.00 & 0.84 & 0.15 & 0.00 \\
 & Word Definitions & 0.96 & 0.01 & 0.03 & 0.00 & 0.88 & 0.06 & 0.07 & 0.00 & 0.71 & 0.16 & 0.13 & 0.00 \\
\cline{1-14}
\multirow[c]{3}{*}{Gemma-2-9B-it} & City Locations & 1.00 & 0.00 & 0.00 & 0.00 & 1.00 & 0.00 & 0.00 & 0.00 & 1.00 & 0.00 & 0.00 & 0.00 \\
 & Medical Indications & 1.00 & 0.00 & 0.00 & 0.00 & 1.00 & 0.00 & 0.00 & 0.00 & 1.00 & 0.00 & 0.00 & 0.00 \\
 & Word Definitions & 0.95 & 0.00 & 0.05 & 0.00 & 0.61 & 0.09 & 0.31 & 0.00 & 0.55 & 0.15 & 0.30 & 0.00 \\
\cline{1-14}
\multirow[c]{3}{*}{Gemma-7B} & City Locations & 1.00 & 0.00 & 0.00 & 0.00 & 0.97 & 0.01 & 0.02 & 0.00 & 1.00 & 0.00 & 0.00 & 0.00 \\
 & Medical Indications & 0.27 & 0.01 & 0.72 & 0.00 & 0.18 & 0.03 & 0.79 & 0.00 & 0.40 & 0.05 & 0.55 & 0.00 \\
 & Word Definitions & 0.82 & 0.00 & 0.17 & 0.00 & 0.53 & 0.03 & 0.44 & 0.00 & 0.53 & 0.03 & 0.44 & 0.00 \\
\cline{1-14}
\multirow[c]{3}{*}{Gemma-7B-it} & City Locations & 0.90 & 0.01 & 0.09 & 0.00 & 0.02 & 0.82 & 0.16 & 0.00 & 0.11 & 0.51 & 0.37 & 0.00 \\
 & Medical Indications & 1.00 & 0.00 & 0.00 & 0.00 & 1.00 & 0.00 & 0.00 & 0.00 & 1.00 & 0.00 & 0.00 & 0.00 \\
 & Word Definitions & 0.88 & 0.01 & 0.11 & 0.00 & 0.37 & 0.15 & 0.48 & 0.00 & 0.52 & 0.14 & 0.35 & 0.00 \\
\cline{1-14}
\multirow[c]{3}{*}{Llama-3-8B} & City Locations & 0.96 & 0.03 & 0.02 & 0.00 & 0.42 & 0.49 & 0.09 & 0.00 & 0.54 & 0.40 & 0.06 & 0.00 \\
 & Medical Indications & 0.82 & 0.10 & 0.08 & 0.00 & 0.12 & 0.78 & 0.10 & 0.00 & 0.44 & 0.45 & 0.11 & 0.00 \\
 & Word Definitions & 0.95 & 0.01 & 0.04 & 0.00 & 0.86 & 0.02 & 0.12 & 0.00 & 0.78 & 0.06 & 0.16 & 0.00 \\
\cline{1-14}
\multirow[c]{3}{*}{Llama-3.1-8B-Instruct} & City Locations & 0.91 & 0.00 & 0.09 & 0.00 & 0.00 & 0.76 & 0.24 & 0.00 & 0.05 & 0.35 & 0.59 & 0.00 \\
 & Medical Indications & 1.00 & 0.00 & 0.00 & 0.00 & 1.00 & 0.00 & 0.00 & 0.00 & 1.00 & 0.00 & 0.00 & 0.00 \\
 & Word Definitions & 1.00 & 0.00 & 0.00 & 0.00 & 1.00 & 0.00 & 0.00 & 0.00 & 1.00 & 0.00 & 0.00 & 0.00 \\
\cline{1-14}
\multirow[c]{3}{*}{Llama-3.2-3B} & City Locations & 0.91 & 0.04 & 0.05 & 0.00 & 0.05 & 0.87 & 0.08 & 0.00 & 0.24 & 0.53 & 0.23 & 0.00 \\
 & Medical Indications & 0.75 & 0.05 & 0.20 & 0.00 & 0.12 & 0.41 & 0.47 & 0.00 & 0.18 & 0.48 & 0.34 & 0.00 \\
 & Word Definitions & 0.75 & 0.00 & 0.24 & 0.00 & 0.43 & 0.02 & 0.55 & 0.00 & 0.41 & 0.03 & 0.56 & 0.00 \\
\cline{1-14}
\multirow[c]{3}{*}{Llama-3.2-3B-Instruct} & City Locations & 1.00 & 0.00 & 0.00 & 0.00 & 1.00 & 0.00 & 0.00 & 0.00 & 1.00 & 0.00 & 0.00 & 0.00 \\
 & Medical Indications & 1.00 & 0.00 & 0.00 & 0.00 & 1.00 & 0.00 & 0.00 & 0.00 & 1.00 & 0.00 & 0.00 & 0.00 \\
 & Word Definitions & 1.00 & 0.00 & 0.00 & 0.00 & 1.00 & 0.00 & 0.00 & 0.00 & 1.00 & 0.00 & 0.00 & 0.00 \\
\cline{1-14}
\multirow[c]{3}{*}{Llama3-Med42-8B} & City Locations & 1.00 & 0.00 & 0.00 & 0.00 & 1.00 & 0.00 & 0.00 & 0.00 & 1.00 & 0.00 & 0.00 & 0.00 \\
 & Medical Indications & 0.94 & 0.01 & 0.05 & 0.00 & 0.67 & 0.17 & 0.16 & 0.00 & 0.89 & 0.04 & 0.07 & 0.00 \\
 & Word Definitions & 1.00 & 0.00 & 0.00 & 0.00 & 1.00 & 0.00 & 0.00 & 0.00 & 1.00 & 0.00 & 0.00 & 0.00 \\
\cline{1-14}
\multirow[c]{3}{*}{Mistral-7B-Instruct-v0.3} & City Locations & 0.88 & 0.01 & 0.11 & 0.00 & 0.01 & 0.86 & 0.13 & 0.00 & 0.26 & 0.16 & 0.59 & 0.00 \\
 & Medical Indications & 1.00 & 0.00 & 0.00 & 0.00 & 1.00 & 0.00 & 0.00 & 0.00 & 1.00 & 0.00 & 0.00 & 0.00 \\
 & Word Definitions & 0.84 & 0.01 & 0.15 & 0.00 & 0.76 & 0.03 & 0.21 & 0.00 & 0.79 & 0.02 & 0.19 & 0.00 \\
\cline{1-14}
\multirow[t]{3}{*}{Mistral-7B-v0.3} & City Locations & 1.00 & 0.00 & 0.00 & 0.00 & 0.98 & 0.00 & 0.02 & 0.00 & 1.00 & 0.00 & 0.00 & 0.00 \\
 & Medical Indications & 0.80 & 0.06 & 0.15 & 0.00 & 0.11 & 0.66 & 0.23 & 0.00 & 0.24 & 0.49 & 0.27 & 0.00 \\
 & Word Definitions & 0.87 & 0.02 & 0.10 & 0.00 & 0.49 & 0.19 & 0.32 & 0.00 & 0.42 & 0.30 & 0.28 & 0.00 \\
\cline{1-14}
\multirow[t]{3}{*}{Qwen-2.5-14B} & City Locations & 0.98 & 0.00 & 0.02 & 0.00 & 0.45 & 0.32 & 0.24 & 0.00 & 0.53 & 0.37 & 0.10 & 0.00 \\
 & Medical Indications & 0.83 & 0.06 & 0.11 & 0.00 & 0.26 & 0.37 & 0.37 & 0.00 & 0.17 & 0.49 & 0.34 & 0.00 \\
 & Word Definitions & 0.96 & 0.00 & 0.03 & 0.00 & 0.87 & 0.03 & 0.10 & 0.00 & 0.89 & 0.02 & 0.09 & 0.00 \\
\cline{1-14}
\multirow[t]{3}{*}{Qwen-2.5-14B-Instruct} & City Locations & 1.00 & 0.00 & 0.00 & 0.00 & 1.00 & 0.00 & 0.00 & 0.00 & 1.00 & 0.00 & 0.00 & 0.00 \\
 & Medical Indications & 0.84 & 0.08 & 0.09 & 0.00 & 0.10 & 0.82 & 0.09 & 0.00 & 0.15 & 0.71 & 0.14 & 0.00 \\
 & Word Definitions & 1.00 & 0.00 & 0.00 & 0.00 & 1.00 & 0.00 & 0.00 & 0.00 & 1.00 & 0.00 & 0.00 & 0.00 \\
\cline{1-14}
\multirow[t]{3}{*}{Qwen-2.5-7B} & City Locations & 0.93 & 0.01 & 0.06 & 0.00 & 0.19 & 0.54 & 0.28 & 0.00 & 0.22 & 0.51 & 0.27 & 0.00 \\
 & Medical Indications & 0.36 & 0.05 & 0.59 & 0.00 & 0.32 & 0.08 & 0.59 & 0.00 & 0.23 & 0.17 & 0.60 & 0.00 \\
 & Word Definitions & 0.97 & 0.00 & 0.03 & 0.00 & 0.92 & 0.01 & 0.06 & 0.00 & 0.97 & 0.01 & 0.03 & 0.00 \\
\cline{1-14}
\multirow[t]{3}{*}{Qwen-2.5-7B-Instruct} & City Locations & 1.00 & 0.00 & 0.00 & 0.00 & 1.00 & 0.00 & 0.00 & 0.00 & 1.00 & 0.00 & 0.00 & 0.00 \\
 & Medical Indications & 1.00 & 0.00 & 0.00 & 0.00 & 1.00 & 0.00 & 0.00 & 0.00 & 1.00 & 0.00 & 0.00 & 0.00 \\
 & Word Definitions & 0.85 & 0.09 & 0.06 & 0.00 & 0.16 & 0.77 & 0.07 & 0.00 & 0.36 & 0.53 & 0.11 & 0.00 \\
\cline{1-14}
\bottomrule
\end{tabular}
\end{adjustbox}
\caption{\textbf{Row-wise confusion matrices for supervised PCA probe with conformal prediction (\texttt{sPCA+CP}) across all $\langle$model-dataset pairs$\rangle$} (evaluated on the \underline{\textit{bag}}).  
Each row corresponds to a specific model and a dataset. 
Columns are grouped by the ground-truth labels (\textit{True}, \textit{False}, \textit{Neither}) with groups of subcolumns that specify the distribution of predictions (\textit{true}, \textit{false}, \textit{neither}, \textit{abstain}).  
For each statement in a dataset, the predicted class is the class with the highest probability (as estimated by \texttt{sPCA+CP}).
The values in each group of four subcolumns sum to \(1\) because they are normalized counts.
For example, in the first row under the \textit{True} ground-truth label, we see that \textit{true} predictions have the value of \(1.00\); that is, \(100\%\) of all the true statements are classified as true.}
\label{sup_tab:cm_spca_bag}
\end{table}
\end{small}

\begin{small}
\begin{table}
\centering
\begin{adjustbox}{width=\textwidth}
\begin{tabular}{ll|cccc|cccc|cccc}
\toprule
 & True Labels \(\rightarrow\)
   & \multicolumn{4}{c|}{True } 
   & \multicolumn{4}{c|}{False} 
   & \multicolumn{4}{c}{Neither} \\
 & Predicted  \(\rightarrow\)
   & True & False & Neither & Abstain 
   & True & False & Neither & Abstain 
   & True & False & Neither & Abstain 
\\
Model  \(\downarrow\)& Dataset \(\downarrow\)
   &  &  &  &  
   &  &  &  &  
   &  &  &  &   \\
\midrule
\multirow[c]{3}{*}{Bio-Medical-Llama-3-8B} & City Locations & 0.80 & 0.01 & 0.00 & 0.18 & 0.00 & 0.88 & 0.00 & 0.11 & 0.00 & 0.00 & 1.00 & 0.00 \\
 & Medical Indications & 0.77 & 0.10 & 0.00 & 0.12 & 0.10 & 0.81 & 0.01 & 0.08 & 0.00 & 0.00 & 1.00 & 0.00 \\
 & Word Definitions & 0.86 & 0.09 & 0.01 & 0.03 & 0.11 & 0.87 & 0.01 & 0.02 & 0.00 & 0.01 & 0.98 & 0.01 \\
\cline{1-14}
\multirow[c]{3}{*}{Gemma-2-9B} & City Locations & 0.87 & 0.00 & 0.00 & 0.13 & 0.00 & 0.85 & 0.00 & 0.15 & 0.00 & 0.00 & 1.00 & 0.00 \\
 & Medical Indications & 0.81 & 0.08 & 0.01 & 0.11 & 0.10 & 0.73 & 0.01 & 0.16 & 0.01 & 0.00 & 0.97 & 0.02 \\
 & Word Definitions & 0.86 & 0.10 & 0.00 & 0.04 & 0.11 & 0.85 & 0.01 & 0.03 & 0.01 & 0.01 & 0.98 & 0.00 \\
\cline{1-14}
\multirow[c]{3}{*}{Gemma-2-9B-it} & City Locations & 0.85 & 0.02 & 0.02 & 0.12 & 0.02 & 0.87 & 0.00 & 0.11 & 0.00 & 0.00 & 0.98 & 0.02 \\
 & Medical Indications & 0.76 & 0.10 & 0.01 & 0.13 & 0.07 & 0.84 & 0.01 & 0.09 & 0.01 & 0.00 & 0.99 & 0.01 \\
 & Word Definitions & 0.83 & 0.08 & 0.00 & 0.08 & 0.07 & 0.88 & 0.01 & 0.05 & 0.00 & 0.01 & 0.98 & 0.01 \\
\cline{1-14}
\multirow[c]{3}{*}{Gemma-7B} & City Locations & 0.86 & 0.01 & 0.00 & 0.13 & 0.01 & 0.89 & 0.00 & 0.11 & 0.00 & 0.00 & 1.00 & 0.00 \\
 & Medical Indications & 0.75 & 0.09 & 0.00 & 0.17 & 0.08 & 0.73 & 0.01 & 0.18 & 0.01 & 0.01 & 0.95 & 0.04 \\
 & Word Definitions & 0.82 & 0.11 & 0.04 & 0.02 & 0.15 & 0.81 & 0.01 & 0.03 & 0.02 & 0.01 & 0.95 & 0.01 \\
\cline{1-14}
\multirow[c]{3}{*}{Gemma-7B-it} & City Locations & 0.86 & 0.02 & 0.01 & 0.11 & 0.02 & 0.87 & 0.00 & 0.11 & 0.00 & 0.00 & 0.99 & 0.01 \\
 & Medical Indications & 0.58 & 0.05 & 0.01 & 0.35 & 0.15 & 0.50 & 0.00 & 0.35 & 0.00 & 0.01 & 0.94 & 0.05 \\
 & Word Definitions & 0.69 & 0.16 & 0.04 & 0.11 & 0.12 & 0.77 & 0.03 & 0.07 & 0.01 & 0.01 & 0.96 & 0.01 \\
\cline{1-14}
\multirow[c]{3}{*}{Llama-3-8B} & City Locations & 0.87 & 0.01 & 0.00 & 0.12 & 0.01 & 0.85 & 0.00 & 0.14 & 0.00 & 0.00 & 1.00 & 0.00 \\
 & Medical Indications & 0.78 & 0.09 & 0.01 & 0.12 & 0.11 & 0.77 & 0.01 & 0.11 & 0.00 & 0.01 & 0.98 & 0.01 \\
 & Word Definitions & 0.87 & 0.12 & 0.00 & 0.00 & 0.13 & 0.85 & 0.01 & 0.00 & 0.01 & 0.01 & 0.97 & 0.00 \\
\cline{1-14}
\multirow[c]{3}{*}{Llama-3.1-8B-Instruct} & City Locations & 0.87 & 0.02 & 0.00 & 0.11 & 0.02 & 0.86 & 0.00 & 0.13 & 0.00 & 0.00 & 1.00 & 0.00 \\
 & Medical Indications & 0.80 & 0.12 & 0.00 & 0.09 & 0.09 & 0.85 & 0.00 & 0.06 & 0.00 & 0.00 & 1.00 & 0.00 \\
 & Word Definitions & 0.87 & 0.06 & 0.00 & 0.07 & 0.07 & 0.86 & 0.00 & 0.06 & 0.00 & 0.01 & 0.98 & 0.01 \\
\cline{1-14}
\multirow[c]{3}{*}{Llama-3.2-3B} & City Locations & 0.84 & 0.03 & 0.00 & 0.12 & 0.04 & 0.86 & 0.01 & 0.09 & 0.00 & 0.01 & 0.99 & 0.01 \\
 & Medical Indications & 0.71 & 0.10 & 0.00 & 0.19 & 0.09 & 0.70 & 0.01 & 0.20 & 0.00 & 0.00 & 0.98 & 0.02 \\
 & Word Definitions & 0.74 & 0.14 & 0.02 & 0.10 & 0.11 & 0.77 & 0.02 & 0.10 & 0.01 & 0.01 & 0.96 & 0.02 \\
\cline{1-14}
\multirow[c]{3}{*}{Llama-3.2-3B-Instruct} & City Locations & 0.86 & 0.02 & 0.00 & 0.11 & 0.02 & 0.82 & 0.00 & 0.16 & 0.00 & 0.00 & 1.00 & 0.00 \\
 & Medical Indications & 0.67 & 0.12 & 0.00 & 0.21 & 0.11 & 0.72 & 0.00 & 0.18 & 0.00 & 0.00 & 0.99 & 0.01 \\
 & Word Definitions & 0.86 & 0.11 & 0.03 & 0.00 & 0.14 & 0.85 & 0.01 & 0.00 & 0.01 & 0.01 & 0.98 & 0.00 \\
\cline{1-14}
\multirow[c]{3}{*}{Llama3-Med42-8B} & City Locations & 0.87 & 0.01 & 0.00 & 0.12 & 0.01 & 0.84 & 0.00 & 0.15 & 0.00 & 0.00 & 0.99 & 0.01 \\
 & Medical Indications & 0.81 & 0.10 & 0.00 & 0.10 & 0.11 & 0.82 & 0.00 & 0.07 & 0.00 & 0.00 & 1.00 & 0.00 \\
 & Word Definitions & 0.86 & 0.06 & 0.01 & 0.06 & 0.08 & 0.85 & 0.00 & 0.07 & 0.01 & 0.01 & 0.97 & 0.02 \\
\cline{1-14}
\multirow[c]{3}{*}{Mistral-7B-Instruct-v0.3} & City Locations & 0.86 & 0.01 & 0.00 & 0.13 & 0.01 & 0.79 & 0.00 & 0.19 & 0.00 & 0.00 & 1.00 & 0.00 \\
 & Medical Indications & 0.77 & 0.09 & 0.00 & 0.14 & 0.10 & 0.76 & 0.00 & 0.14 & 0.00 & 0.01 & 0.99 & 0.01 \\
 & Word Definitions & 0.84 & 0.08 & 0.03 & 0.05 & 0.09 & 0.87 & 0.01 & 0.03 & 0.01 & 0.01 & 0.98 & 0.01 \\
\cline{1-14}
\multirow[c]{3}{*}{Mistral-7B-v0.3} & City Locations & 0.86 & 0.01 & 0.00 & 0.13 & 0.01 & 0.86 & 0.00 & 0.13 & 0.00 & 0.00 & 1.00 & 0.00 \\
 & Medical Indications & 0.75 & 0.08 & 0.00 & 0.17 & 0.09 & 0.76 & 0.00 & 0.15 & 0.00 & 0.00 & 0.99 & 0.01 \\
 & Word Definitions & 0.87 & 0.11 & 0.02 & 0.00 & 0.14 & 0.85 & 0.01 & 0.00 & 0.01 & 0.01 & 0.98 & 0.00 \\
\cline{1-14}
\multirow[c]{3}{*}{Qwen-2.5-14B} & City Locations & 0.87 & 0.01 & 0.00 & 0.12 & 0.01 & 0.86 & 0.00 & 0.13 & 0.00 & 0.00 & 0.99 & 0.01 \\
 & Medical Indications & 0.78 & 0.12 & 0.00 & 0.10 & 0.11 & 0.76 & 0.00 & 0.13 & 0.00 & 0.00 & 0.99 & 0.00 \\
 & Word Definitions & 0.86 & 0.12 & 0.02 & 0.00 & 0.13 & 0.86 & 0.01 & 0.00 & 0.01 & 0.02 & 0.98 & 0.00 \\
\cline{1-14}
\multirow[c]{3}{*}{Qwen-2.5-14B-Instruct} & City Locations & 0.82 & 0.01 & 0.00 & 0.17 & 0.01 & 0.87 & 0.00 & 0.11 & 0.00 & 0.00 & 1.00 & 0.00 \\
 & Medical Indications & 0.80 & 0.12 & 0.01 & 0.07 & 0.07 & 0.86 & 0.01 & 0.07 & 0.00 & 0.00 & 0.99 & 0.00 \\
 & Word Definitions & 0.86 & 0.07 & 0.01 & 0.05 & 0.06 & 0.86 & 0.00 & 0.08 & 0.00 & 0.01 & 0.98 & 0.01 \\
\cline{1-14}
\multirow[c]{3}{*}{Qwen-2.5-7B} & City Locations & 0.84 & 0.01 & 0.00 & 0.16 & 0.01 & 0.84 & 0.00 & 0.14 & 0.00 & 0.00 & 1.00 & 0.00 \\
 & Medical Indications & 0.74 & 0.10 & 0.01 & 0.15 & 0.10 & 0.77 & 0.01 & 0.13 & 0.00 & 0.00 & 0.98 & 0.01 \\
 & Word Definitions & 0.83 & 0.12 & 0.02 & 0.03 & 0.10 & 0.87 & 0.01 & 0.03 & 0.01 & 0.01 & 0.97 & 0.01 \\
\cline{1-14}
\multirow[c]{3}{*}{Qwen-2.5-7B-Instruct} & City Locations & 0.82 & 0.02 & 0.01 & 0.15 & 0.02 & 0.85 & 0.00 & 0.13 & 0.00 & 0.00 & 0.99 & 0.01 \\
 & Medical Indications & 0.78 & 0.08 & 0.01 & 0.13 & 0.14 & 0.72 & 0.02 & 0.12 & 0.01 & 0.01 & 0.98 & 0.01 \\
 & Word Definitions & 0.83 & 0.13 & 0.02 & 0.02 & 0.09 & 0.87 & 0.02 & 0.02 & 0.01 & 0.01 & 0.97 & 0.01 \\
\bottomrule
\end{tabular}
\end{adjustbox}
\caption{\textbf{Row-wise confusion matrices for the multiclass \texttt{sAwMIL} across all $\langle$model-dataset pairs$\rangle$} (evaluated on the \underline{\textit{bag}}).  
Each row corresponds to a specific model and a dataset. 
Columns are grouped by the ground-truth labels (\textit{True}, \textit{False}, \textit{Neither}) with groups of subcolumns that specify the distribution of predictions (\textit{true}, \textit{false}, \textit{neither}, \textit{abstain}).  
For each statement in a dataset, the predicted class is the class with the highest probability (as estimated by multiclass \texttt{sAwMIL}).
The values in each group of four subcolumns sum to \(1\) because they are normalized counts.
For example, in the first row under the \textit{True} ground-truth label, we see that \textit{true} predictions have the value of \(0.80\); that is, \(80\%\) of all the true statements are classified as true.
}
\label{sup_tab:cm_sawmil_bag}
\end{table}
\end{small}

\clearpage
\onecolumn
\section{Related Resources}
\label{supsec:related}

\subsection*{Code}
{\raggedright
The code associated with this manuscript is publicly available on GitHub at \href{https://github.com/carlomarxdk/trilemma-of-truth}{\texttt{carlomarxdk/trilemma-of-truth}} (release version~1.0.0).
\par}

\subsection*{Data}
{\raggedright
The data used in this study are publicly available on Hugging Face at \href{https://huggingface.co/datasets/carlomarxx/trilemma-of-truth}{\texttt{carlomarxx/trilemma-of-truth}}   (DOI:~\href{https://doi.org/10.57967/hf/5900}{\texttt{10.57967/hf/5900}}).
\par}


\end{document}